\pdfoutput=1
\documentclass[]{youtu}

\usepackage{mathpazo}
\usepackage{graphicx}
\usepackage{float}
\usepackage[numbers,sort&compress]{natbib}
\usepackage{algorithm}
\usepackage{algpseudocode}
\usepackage{amsthm}
\usepackage{colortbl}
\definecolor{lmHeader}{RGB}{238,242,247}
\definecolor{lmOurs}{RGB}{232,243,253}
\definecolor{lmAccent}{RGB}{47,79,111}
\newtcolorbox{keytakeaway}{
    enhanced, breakable,
    colback=lmOurs, colframe=lmAccent, boxrule=0.6pt,
    left=4pt, right=4pt, top=3pt, bottom=3pt, arc=1.5pt,
    fontupper=\small,
}

\usepackage[margin=1in]{geometry}
\usepackage[table]{xcolor}
\usepackage{booktabs}
\usepackage{array}
\usepackage{multirow}
\usepackage{graphicx}
\usepackage[numbers]{natbib}
\usepackage{url}
\usepackage{hyperref}
\usepackage[skip=6pt]{caption}
\usepackage{tabularx}
\definecolor{youtuhl}{RGB}{232,241,252}

\renewcommand{\methodname}{Youtu-VL 2.0}

\newcommand{\papertitle}{Youtu-VL 2.0: Unlocking Native Multimodal Intelligence with Modality-Aware Capacity, Routing, and Memory}
\title{\fontsize{16}{19}\selectfont\papertitle}
\hypersetup{pdftitle={\papertitle}}
\newcommand{\authorfootnote}{\begingroup\renewcommand{\thefootnote}{\fnsymbol{footnote}}%
  \footnote{Contributors are listed in Appendix~\ref{sec:contributions}.}%
  \addtocounter{footnote}{-1}\endgroup}
\author{Youtu Foundation Model Team\protect\authorfootnote}

\youtufinalcopy

\begin{document}
\abstract{We introduce Youtu-VL 2.0, a native multimodal Mixture-of-Experts architecture, and pre-train Youtu-VL2.0-32B-A4B-Base on this design. Visual features from a Qwen3-VL vision transformer enter a shared autoregressive decoder that interleaves KDA and gated DSA layers at a 3:1 ratio. The backbone pairs fixed-size recurrent states with sparsely attended, latent-compressed key-value caches, and four-stream Gated Residual connections regulate information flow across depth. To balance cross-modal fusion with modality-specific specialization, a single token-level modality label coordinates three mechanisms. Specifically, MMoE partitions always-active shared experts by modality while preserving a common routed pool, modality-aware Quantile Balancing regulates routed capacity using independent text and vision bias states without auxiliary losses, and text-selective Engram provides lexical $N$-gram memory while zeroing lookup residuals at visual positions. We train the model on 15.2T tokens using a multi-stage curriculum that enforces exact per-step vision-language sample quotas per data-parallel rank, scaling to a 256K context window with square-root sample-length reweighting and a warm-up transition to sparse DSA. A co-designed training stack fuses routing, shared-expert packing, and node-local memory lookups, complemented by continuous telemetry and event-triggered loss-spike attribution. Across standard text protocols, continuation-based multimodal benchmarks, and 256K long-context evaluations, Youtu-VL2.0-32B-A4B-Base achieves strong performance.
}
\maketitle

\vspace{-.1em}

\begin{figure}[htbp]
    \centering
    \includegraphics[width=1.0\textwidth]{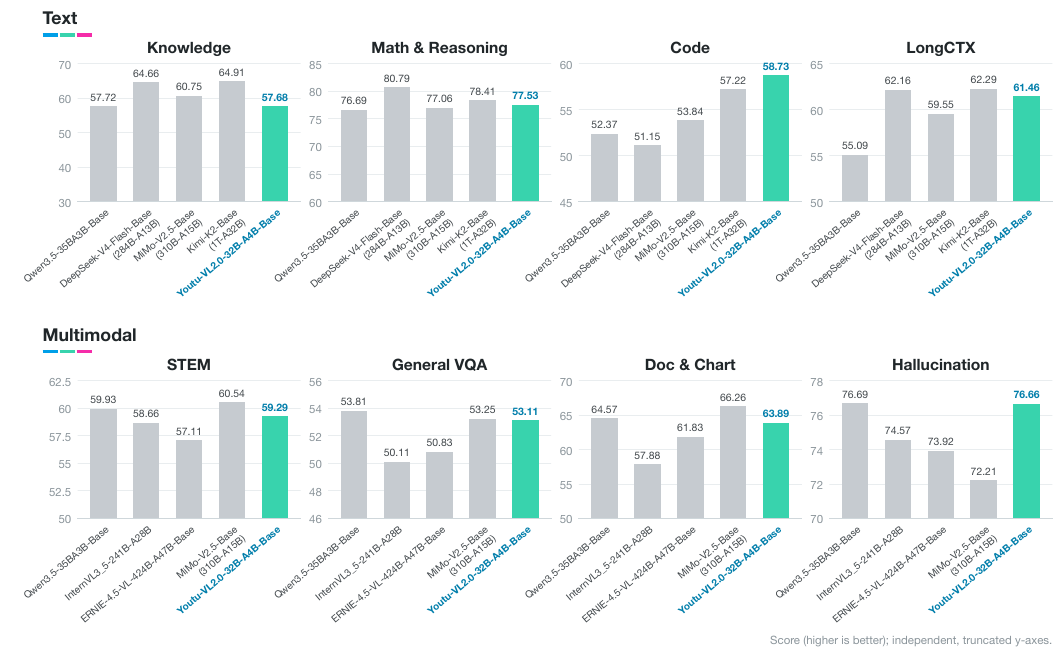}
    \caption{Youtu-VL2.0-32B-A4B-Base main results.}
    \label{fig:main_results}
\end{figure}

\section{Introduction}
\label{sec:intro}

Multimodal large language models (MLLMs) face a fundamental architectural tension between early cross-modal fusion and modality-specific specialization. Early fusion is motivated by the need for text and visual inputs to interact early, grounding shared concepts and enabling cross-modal reasoning. Yet the two modalities differ markedly in local structure, token statistics, spatial layout, and optimization dynamics. Fully separating their computation can limit parameter reuse and cross-modal interaction, whereas a single shared path may struggle to accommodate modality-specific features. Compounding this challenge, high-resolution visual inputs increase sequence length, placing additional pressure on attention computation and key-value (KV) cache storage \cite{qiu2026qwen38next,zai2026glm53flash}.

Recent architectures address computational efficiency through sparse conditional computation and hybrid sequence modeling. Mixture-of-Experts (MoE) layers decouple total parameter capacity from per-token compute \cite{shazeer2017outrageously,lepikhin2021gshard,fedus2022switch}. Multi-head Latent Attention (MLA) compresses the attention cache into a compact latent representation \cite{deepseekai2024deepseekv2}. DeltaNet and Kimi Delta Attention (KDA) summarize sequence history in fixed-size recurrent states \cite{yang2024parallelizing,kimiteam2025kimilinear}. Kimi Linear, Kimi K3, and the Ling-3 series interleave KDA with periodic MLA layers \cite{kimiteam2025kimilinear,kimiteam2026kimik3,inclusionai2026ling30flash,inclusionai2026ling31flash}. Qwen3.8-Flash-Next and GLM-5.3-Flash pair recurrent layers with sparse attention, limiting the number of positions read by the main attention \cite{qiu2026qwen38next,zai2026glm53flash}. Sparse selection and latent cache compression address complementary costs and can be combined. These efficiency mechanisms do not, by themselves, determine how a native multimodal MoE should allocate dedicated capacity, regulate routed expert loads, or apply parametric memory across text and vision.

To bridge this gap, we introduce \textbf{Youtu-VL2.0-32B-A4B-Base}, the first and most compact member of the Youtu-VL 2.0 family. Visual features extracted by a Qwen3-VL vision transformer (ViT) enter the same autoregressive decoder alongside text tokens from the onset of pre-training, fostering unified cross-modal representations throughout optimization. The 48-layer decoder adopts the 3:1 hybrid ratio of Kimi Linear, alternating between three KDA layers and one gated DeepSeek Sparse Attention (DSA) layer \cite{deepseekai2025deepseekv32}. DSA couples the latent key-value compression of MLA with a lightweight indexer that selects a bounded set of causal positions per query. These layers operate as dense gated MLA during early training stages and transition to sparse selection in the final 256K-context stage following an indexer-only warm-up. Throughout the decoder's depth, four-stream Gated Residual (GR) connections propagate hidden states \cite{qiu2026qwen38next}. Except for a dense feed-forward network in the first layer, all subsequent layers employ sparse MoE with 128 routed experts and top-8 activation. A conditional memory module at the second decoder layer adds hashed $N$-gram embeddings to the residual stream.

Within this shared decoder, we preserve cross-modal interaction while introducing modality-specific specialization in selected components. Specifically, a single token-level modality label governs three allocation decisions: where dedicated capacity resides, how routed capacity is allocated, and which positions access parametric memory. First, for \textbf{modality-dedicated capacity}, we introduce Multimodal Shared-expert MoE (MMoE): routed experts remain shared across modalities to support cross-modal transfer, whereas the always-active shared path is partitioned into modality-specific branches, reserving dedicated capacity without splitting the backbone. Related work also reports benefits from per-modality shared experts in controlled settings \cite{tong2026beyond}. Here, we integrate this design into 32B multi-stage pre-training alongside modality-aware routing, conditional memory, and fused kernel execution.

Second, for \textbf{modality-aware routing}, we extend auxiliary-loss-free load balancing to heterogeneous token distributions. Kimi K3 balances expert loads without auxiliary losses via Quantile Balancing (QB), which updates a single expert-wise bias vector across all routed tokens \cite{kimiteam2026kimik3}. However, pooled load balancing can conceal imbalances within either modality, particularly when text and vision contribute different token counts. Inspired by the separate text and image bias updates in DeepSeek-V4.1-Flash \cite{deepseekai2026deepseekv41flash}, we bring modality separation to QB: each router maintains independent bias states and required-bias histograms for text and visual tokens, with the same target selection fraction for each expert in both modalities. Expert selection uses the bias corresponding to each token's modality, preserving QB's auxiliary-loss-free design while enabling modality-conditional load control.

Third, for \textbf{position-selective parametric memory}, we apply Engram \cite{cheng2026conditional}, a conditional memory module based on scalable $N$-gram lookup, only at text positions. Text tokens receive hashed bigram and trigram residuals, whereas visual positions receive zero residual and reset the $N$-gram and short-convolution states. Since visual positions carry placeholder token IDs, hashing repeated placeholders would access the same small set of $N$-gram rows regardless of image content. Masking these positions keeps the lookup table lexical. The table holds roughly 8 billion lookup values and is maintained through data-parallel row sharding, block-scaled FP8 storage, and sparse updates.

To jointly optimize the shared backbone and its modality-conditioned components, we extend our prior multi-stage data and context curriculum \cite{lu2025youtullm} with stricter modality handling, data validation, and stage-transition checks. At every global step, deterministic blending converts target data mixtures into exact vision-language sample quotas per data-parallel (DP) rank. These quotas fix the sample-type composition of each step, while visual-token exposure and encoder workload still depend on image counts and resolutions. Early stages are text-dominant, with multimodal tokens accounting for 10\% to 15\% of the trained volume. Later stages increase the share of VQA and trajectory data within the vision-language mixture. Shorter-context stages compute token-averaged losses. At 256K, square-root sample-length reweighting over logical packed samples reduces the relative loss contribution of exceptionally long samples.

To support this curriculum at scale, we develop an infrastructure stack tailored to the hybrid architecture. It uses fused kernels for KDA and gated DSA token mixing as well as sparse MoE execution, while managing architecture-specific state such as GR streams and the FP8 Engram table across devices. Checkpoint conversion and serving implementations preserve the trained model's numerical behavior.

We also integrate two complementary monitoring mechanisms into the training infrastructure: continuous telemetry and event-triggered diagnostics. Continuous telemetry tracks modality-separated losses, component gradient norms, and Engram update statistics. Routing diagnostics include expert loads and maximum load violation (MaxVio) for each modality, along with shared-expert activations. Throughput and device memory usage are recorded to monitor systems health. When a loss spike is detected, the monitor records source indices and reconstruction metadata for high-loss micro-batches without interrupting training. Offline tools use these records to reconstruct the corresponding text and images for sample-level diagnosis, informing data curation and pipeline fixes for subsequent model iterations.

Finally, we evaluate the resulting base checkpoint across three core capability axes: text, multimodal, and long-context understanding. The text track follows standard likelihood- and generation-based protocols spanning world knowledge, reasoning, mathematics, multilinguality, coding, and repository completion \cite{hendrycks2021mmlu,rein2024gpqa,cobbe2021gsm8k,suzgun2023bbh,chen2021humaneval,austin2021mbpp,cassano2023multiple,liu2024repobench}. The multimodal track reformulates instruction-oriented benchmarks into few-shot conditional continuation tasks, reducing reliance on chat templates by specifying the expected answer format through in-context demonstrations \cite{yue2024mmmu,liu2024mmbench,liu2024ocrbench,masry2022chartqa,du2024embspatial}. The long-context track assesses synthetic retrieval, realistic long-document comprehension, and multi-turn conversations at context lengths up to 256K \cite{hsieh2024ruler,bai2025longbenchv2,vodrahalli2024michelangelo}.

Our contributions are summarized as follows:

\begin{enumerate}
  \item \textbf{Unified, multi-stage vision-language pre-training.} Text and vision-language data are mixed throughout pre-training under deterministic per-step blending, assigning every data-parallel rank an exact vision-language sample quota in each stage. The context curriculum extends to 256K with square-root sample-length reweighting over logical packed samples.
  \item \textbf{Token-level modality-aware capacity, routing, and memory.} A unified modality label governs modality-specific shared capacity in MMoE, a modality-aware extension of Kimi K3 QB with independent text and vision bias states, and position-selective Engram injection.
  \item \textbf{Infrastructure co-designed with the architecture.} UniRoute fuses routing, selection, and balancing statistics into a single kernel whose metadata the dispatcher consumes directly, eliminating dense routing maps and per-layer synchronization in sparse routing. ModalPack executes modality-specific shared experts entirely on-device. DistEngram shards the conditional-memory table so that lookups stay within a node, while sparse table gradients are collected during the backward pass and communicated only after backbone gradient synchronization.
  \item \textbf{Integrated observability and data attribution.} Interpretable online telemetry is complemented by event-triggered source reconstruction, enabling data and preprocessing remediation across model versions.
  \item \textbf{Interface-controlled base-model evaluation.} We characterize Youtu-VL2.0-32B-A4B-Base across established text protocols, continuation-based multimodal evaluation, and long-context benchmarks under a common base-model interface.
\end{enumerate}

Because the capacity, routing, and memory mechanisms are indexed by modality rather than specialized to vision, additional modalities can be incorporated as extra shared branches, bias states, and memory masks. This report instantiates and evaluates the vision-language case only. All architecture-specific claims and evaluations in this report refer to Youtu-VL2.0-32B-A4B-Base. Larger members of the family share these design principles but are not characterized here.

\section{Related Work}
\label{sec:related}
\subsection{Efficient sequence modeling and sparse capacity}

The Transformer provides direct content-addressable retrieval but incurs quadratic attention cost and a KV cache that grows with sequence length~\cite{vaswani2017attention}. MLA reduces cache storage through latent key-value representations~\cite{deepseekai2024deepseekv2}, while recurrent linear-attention methods maintain fixed-size state. DeltaNet interprets this state as a fast-weight memory updated by the delta rule~\cite{yang2024parallelizing}, and Kimi Linear develops KDA for large-scale language modeling~\cite{kimiteam2025kimilinear}. Hybrid architectures combine recurrent propagation with periodic retrieval: Kimi Linear establishes a 3:1 KDA-MLA pattern, which Kimi K3 and the Ling-3 models adopt with related allocations~\cite{kimiteam2026kimik3,inclusionai2026ling30flash,inclusionai2026ling31flash}. Qwen3.8-Flash-Next instead pairs Gated DeltaNet~\cite{yang2025gated} with full attention that is replaced by sparse attention at continued pre-training~\cite{qiu2026qwen38next}, while GLM-5.3-Flash combines KDA with DeepSeek Sparse Attention (DSA)~\cite{deepseekai2025deepseekv32,zai2026glm53flash}, introduced in DeepSeek-V3.2: the MLA latent cache is retained, but a lightweight indexer scores earlier positions and each query attends to a bounded subset of them. DeepSeek-V4 combines compressed sparse attention with heavily compressed attention for million-token contexts~\cite{deepseekai2026deepseekv4}, and DeepSeek-V4.1-Flash extends this line with cross-layer reuse of sparse selections, low-precision cache storage, and an encoder-decoder split of the global cache~\cite{deepseekai2026deepseekv41flash}.

Sparse MoE provides a complementary form of efficiency by activating only a subset of feed-forward parameters~\cite{shazeer2017outrageously}. GShard establishes large-scale conditional computation with automatic sharding~\cite{lepikhin2021gshard}, and Switch Transformer simplifies sparse routing while documenting its stability constraints~\cite{fedus2022switch}. DeepSeek-V2 combines fine-grained routed experts with shared experts~\cite{deepseekai2024deepseekv2}, and DeepSeek-V3 develops auxiliary-loss-free balancing through correction biases~\cite{deepseekai2024deepseekv3}.

Residual transport is a third capacity axis. Hyper-connections learn how transformer layers read and write multiple residual streams~\cite{zhu2024hyperconnections}, while manifold-constrained hyper-connections regularize this transport~\cite{xie2025mhc}. Qwen3.8-Flash-Next introduces Gated Residual, which spends its expressiveness on an elementwise read and drops inter-stream mixing, and evaluates its effect on training stability~\cite{qiu2026qwen38next}. Youtu-VL2.0-32B-A4B-Base adopts hybrid recurrent-sparse-attention mixing, sparse feed-forward capacity, and four-stream GR transport as its backbone. Its primary architectural distinction lies in how this backbone is extended to native multimodal learning inside one shared decoder.

\subsection{Native multimodal representation learning}

Early large vision-language systems commonly connect separately pre-trained components. Flamingo inserts gated cross-attention between a frozen visual encoder and language model~\cite{alayrac2022flamingo}, whereas LLaVA projects pre-trained visual features into a language model before visual instruction tuning~\cite{liu2023visual}. NaViT introduces variable-resolution patch-and-pack processing~\cite{dehghani2023patch}, and PaliGemma combines a pre-trained encoder and decoder into a compact transferable VLM~\cite{beyer2024paligemma}. Qwen2-VL develops native dynamic resolution and multimodal rotary position encoding~\cite{wang2024qwen2vl}, and Qwen3-VL, whose vision transformer we adopt, extends this line toward perception, spatial reasoning, and agentic interaction~\cite{bai2025qwen3vl}. Our earlier Youtu-VL studies unified supervision across heterogeneous visual tasks~\cite{wei2026youtuvl}.

Recent models increasingly incorporate visual data during base-model training. Kimi K3 is a native multimodal KDA-MLA MoE whose vision encoder is trained from scratch with next-token prediction~\cite{kimiteam2026kimik3}. GLM-5.3-Flash combines native multimodality with recurrent and sparse-attention layers~\cite{zai2026glm53flash}. DeepSeek-V4.1-Flash jointly trains a multimodal MoE with a dedicated vision encoder~\cite{deepseekai2026deepseekv41flash}. Muse Glimmer pairs a dense language model with a perception encoder~\cite{abtahizadeh2026museglimmer}. Youtu-VL2.0-32B-A4B-Base adopts a pre-trained visual encoder rather than training one from scratch, and mixes text and vision-language data throughout pre-training. Its distinction lies in how capacity, routing, and parametric memory are allocated between modalities inside the shared decoder, under the single token-level modality signal described next.

\subsection{Modality-aware conditional computation and memory}

Multimodal MoE routing must reconcile cross-modal transfer with modality-conditional load imbalance. ERNIE~4.5 dedicates routed experts to each modality, routes text and visual tokens in isolation, and balances them with auxiliary losses~\cite{ernie2025technicalreport}. Kimi K3 introduces QB, estimating one expert-wise bias vector from required-bias histograms pooled over the routed-token population~\cite{kimiteam2026kimik3}. Its published formulation is modality agnostic. DeepSeek-V4.1-Flash instead maintains separate text and image correction biases under a sign-based auxiliary-loss-free update~\cite{deepseekai2026deepseekv41flash}. Concurrent work also studies per-modality shared experts in unified multimodal pre-training and reports gains over global shared experts~\cite{tong2026beyond}. In contrast to partitioning routed experts, the MMoE of Youtu-VL 2.0 splits only the shared experts by modality and keeps a single learned-router pool open to both. It extends QB, rather than adding balancing losses, with modality-indexed bias vectors and required-bias histograms, together with a fused execution path for 32B multi-stage pre-training.

Engram introduces conditional memory as a complementary sparsity axis based on scalable $N$-gram lookup~\cite{cheng2026conditional}. Qwen3.8-Flash-Next incorporates an $N$-gram embedding layer~\cite{qiu2026qwen38next}, and DeepSeek-V4.1-Flash adopts Engram within a broader cache-compression architecture~\cite{deepseekai2026deepseekv41flash}. Youtu-VL 2.0 follows the FP8 row-sharded storage and momentum-Sinkhorn table updates of DeepSeek-V4.1-Flash, and adds what multimodal pre-training requires: visual-position masking and boundary-aware state reset.

\subsection{Training stability and operational observability}

Large-model instabilities have been associated with learning rates, activation outliers, gradient growth, and architectural scaling. Wortsman et al. show that selected large-scale instabilities can be reproduced with smaller proxy models~\cite{wortsman2023smallscale}, while Qwen3.8-Flash-Next studies learning-rate stress, residual transport, and Muon optimization~\cite{qiu2026qwen38next}. These works primarily investigate the mechanisms that produce instability.

Distributed multimodal training additionally requires attribution across data composition, modality routing, model components, and systems state. Online losses, expert loads, activation statistics, gradient norms, and memory or throughput measurements diagnose many failure modes directly, but sample-level causes require data provenance, which these aggregate signals do not retain. Youtu-VL 2.0 therefore separates continuous telemetry from event-triggered source reconstruction, supporting both immediate run diagnosis and data remediation in later model versions.

\subsection{Base-model evaluation under interface dependence}

Text base-model evaluation has mature task-specific protocols and standardized harnesses~\cite{eval-harness}. MMLU measures broad academic knowledge~\cite{hendrycks2021mmlu}, GPQA targets graduate-level scientific reasoning~\cite{rein2024gpqa}, GSM8K and BIG-Bench Hard evaluate mathematical and compositional reasoning~\cite{cobbe2021gsm8k,suzgun2023bbh}, and HumanEval, MBPP, MultiPL-E, and RepoBench cover code generation from single functions to whole repositories~\cite{chen2021humaneval,austin2021mbpp,cassano2023multiple,liu2024repobench}.

For multimodal base models, standard evaluation prompts can conflate visual understanding with instruction following. MMMU, MMBench, OCRBench, ChartQA, and EmbSpatial-Bench cover expert visual reasoning, broad perception, OCR, chart understanding, and spatial relations~\cite{yue2024mmmu,liu2024mmbench,liu2024ocrbench,masry2022chartqa,du2024embspatial}, and toolkits such as VLMEvalKit standardize their evaluation~\cite{duan2024vlmevalkit}. These protocols commonly rely on chat-oriented prompts and answer-format instructions. Continuation-based evaluation reduces this confound by presenting demonstrations and a stable completion boundary while retaining only scorer-essential constraints.

\section{Model Architecture}
\label{sec:method}
\subsection{Architecture overview}

\begin{figure}[t]
    \centering
    \includegraphics[width=0.98\textwidth]{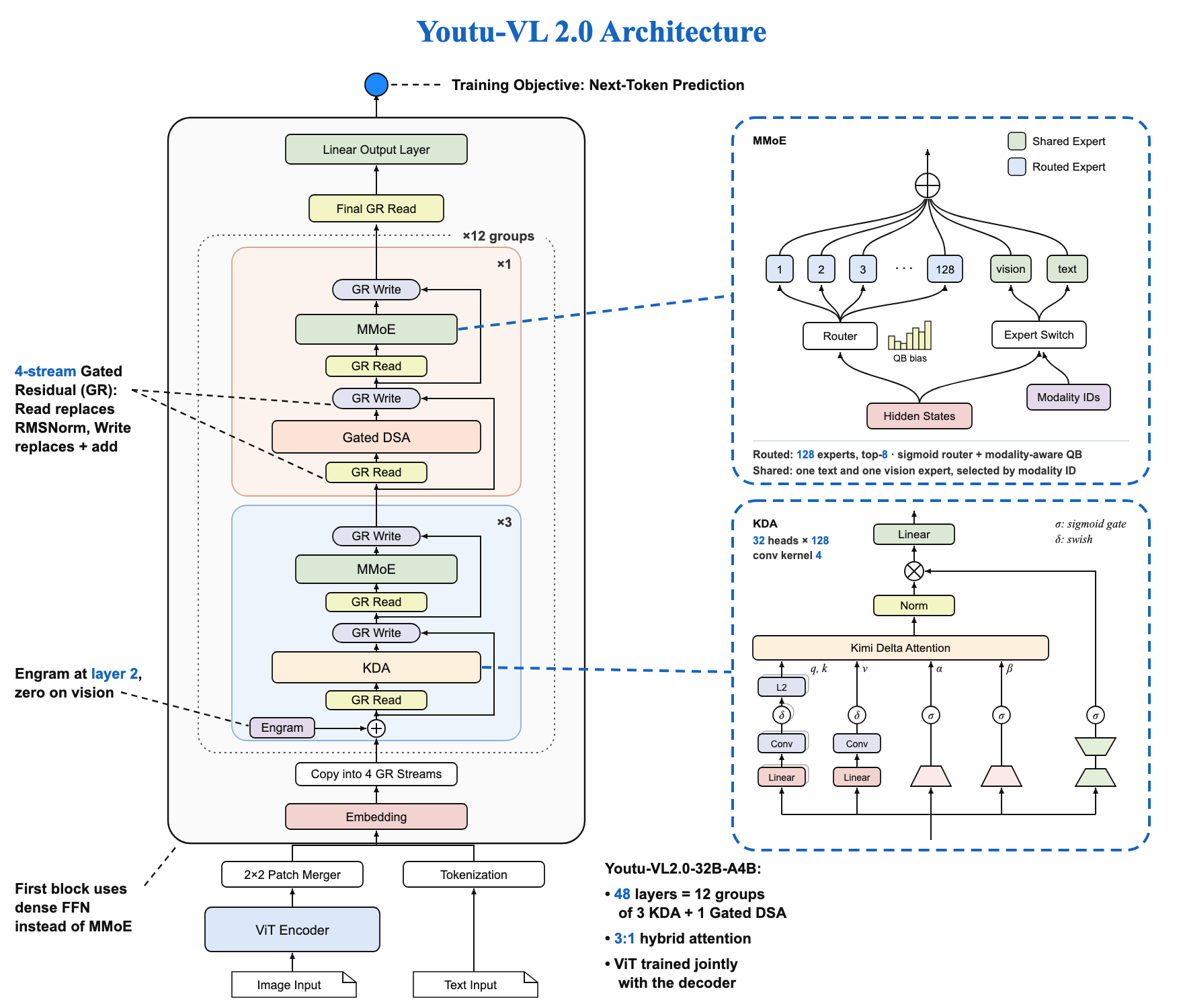}
    \caption{Youtu-VL 2.0 architecture. Left: a Qwen3-VL ViT and patch merger feed the shared decoder, which repeats three KDA layers and one gated DSA layer with four-stream GR and text-selective Engram. Upper right: MMoE combines a common routed pool with modality-switched shared experts and modality-aware QB biases. Lower right: KDA projection, recurrent mixing, normalization, and output-gating paths. The first decoder layer uses a dense FFN instead of MMoE.}
    \label{fig:architecture}
\end{figure}

Youtu-VL 2.0 combines a shared multimodal backbone with token-level specialization. Figure~\ref{fig:architecture} illustrates the complete pipeline from visual and textual inputs to autoregressive prediction. Specifically, a hybrid Mixture-of-Experts (MoE) decoder performs shared contextual modeling, while a single token-level modality label governs three allocation decisions: dedicated capacity, routed capacity, and parametric memory. This design preserves cross-modal interaction throughout the network while allowing selected components to maintain modality-specific parameters and statistics.

For visual inputs, we adopt a pre-trained Qwen3-VL ViT~\cite{bai2025qwen3vl}, which handles spatial positional encoding internally, paired with a $2\times2$ spatial patch merger. The merger groups adjacent $2\times2$ patch features and projects them to the decoder hidden dimension, compressing the visual token sequence fed to the backbone. In the serialized input sequence, these projected features replace visual placeholder embeddings, while ordinary text tokens and textual image-boundary markers retain their standard embeddings. Unlike Qwen3-VL, we extract features exclusively from the final ViT layer without intermediate-layer DeepStack injection. Consequently, visual and textual tokens form a single unified embedding sequence and interact directly within the shared autoregressive decoder, eliminating the need for dedicated cross-attention layers.

Youtu-VL2.0-32B-A4B-Base consists of $L=48$ decoder layers with a hidden dimension of $d=2{,}048$. The decoder holds 31.05B parameters, excluding the input embedding, the untied language-model head, the vision encoder with its merger, and the Engram table. The Engram table adds 8.41B parameters, 8.39B of which are lookup values stored separately in FP8 (Section~\ref{sec:engram}), and is not counted in the 32B total. Its attention backbone repeats a 3:1 pattern of three Kimi Delta Attention (KDA) layers followed by one gated DeepSeek Sparse Attention (DSA) layer. DSA retains the latent key-value compression of Multi-head Latent Attention (MLA) while adding sparse key selection, replacing dense gated MLA during the final 256K-context stage. Across depth, four-stream Gated Residual (GR) connections link the sublayers. For the feed-forward blocks, the first layer employs a dense feed-forward network (FFN), whereas all subsequent layers use Multimodal Shared-expert MoE (MMoE) with 128 routed experts and top-8 routing.

The hybrid attention backbone and four-stream GR connections follow prior designs \cite{kimiteam2025kimilinear,kimiteam2026kimik3,deepseekai2025deepseekv32,qiu2026qwen38next} to support efficient sequence modeling and learned cross-layer transport. Our multimodal architecture implements three modality-conditioned mechanisms on top of this foundation. Specifically, MMoE allocates dedicated shared-expert capacity per modality, modality-aware Quantile Balancing (QB) regulates routing competition within the shared expert pool, and text-selective Engram supplies parametric lexical memory without hashing visual placeholders as discrete text patterns.

\subsection{Hybrid decoder}

Modeling long multimodal sequences requires both efficient history aggregation and direct access to specific earlier tokens. Following the 3:1 allocation in Kimi Linear \cite{kimiteam2025kimilinear}, we interleave three recurrent KDA layers with one gated DSA layer. In most layers, KDA compresses sequence history into a fixed-size recurrent state, whereas every fourth layer uses gated DSA to retain token-indexed representations for direct causal retrieval. Both token mixers operate continuously across the unified text-visual sequence within each logical sample: visual spans do not reset the recurrent state or attention context, while boundaries between independent packed samples remain strictly isolated.

\subsubsection{Kimi Delta Attention}

KDA aggregates sequence history through a recurrent associative state rather than storing a separate key-value pair for every preceding token. Specifically, it refines the scalar decay of Gated DeltaNet~\cite{yang2025gated} into a channel-wise diagonal gate $\operatorname{Diag}(\alpha_t)$, allowing memory retention to be controlled independently per key channel rather than by a single head-shared scalar. As illustrated in the lower-right panel of Figure~\ref{fig:architecture}, linear projections followed by a causal depthwise short convolution produce queries, keys, and values, with queries and keys $\ell_2$-normalized prior to the recurrent update. For a single head, let $S_t\in\mathbb{R}^{d_k\times d_v}$ denote the key-by-value associative state. Its state update and readout are given by
\begin{equation}
S_t=
\left(I-\beta_t k_t k_t^\top\right)
\operatorname{Diag}(\alpha_t)
S_{t-1}
+\beta_t k_t v_t^\top,
\qquad
o_t=\frac{S_t^\top q_t}{\sqrt{d_k}}\in\mathbb R^{d_v}.
\end{equation}

Defining $\widetilde{S}_t=\operatorname{Diag}(\alpha_t)S_{t-1}$ as the decayed state, the update can be rewritten as
\begin{equation}
S_t=
\widetilde S_t
+\beta_t k_t
\left(v_t-\widetilde S_t^\top k_t\right)^\top,
\end{equation}
which makes its delta-rule structure explicit: $\widetilde{S}_t^\top k_t$ predicts the current value from the decayed state, and the update writes the prediction residual along $k_t$ scaled by the sigmoid gate $\beta_t$. Each recurrent layer thus cleanly decouples forgetting, corrective writing, and contextual readout.

The channel-wise decay $\alpha_t$ is computed as the exponential of a bounded log-decay produced by a low-rank projection of the layer input. Before projecting back to the decoder hidden dimension, the recurrent output is RMS-normalized per head and modulated by a channel-wise sigmoid gate. Following Kimi K3, the gate logits are produced by a full-rank projection rather than the low-rank projection used in Kimi Linear~\cite{kimiteam2026kimik3,kimiteam2025kimilinear}. Under a fixed head geometry, neither the recurrent state nor the short-convolution state grows with sequence length. Consequently, KDA provides constant-memory history aggregation, complementing the periodic gated DSA layers that retain uncollapsed token representations for direct retrieval.

\subsubsection{Gated DeepSeek Sparse Attention}
\label{sec:methods_dsa}

The DSA layers complement the recurrent layers with content-addressable retrieval. Following MLA \cite{deepseekai2024deepseekv2}, each source token $r$ is compressed into a joint key-value latent $c_r$ and a head-shared key component $k_r^{\mathrm s}$, which, together with the indexer key introduced below, are the only quantities cached. These layers apply no positional encoding (NoPE), delegating position and recency modeling to the KDA layers and avoiding rotary retuning as context length increases \cite{kimiteam2025kimilinear}. Without a rotary transformation between query and key projections, the key and value up-projections can be absorbed into the query and output projections, respectively. Attention is therefore evaluated directly over the head-shared cached pair:
\begin{equation}
A_{t,r,j}=
\operatorname{softmax}_{r\in\mathcal S_t}
\left(\frac{\tilde q_{t,j}^\top c_r+q_{t,j}^{\mathrm s\top}k_r^{\mathrm s}}{\sqrt{d_{\mathrm{qk}}}}\right),
\qquad
o_{t,j}=\sum_{r\in\mathcal S_t}A_{t,r,j}\,c_r,
\end{equation}
where $\tilde q_{t,j}$ denotes the absorbed content query and $q_{t,j}^{\mathrm s}$ is the shared-key component. A sigmoid gate projected from the decoder hidden dimension modulates each head output as $\bar o_{t,j}=\sigma(a_{t,j})o_{t,j}$. Because this gate is a single scalar per head, it commutes with the absorbed value projection, in contrast to the channel-wise output gate used in the KDA layers.

The attended set $\mathcal S_t$ evolves across curriculum stages. Prior to the 256K stage, $\mathcal S_t$ includes all valid causal positions within the logical sample, so the layer operates as dense gated MLA: the latent cache narrows each entry, though attention complexity still scales with context length. At 256K, we restrict $\mathcal S_t$ to the top-$k_{\mathrm{idx}}$ positions selected by a lightning indexer, with $k_{\mathrm{idx}}=2048$, yielding DeepSeek Sparse Attention (DSA)~\cite{deepseekai2025deepseekv32}:
\begin{equation}
\mathrm{score}_{t,r}=
\sum_{h}w_{t,h}\,
\operatorname{ReLU}\!\left(\iota_{t,h}^\top\kappa_r\right).
\end{equation}
Here, the indexer queries $\iota_{t,h}$ are projected from the same low-rank query latent as the main attention queries, $\kappa_r$ is a single normalized key per position, and the head weights $w_{t,h}$ are projected from the layer input. The indexer thus acts as a single-key, softmax-free scorer rather than a full secondary attention stack. It employs 16 query heads, against 32 in the main attention, and executes in BF16 without the Hadamard rotation used for low-precision scoring in the original DSA design, and carries no rotary channels under NoPE. Because the indexer still scores all preceding causal positions, sparse selection bounds the cost of the main attention rather than of scoring. When the context length is below $k_{\mathrm{idx}}$, the layer remains dense and short-context behavior is preserved.

Following DeepSeek-V3.2~\cite{deepseekai2025deepseekv32}, DSA is introduced during the 256K stage in two phases. First, during a dense warm-up phase of 100 steps, attention remains dense and the backbone is frozen while the randomly initialized indexer is trained with a learning rate of $10^{-3}$ to match the dense attention distribution by minimizing $\operatorname{KL}\big(p_t\,\|\,\operatorname{softmax}_r(\mathrm{score}_{t,r})\big)$ over the causal positions of each logical sample, where $p_t$ is the head-summed and renormalized attention distribution $A_{t,\cdot,j}$. Second, during sparse training, main attention is restricted to $\mathcal S_t$ and the entire model is updated via the cross-entropy loss, while the indexer continues optimizing its KL objective restricted to $\mathcal S_t$. In this second phase, both the indexer inputs and target distribution $p_t$ are detached, and top-$k_{\mathrm{idx}}$ selection passes no gradients, ensuring that the backbone is supervised solely by language modeling and the indexer solely by KL alignment. Deferring sparse selection until the long-context stage follows Qwen3.8-Flash-Next, which similarly transitions from full to sparse attention during continued pre-training~\cite{qiu2026qwen38next}. Together, the decoder integrates bounded recurrent state readout with sparse retrieval across a unified multimodal sequence.

\subsubsection{Residual transport and feed-forward capacity}

While KDA and DSA mix information across tokens, GR~\cite{qiu2026qwen38next} regulates information transport across depth. It maintains four parallel residual streams, each initialized from the input embedding. Given stream states $h_{t,i}$ and their independently normalized representations $\hat h_{t,i}$, a sublayer $F$ computes
\begin{equation}
x_t=\frac{1}{n}\sum_{i=1}^{n}\alpha_{t,i}\odot\hat h_{t,i},
\qquad
y_t=F(x_t),
\qquad
h'_{t,i}=h_{t,i}+\beta_{t,i}y_t,
\quad n=4.
\end{equation}

\begin{figure}[htbp]
    \centering
    \includegraphics[width=1.0\textwidth]{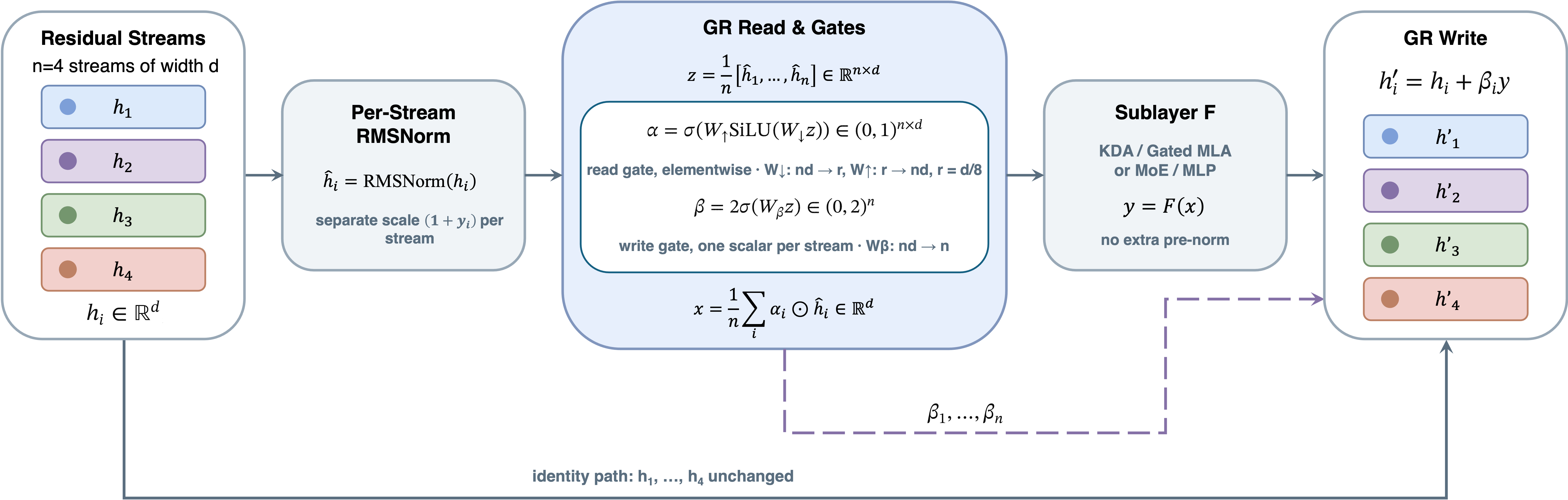}
    \caption{Four-stream Gated Residual. Elementwise read gates aggregate independently normalized streams into the sublayer input, while a per-stream scalar write gate controls residual injection and preserves an identity path across the depth of the decoder.}
    \label{fig:gated_residual}
\end{figure}

Both gates are predicted from the concatenation of all normalized streams: the elementwise read gates $\alpha_{t,i}$ via a low-rank bottleneck, and the scalar per-stream write gates $\beta_{t,i}\in(0,2)$ via a direct linear projection, as shown in Figure~\ref{fig:gated_residual}. These gates are distinct from the KDA gates $\alpha_t$ and $\beta_t$. Consistent with findings in Qwen3.8-Flash-Next~\cite{qiu2026qwen38next}, refining the read gate drives most of the performance gain, whereas channel-wise write gates offer negligible benefit. Since the read step already normalizes and gates its input, GR replaces standard sublayer pre-normalization rather than preceding it, avoiding extra normalization layers across the widened stream. Moreover, no direct stream-to-stream mixing is applied to the residual states, i.e., cross-stream aggregation occurs solely through the gated read, allowing each stream to preserve an identity path while receiving a gated sublayer update. Sequence-mixing and feed-forward sublayers maintain independent GR parameters, and a final learned read merges the four streams into the input of the language-model head.

Sparse feed-forward layers provide conditional channel capacity after sequence mixing, decoupling total stored parameters from per-token compute. Following the dense first layer, each MoE layer combines selected routed experts with a modality-specific shared branch. All experts operate directly at the decoder hidden dimension, omitting the latent bottleneck of LatentMoE~\cite{elango2026latentmoe} used in Kimi K3, and adopt the Sigmoid Tanh Unit GLU activation of Kimi K3 to bound both multiplicative branches~\cite{kimiteam2026kimik3}. Section~\ref{sec:discussion_latentmoe} compares this choice against a latent variant. Section~\ref{sec:modality-aware} details the shared-branch allocation and routed expert selection.

\subsection{Modality-aware capacity, routing, and memory}
\label{sec:modality-aware}

The shared backbone allows cross-modal interaction, while modality-specific structure motivates a more selective allocation of computation. We distinguish three requirements: a dedicated path for features specific to one modality, modality-conditional load control within the routed pool, and lexical memory restricted to discrete text patterns. These requirements are addressed by complementary mechanisms governed by one source-token modality label $m_t\in\{\mathrm{text},\mathrm{vision}\}$. Ordinary text and textual image-boundary markers remain text even within a vision-language sample. Only positions occupied by visual features are visual. Throughout this section a valid token is a non-padding position. Padding carries no modality label and is excluded from all three mechanisms.

\subsubsection{Multimodal Shared-expert MoE}
Shared experts provide an always-active path for every token in an MoE layer, independent of routed selection. In multimodal pre-training, however, forcing a single shared branch to accommodate both textual and visual representations can cause cross-modal interference, as the two modalities differ substantially in feature statistics and optimization dynamics. To prevent this bottleneck while preserving cross-modal parameter sharing, MMoE partitions the shared path into modality-specific branches while retaining a common pool of routed experts, as illustrated in the upper-right panel of Figure~\ref{fig:architecture}. Given the sublayer input $x_t$, the MMoE output is computed as
\begin{equation}
y_t=
\sum_{e\in\mathcal I_t}w_{t,e}F_e(x_t)
+F_{\mathrm{shared},m_t}(x_t),
\qquad m_t\in\{\mathrm{text},\mathrm{vision}\},
\end{equation}
where $\mathcal I_t$ denotes the selected routed experts, $w_{t,e}$ denotes their mixture weights defined in Section~\ref{sec:qb}, and $F_{\mathrm{shared},m_t}$ is the shared branch corresponding to source modality $m_t$. Shared-branch assignment is deterministic: text tokens use the text branch and visual tokens use the vision branch, with no additional output gate or routing bias. By contrast, routed selection remains content-dependent, allowing tokens of either modality to access any expert in the common pool.

The two paths serve complementary roles: dedicated shared branches reserve capacity for modality-specific features, while the larger routed pool enables parameter sharing across modalities. Crucially, this specialization operates on representations already contextualized by the shared token mixers, where text tokens can attend to visual context and visual tokens can attend to preceding text. Modality-specific shared experts therefore preserve cross-modal reasoning without duplicating the sequence-modeling backbone.

Each shared branch has the same intermediate dimension as a single routed expert. Each MoE layer stores two shared branches, and each token activates only the branch of its own modality. With $P_e$ parameters per expert, $L_{\mathrm{MoE}}$ MoE layers, $E$ routed experts, and $M=2$ modality branches, the stored and per-token active expert parameter counts are
\begin{equation}
P_{\mathrm{stored}}=L_{\mathrm{MoE}}(E+M)P_e,
\qquad
P_{\mathrm{active}}=L_{\mathrm{MoE}}(k+1)P_e,
\end{equation}
since every token activates one shared branch alongside its $k$ routed experts. Splitting the shared path thus increases stored model capacity while leaving per-token compute unchanged. Moreover, deterministic per-step blending assigns every data-parallel rank at least one text and one vision-language sample per step, ensuring that both shared branches receive gradients at every optimization step and neither idles even during text-dominant early stages.

\subsubsection{Modality-aware Quantile Balancing}
\label{sec:qb}

A router balanced globally without distinguishing modalities can conceal modality-conditional imbalance, particularly when text and vision contribute vastly different token counts. Since MMoE's shared branches do not regulate competition among routed experts, we extend Quantile Balancing (QB)~\cite{kimiteam2026kimik3} with independent text and vision bias vectors at every MoE layer. At optimization step $u$, given sigmoid router scores $s_{t,e}=\sigma(a_{t,e})$, expert selection and mixture weighting are defined as
\begin{equation}
\mathcal I_t=
\operatorname{TopK}_e\left(s_{t,e}+b_{m_t,e}^{(u)},k\right),
\qquad
w_{t,e}=\lambda\,
\frac{s_{t,e}}{\sum_{j\in\mathcal I_t}s_{t,j}},
\quad e\in\mathcal I_t,
\end{equation}
where $k$ denotes the number of selected experts and $\lambda=2.83$ is the routed scaling factor. Biases are applied solely during expert selection, while mixture weights are computed from the unbiased scores normalized over $\mathcal I_t$. As a result, altering the selected set $\mathcal I_t$ indirectly affects both the mixture weights and the gradients. These balancing biases are updated independently of gradient-based router optimization, without introducing a balancing auxiliary loss.

QB estimates the bias required for each expert to reach the current selection threshold. Letting $\tau_t^{(u)}$ denote the $(k+1)$-th largest adjusted score, the required bias for expert $e$ is $r_{t,e}^{(u)}=\tau_t^{(u)}-s_{t,e}$. Over the set of valid tokens $\mathcal T_m^{(u)}$ belonging to modality $m$ within a global step, the bias vector is updated via
\begin{equation}
\bar b_{m,e}^{(u+1)}
=Q_{k/E}\!\left(\{r_{t,e}^{(u)}:t\in\mathcal T_m^{(u)}\}\right),
\qquad
b_{m,e}^{(u+1)}
=\bar b_{m,e}^{(u+1)}
-\frac{1}{E}\sum_{j=1}^{E}\bar b_{m,j}^{(u+1)},
\end{equation}
where $Q_p$ denotes the empirical $p$-quantile and $E$ is the number of routed experts per layer. Targeting the $k/E$ quantile aligns the selection fraction across all experts, while mean-centering removes any global offset. Text and vision maintain separate empirical distributions against the same target selection fraction. Unlike hard-partitioning routed experts by modality~\cite{ernie2025technicalreport}, this approach regulates modality load while keeping the entire routed pool accessible to both modalities.

In practice, quantiles are estimated from per-modality histograms accumulated across all micro-batches and data-parallel ranks within each global step. Updates take effect only at step $u+1$, ensuring that no batch is routed using biases derived from its own statistics. Because no balancing auxiliary loss is used and the estimated bias vectors remain fixed at inference, this estimator achieves modality-conditioned load control without request-dependent rebalancing.

\subsubsection{Text-selective Engram}
\label{sec:engram}

\begin{figure}[t]
    \centering
    \includegraphics[width=0.6\textwidth]{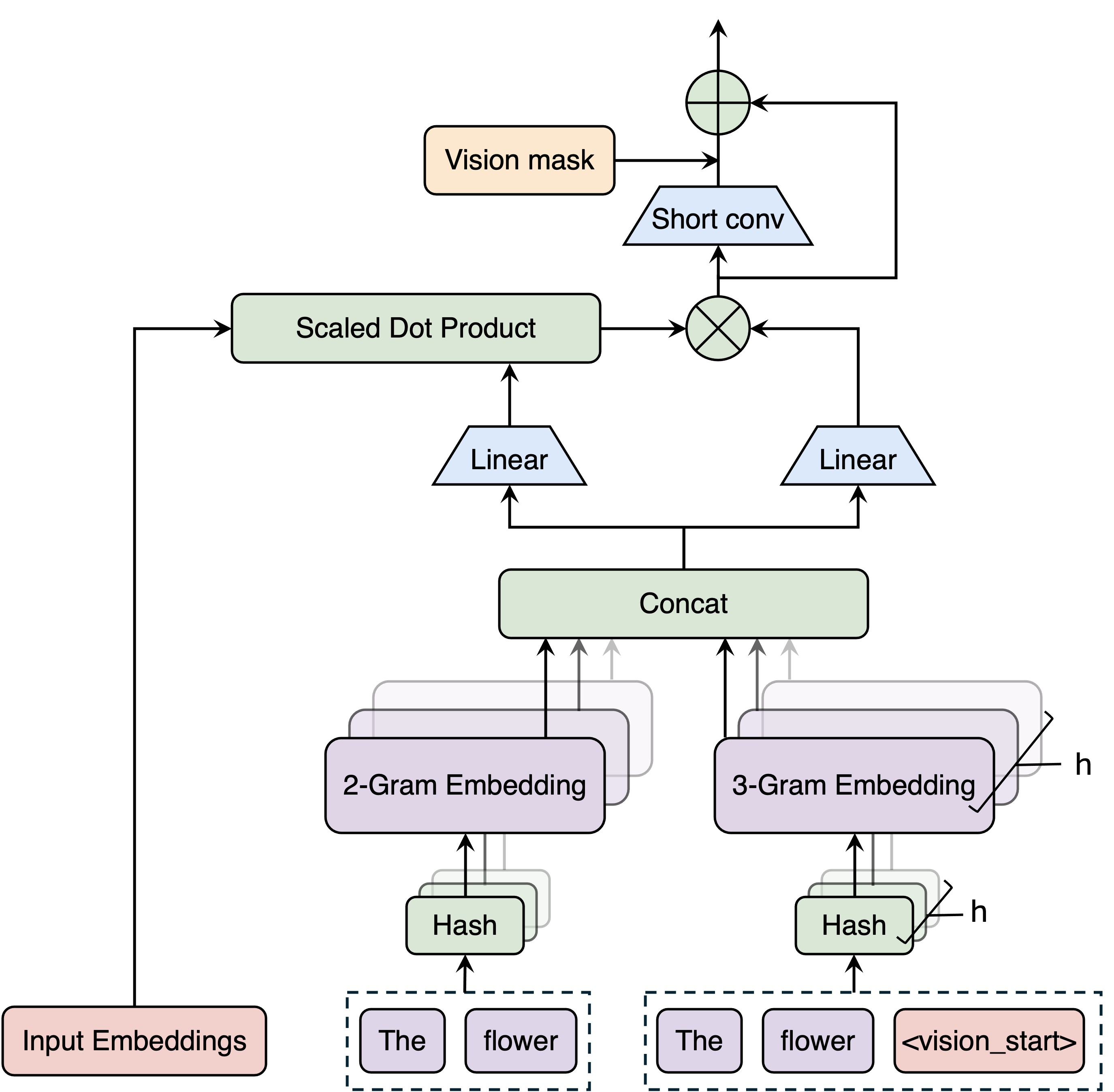}
    \caption{Text-selective Engram. Bigram and trigram lookups are concatenated, projected into keys and values, and gated by the decoder representation. A causal short convolution combines neighboring lookup features, while visual positions receive zero residual and separate the lexical histories on either side of an image. Each stacked layer marked $h$ in the figure is one hash head, indexed by $j$ in the text.}
    \label{fig:engram}
\end{figure}

While expert layers transform contextual representations, Engram~\cite{cheng2026conditional} provides complementary conditional capacity by retrieving embeddings of recurring local token patterns. The decoder adopts a single lookup module at layer 2, immediately ahead of the sequence-mixing GR read, as distributing the same capacity across multiple layers yields no consistent performance improvement \cite{qiu2026qwen38next}. Bigram and trigram lookups form a representation $\psi_t$: given input token IDs $\mathrm{id}_t$, hash head $j$ of order $n$ addresses one row of the embedding table $\mathcal H$ via
\begin{equation}
z_{t,n,j}=
\Delta_{n,j}+
\left[
\left(\bigoplus_{\ell=0}^{n-1}\mu_\ell\,\mathrm{id}_{t-\ell}\right)
\bmod V_{n,j}
\right],
\qquad
\psi_t=\operatorname{Concat}_{n,j}\mathcal H[z_{t,n,j}],
\end{equation}
where $\oplus$ denotes bitwise XOR, $\mu_\ell$ are fixed per-layer multipliers, and each bucket has a distinct prime size $V_{n,j}$ at offset $\Delta_{n,j}$ in $\mathcal H$. Since each head retrieves exactly one row regardless of bucket capacity, total stored capacity scales without increasing per-token lookup cost. Figure~\ref{fig:engram} illustrates this memory path.

Since retrieved entries are static priors that may be unreliable under hash collisions or polysemy, the lookup output is gated by the decoder context rather than injected unconditionally. The query, key, and convolution-input normalization layers occupy independent parameters for each GR stream. Letting $q_{t,i}=\operatorname{Norm}_q(h_{t,i})$ and $k_{t,i}=\operatorname{Norm}_k(W_k\psi_t)_i$, contextual gating and local composition are computed as
\begin{equation}
g_{t,i}=\phi\!\left(\frac{\langle q_{t,i},k_{t,i}\rangle}{\sqrt d}\right),
\qquad
U_t=\operatorname{Concat}_i\left(g_{t,i}W_v\psi_t\right),
\end{equation}
\begin{equation}
R_t=U_t+
\operatorname{SiLU}\!\left(
\operatorname{DWConv}
(\operatorname{Norm}_c(U))_t
\right).
\end{equation}
Here $\phi(s)=\sigma(\operatorname{sign}(s)\sqrt{|s|})$ applies a signed square root before the sigmoid to compress large-magnitude logits and mitigate saturation, whereas the original Engram applies the sigmoid directly~\cite{cheng2026conditional}. The key projection $W_k$ produces one key per GR stream, while the value projection $W_v$ is shared across streams. A causal depthwise convolution then composes neighboring lookup features, with its dilation set to the maximum $N$-gram order so that successive taps read disjoint $N$-grams. Finally, $R_t$ is reshaped and added into the residual streams. In this way, local lexical memory is modulated by the current context, while long-range and cross-modal dependencies remain handled by the hybrid backbone.

Visual features lie outside the discrete lexical space of $N$-gram hashing. Accordingly, visual positions receive zero Engram residual and reset both the lexical and convolutional histories, ensuring that text following an image begins a fresh lexical segment. Importantly, the boundary is strictly local to Engram, while the KDA and DSA layers continue to preserve cross-modal context across the visual span.

The lookup table accounts for most of Engram's parameters, storing roughly 8 billion lookup values. Each row is stored in FP8 as E4M3 values with E8M0 scales per row and feature block, and sharded by row across the data-parallel group. Since lookup indices depend only on the input token sequence, addressing is fully deterministic. Maintaining standard Adam states for a table of this scale would dominate optimizer memory. Following DeepSeek-V4.1-Flash~\cite{deepseekai2026deepseekv41flash}, sparse row gradients are instead routed back to their owning rank and applied via a momentum update with Sinkhorn balancing. Meanwhile, a rank-local FP32 master copy accumulates sub-quantization updates smaller than one E4M3 step.

\section{Pre-training Data}
\label{sec:pretrain_data}
\subsection{Data composition and accounting}
\label{sec:data_composition}

We standardize heterogeneous source formats into a common record and archive records by language and content type. The text ledger groups Chinese, English, and multilingual sources into web pages, encyclopedic and knowledge corpora, code, mathematics, multidisciplinary and examination material, synthesized documents, and reasoning trajectories. The vision-language ledger adds captions, OCR, charts and tables, visual grounding, VQA, GUI, video and multi-image examples, and spatial reasoning. Provenance metadata stays attached to each record, so a corpus family can be traced through filtering, sampling, and the stage curriculum. Four quantities remain distinct throughout this section: raw assets or logical examples, packed sample slots, valid supervised tokens, and source-modality tokens. Every text-to-vision ratio we report is tied to one of these denominators, and a ratio over sample slots is never compared directly with a ratio over tokens.

A rule-based stage removes malformed records and applies domain-specific quality filters. MinHash near-duplicate detection collapses repeated documents before sampling, and a separate decontamination stage screens source corpora for overlap with evaluation items. The retained records are bucketed by token length, with separate ledgers at 8K, 32K, 128K, and 256K, so each curriculum stage draws from a declared length distribution. Each dataset row records its sample count, the token mass it contributes to that bucket, its total token mass, its sampling rate, and its quality-check status. Each stage reads a named data version, and corpus updates are recorded against that version rather than against the training step at which they take effect during pre-training.

\subsection{Text data curation and synthesis}
\label{sec:text_curation}

The text corpus combines inherited data, incremental data, and topic-specific synthesis. Iterative filtering reduces the inherited Youtu-LLM~\cite{lu2025youtullm} text corpus from 10.5T to 8.4T tokens, and lower-difficulty post-training examples are moved into the pre-training pool. Incremental data contributes 4.5T tokens under the processing scheme reported in the same technical report. Table~\ref{tab:text_corpus} gives the resulting composition. The five synthesis programs below account for the 545B topic-specific portion.

\begin{table}[!ht]
\centering
\small
\caption{Composition of the text corpus available to pre-training. Volumes use the fixed reference tokenizer, with T and B denoting trillions and billions of tokens. The inherited row reports the volume after iterative filtering, and the synthesis row sums the five programs described in this section.}
\label{tab:text_corpus}
\begin{tabular}{lr}
\toprule
Source & Volume \\
\midrule
Inherited Youtu-LLM text & 8.4T \\
Incremental data & 4.5T \\
Topic-specific synthesis & 545B \\
\midrule
Total & 13.45T \\
\bottomrule
\end{tabular}
\end{table}

Five programs address stylistic degradation in web documents, sparse long-tail knowledge, Chinese-specific linguistic skills, long expository structure, and state tracking over long sequences. The routes of Section~\ref{sec:vl_synthesis} generate verified question answering and spatial trajectories from vision-language sources. The programs below instead rewrite or extend text-only corpora. Table~\ref{tab:text_corpus} decomposes the text volume implied by the stage totals of Table~\ref{tab:curriculum}.

\paragraph{Role-conditioned refinement of low-quality web text.}
Web pages rejected by quality filtering carry useful content inside noisy or templated surfaces. We sample a persona-style prompt for each rewrite, with a style-free option in the candidate set, and apply it to the raw document. The generator preserves the source content while replacing the surface form, which produces 500B tokens of clean text in the selected voice. We randomize the persona so that repeated sampling does not converge on one register. The style-free option keeps the rewrite distribution anchored to unstyled prose.

\paragraph{Long-tail world knowledge.}
Standard quality filtering discards exactly the low-exposure pages in which rare facts concentrate. We reclaim these pages and represent them as a document graph, following the construction of ElephantBench~\cite{pan2026elephantbench}. Nodes are documents, support edges connect documents about the same fact, and conflict edges mark incompatible accounts of the same subject and attribute. Knowledge-point clustering and entity retrieval narrow the candidate pairs before relation verification, which avoids comparing every pair of documents. Verified conflict edges and their subgraphs yield 220 thousand long-tail records and 3B tokens. Each account is checked against its originating documents and authoritative public sources, then reviewed by human annotators. The graph remains an offline construction and audit structure, so every retained record carries a source chain rather than a free-form assertion. ElephantBench supplies the construction procedure here, not a benchmark to train on.

\paragraph{Chinese-specific question synthesis.}
General web corpora underrepresent character-level and phonological reasoning in Chinese. We collect questions from internet forum puzzles, pinyin, error correction, and the interpretation of radicals and character components, then ask several strong models to answer each question independently. Anonymous voting resolves conflicting candidates, and human auditors inspect a sample of the resolved set, giving 10 million records and 30B tokens. Voting establishes agreement between models, not correctness of the reasoning behind an answer, so the human sample is what checks the process itself.

\paragraph{Long-form English textbooks.}
A predefined taxonomy of knowledge points generates retrieval queries over the source corpus. Retrieved chunks are clustered into topical units, a hierarchical table of contents is planned and filtered by a quality gate, and source-grounded sections are written under audience and style briefs before assembly into complete books. We follow the pipeline of~\cite{tao2026textbooks} for 140 thousand books and 6B tokens. That work holds generated content, token count, and retrieval pool fixed across its controls and still finds complete books superior to independent sections, length-matched concatenations, and per-document rewrites. The gain comes from document-level organization rather than rewriting style, which is why we retain book boundaries instead of dispersing the sections into packed sequences.

\paragraph{Structured state-tracking tasks.}
Two tasks exercise explicit state maintenance over long inputs, following the KDA warm-up setting~\cite{kimiteam2025kimilinear}. Palindrome generation asks for a natural article reproduced in reverse order under an instruction. Stack state tracking disperses an article into a stack, applies a random sequence of pops, and asks for the resulting stack state. We retain 40 thousand records and 6B tokens. Both tasks admit deterministic verification from the input alone, which makes them suitable as a warm-up for the long-context stages where recurrent state must persist across many tokens.

\subsection{Multimodal data synthesis and verification}
\label{sec:vl_synthesis}

The multimodal corpus available to pre-training totals 1.7T tokens and
combines four families: interleaved image-text web documents,
vision-language question-answering pairs, trajectory data, and caption and
CV-class data. The families play complementary roles. Interleaved documents
supply long-range image-text context, QA pairs supply dense supervision per
image, trajectories supply sequential grounding, and the caption pool supplies
the single-image alignment signal. Table~\ref{tab:mmcomp} gives the resulting
composition. Volumes use the fixed reference tokenizer, with image tokens
counted at the per-image mean $\bar v = 437.9$ of the vision token budget, so
the totals are budgeting figures rather than packed training counts. The
latter are reported at the stage level.

\begin{table}[!ht]
\centering
\small
\caption{Composition of the multimodal corpus available to pre-training.
Volumes use the fixed reference tokenizer, with T and B denoting trillions
and billions of tokens. The interleaved row reports the volume after the
filter cascade of this section. The caption row collects single-image
supervision and is the remainder of the total.}
\label{tab:mmcomp}
\begin{tabular}{lr}
\toprule
Source & Volume \\
\midrule
Interleaved image-text & 500B \\
Vision-language QA         & 400B \\
Trajectory data             & 150B \\
Caption and CV-class data   & 650B \\
\midrule
Total                       & 1.7T \\
\bottomrule
\end{tabular}
\end{table}

The other three families differ in how quality is established, so each
receives a dedicated pipeline. Interleaved
web documents arrive pre-paired but structurally unreliable, and their
treatment is therefore \emph{subtractive}: a five-stage cascade
(Section~\ref{sec:cleaning}) decides which documents and images survive.
QA pairs and trajectories, by contrast, are \emph{constructed}: the corpus
does not merely select existing samples but synthesizes new ones, with
verification effort scaled to reasoning depth. Direct QA follows a
sketch-free route in which generation is routed by answer type and accepted
only after answer-level verification (Section~\ref{sec:sketch_free}), while
multi-step spatial tasks follow a sketch-assisted route
(Section~\ref{sec:sketch_assisted}) that emits per-step records and subjects
them to rendering, cross-representation consistency, and intervention checks.

\subsubsection{Cleaning interleaved image-text data}
\label{sec:cleaning}

The interleaved corpus is harvested from three broad source families: general
web pages, digitized books, and scientific documents such as arXiv papers.
Web pages contribute the bulk of the volume and the greatest diversity of
image-text layouts, but also the least reliable pairing. Digitized books
supply long, coherent narratives with figures anchored in surrounding prose.
Scientific documents contribute densely referenced figures whose captions
and in-text citations bind images tightly to technical content. The three
families share a common downstream format, so they enter the same processing
belt. They differ sharply in structural reliability, which the cleaning cascade
of this section is designed to absorb. Interleaved image-text web data is abundant but structurally unreliable:
image references outlive the files they point to, pages are padded with
boilerplate, and document quality spans orders of magnitude. Rather than one
heavyweight filter, we run five sequential stages $S_1$ to $S_5$ over every
record, governed by three principles.
\emph{Light rules first}: stages are ordered by unit cost, so regex and
heuristic checks run before any neural scorer, and the most expensive model
sees the smallest possible input. \emph{Explicit granularity}: a rule is
charged either to the whole document or to a single image, never ambiguously
to both. $S_1$, $S_2$, and $S_5$ drop documents, while $S_3$ and $S_4$ drop
individual images. \emph{Logged removals}: every drop records its reason, so
re-tuning a threshold is an offline recomputation over score files, not a
re-run of GPU models. The cascade decides membership in the corpus, not how
often a document is used. Reweighting happens at the mixing stage.

\paragraph{$S_1$ and $S_2$: low-cost filters on text alone.}
Two CPU stages run before any image is touched. $S_1$ flags sensitive content
such as NSFW or violent material and removes the whole document. Exemptions
are built in for documentation placeholders so that legitimate technical
content survives. $S_2$ targets the two most common failure modes of web
crawls at the text level: pages dominated by \emph{boilerplate} (navigation,
cookie banners, template text) are dropped when the boilerplate ratio exceeds
$0.30$, and \emph{near-duplicates} of already-kept documents are dropped as
well. Both stages cost little compute, so they run first and shrink the
stream before the expensive stages see it.

\paragraph{$S_3$: validity, with two granularities.}
$S_3$ inspects every referenced image: it must decode, have a short side of at
least $256$\,px, and an aspect ratio of at most $20$. Geometry failures are
charged to the single image, which is removed while its document stays on the
belt. A record whose images cannot be decoded at all is a different case: the
page that was crawled is no longer reproducible from what is on disk, so the
failure is charged to the whole document. Keeping these two outcomes separate
is what keeps image-level statistics attributable after the fact.

\paragraph{$S_4$: the relevance test.}
$S_4$ is the only rule that asks whether an image and its text actually belong
together, and the only one whose failure removes an image while leaving the
document intact. We do not score the image against the whole page: the topic
of a long document drifts, and a globally computed similarity would retain
decorative images that merely echo the page topic. Instead we build a
\emph{local} context $c_i$ from the characters immediately surrounding the
image ($\pm$350 characters) and compute $s_i = \cos(E_v(v_i), E_t(c_i))$ with
a dual-tower vision-language encoder.

The difficulty is that the absolute value of $s_i$ is not comparable across
images: on correctly paired samples a large fraction of images score near zero
and a non-trivial fraction score negative, even though image-to-text retrieval
on the same encoder remains high. The ranking signal is strong while the
absolute scale is unstable, so any global cosine threshold is simultaneously
too strict for some images and too lax for others. We therefore calibrate each
image against its own null distribution: 256 random texts are drawn from the
corpus and scored against $v_i$, giving $(\mu_i, \sigma_i)$, and the image is
removed when
\begin{equation}
  z_i \;=\; \frac{s_i-\mu_i}{\sigma_i} \;<\; \tau_z = 0 .
\end{equation}
The rule is portable: \emph{keep the image if and only if its own context is more similar
to it than a random text is}. The pool size trades the stability of
$(\mu_i,\sigma_i)$ against encoder cost, and a few hundred negatives suffice
for the statistic to settle in practice.

\paragraph{$S_5$: document quality, the last and most expensive stage.}
The rebuilt document is scored by a compact quality model, and one property is
easy to overlook: the scoring head has a single output unit, i.e.\ it is a
\emph{regression} head emitting a continuous score on roughly $[0, 1.2]$, not
a calibrated probability. The operating threshold must be obtained by
calibration on held-out data. We use $\tau_q = 0.44$ rather than inheriting
the $0.5$ prior of a probabilistic head.
Documents with more than 16 images or more than 8{,}000 estimated tokens
exceed an explicit skip budget. They bypass scoring entirely and are handled
by an explicit policy, so that a skipped document is never mistaken for a
rejected one.

\paragraph{Rebuilding and the document-level verdict.}
After $S_4$, surviving images are re-serialized so that text spans and image
placeholders remain aligned. Two verdicts follow
directly from per-image decisions: a record that hit undecodable assets and a
document left without any image are both dropped. The latter has degenerated
into a text-only sample and no longer belongs in an interleaved corpus. The
remaining documents face $S_5$'s threshold. For budgeting we estimate the
training cost of each document as
$\hat T(\mathrm{doc}) = n_{\mathrm{tok}}(t_{\mathrm{doc}}) + |I_{\mathrm{doc}}|\cdot \bar v$, where $n_{\mathrm{tok}}$ counts the text tokens $t_{\mathrm{doc}}$ and $I_{\mathrm{doc}}$ is the image set. The per-image mean $\bar v = 437.9$ is measured by
reading the true resolution of 20{,}000 sampled images under the encoder's
resize rule. The sampled distribution is heavy-tailed, so we use the sample
mean rather than the median, which would understate the budget by nearly
$2\times$. $\hat T$ is a \emph{budgeting} estimate used only for
length-bin assignment. True training token counts are reported from the packed
artifacts.

\begin{figure}[!ht]
	\centering
	\includegraphics[width=\textwidth]{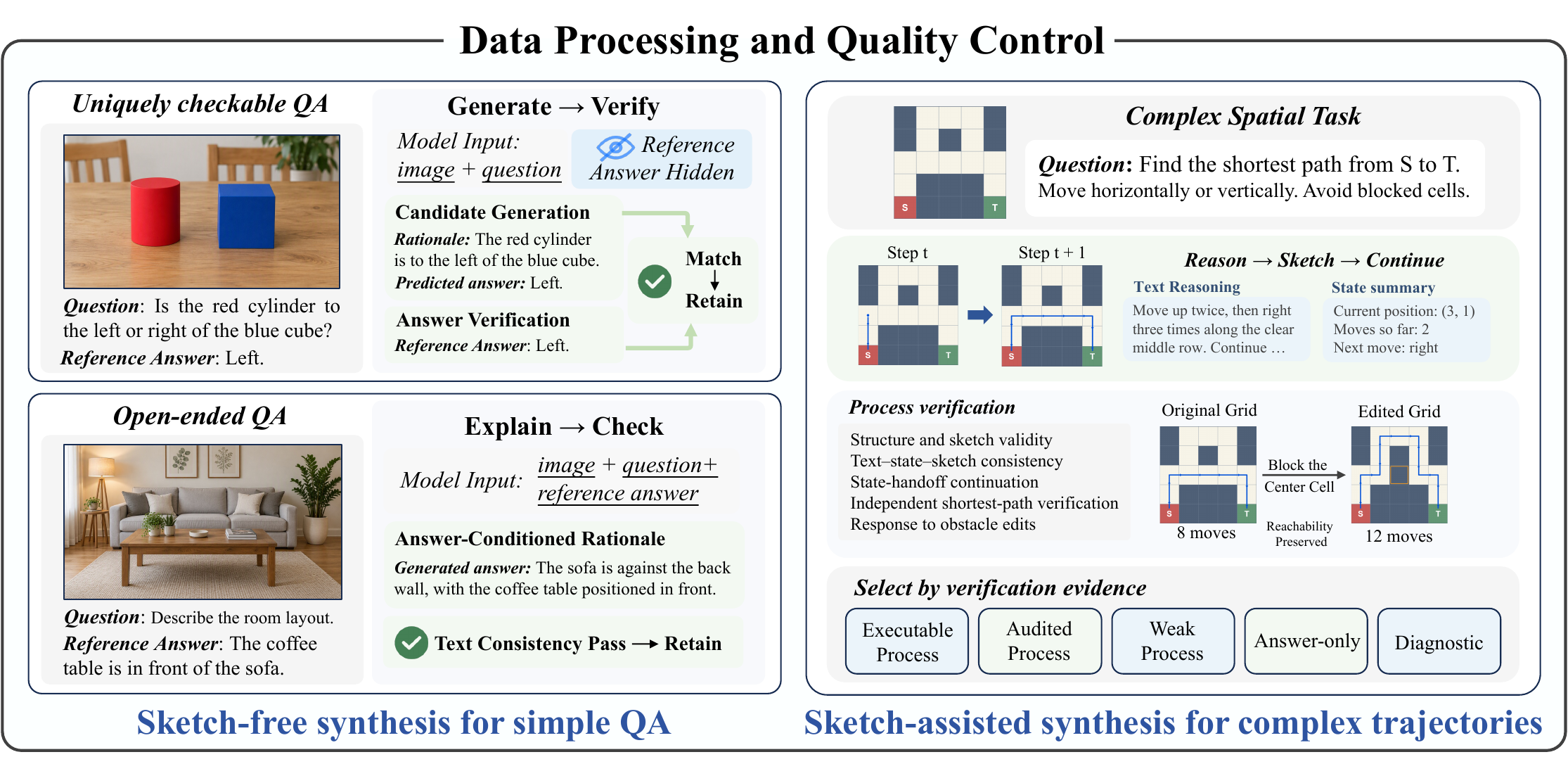}
	\caption{Two complementary routes for data synthesis and quality control. Left: sketch-free QA uses reference-hidden generation with answer verification or reference-conditioned generation with consistency checks. Right: sketch-assisted trajectories combine iterative reasoning and rendering with process verification and evidence-based selection.}
	\label{fig:data_processing}
\end{figure}

\subsubsection{Sketch-free synthesis for simple QA}
\label{sec:sketch_free}

We normalize samples as $x=(I,q,a^{*},\zeta)$, comprising image references, a question, an optional reference answer, and preserved source metadata such as grids, coordinates, and constraints. Reasoning depth and available verification evidence determine two synthesis routes: sketch-free reasoning for direct QA (Section~\ref{sec:sketch_free}) and sketch-assisted trajectories for multi-step spatial reasoning (Section~\ref{sec:sketch_assisted}).

\paragraph{Answer-type-routed generation.}
For annotated QA, we generate concise textual reasoning $r$ and a predicted answer $\hat a$. Let $\omega\in\{\mathrm{u},\mathrm{o}\}$ denote uniquely checkable and open-ended answers, respectively. The reference answer is withheld for the former and supplied for the latter:
\begin{equation}
\label{eq:sketchfree_gen}
(r,\hat a)\sim p_{\theta}\bigl(r,\hat a \mid I,q,a^{*}_{\omega}\bigr),
\qquad
a^{*}_{\omega}=
\begin{cases}
\emptyset, & \omega=\mathrm{u},\\[2pt]
a^{*}, & \omega=\mathrm{o}.
\end{cases}
\end{equation}
Here, $p_{\theta}$ is the generator. This supports rejection sampling for uniquely checkable answers and answer-conditioned rationalization for open-ended responses.

\paragraph{Answer verification.}
Task-specific matchers $\mathcal{M}_{\omega}$ and threshold $\eta_{\omega}$ determine acceptance:
\begin{equation}
\label{eq:sketchfree_verif}
\mathcal{V}_{\omega}(\hat a,a^{*})=\mathbb{1}\Bigl[\max_{\mu\in\mathcal{M}_{\omega}}\mu(\hat a,a^{*})\ge \eta_{\omega}\Bigr],
\qquad
\mathcal{D}_{\omega}=\bigl\{(I,q,a^{*},r)\ :\ \mathcal{V}_{\omega}(\hat a,a^{*})=1\bigr\}.
\end{equation}
For uniquely checkable answers, matchers test normalized equality, extracted choices, yes/no polarity, numerical equivalence, or constrained phrase containment. Open-ended answers use normalized containment, normalized edit similarity, or token recall. Acceptance establishes answer-level agreement only. Neither branch certifies intermediate reasoning. Parsing or verification failures trigger resampling within a fixed budget $B$, without exposing $a^{*}$ for $\omega=\mathrm{u}$. Rejected candidates retain diagnostic labels. Timeouts and rate limits are recorded separately as infrastructure failures and excluded from quality accounting.

\subsubsection{Sketch-assisted synthesis for complex trajectories}
\label{sec:sketch_assisted}

For multi-step spatial tasks, the pipeline assembles a trajectory of incremental records:
\begin{equation}
\label{eq:step_record}
\mathcal R_t=\bigl(r_t,\ s_t,\ d_t,\ z_t,\ v_t,\ c_t\bigr),
\qquad
\xi=(\mathcal R_1,\dots,\mathcal R_T).
\end{equation}
Here, $T$ is the trajectory length, $r_t$ captures incremental spatial reasoning, $s_t$ records entities, coordinates, relations, constraints, and derived values, and $d_t$ tracks dependencies for identifying the effects of local edits. The optional sketch $z_t$ encodes spatial state in an editable domain-specific language (DSL). $v_t$ is its deterministic rendering, and $c_t$ stores structural and consistency checks. The model generates $(r_t,s_t,d_t,z_t)$, while rendering and verification supply $(v_t,c_t)$. Each round receives the original input and prior reasoning, states, and rendered sketches. A separate \texttt{finish} flag and final answer terminate generation.

\paragraph{Format checks and sketch rendering.}
Validators check record completeness, parsability, artifact references, and final-answer extraction. For steps containing sketches, rendering proceeds deterministically only after syntax and task-specific geometric checks:
\begin{equation}
\label{eq:render}
v_t=
\begin{cases}
g(z_t), & \operatorname{Valid}(z_t)=1,\\[2pt]
\bot, & \text{otherwise},
\end{cases}
\end{equation}
where $g$ is the renderer and $\bot$ denotes validation failure. Online rendering and offline completion use identical validation rules.

\paragraph{Consistency across text, states, and sketches.}
For shared, identifiable spatial quantities, representation-specific mappings $\phi$ recover canonical values from text, structured state, symbolic sketches, and images:
\begin{equation}
\label{eq:consistency}
\phi_{\mathrm{txt}}(r_t)\;\equiv\;\phi_{\mathrm{st}}(s_t)\;\equiv\;\phi_{\mathrm{sk}}(z_t)\;\equiv\;\phi_{\mathrm{img}}(v_t).
\end{equation}
Conflicts are flagged, and insufficient evidence is marked uncertain rather than accepted.

\paragraph{Measuring the contribution of sketches.}
At the last nonterminal step containing a sketch, we neutralize answer cues and test continuation under five conditions: text only, sketch only, text plus sketch, neutral text plus sketch, and an externally verified state. The text-only and combined conditions exclude explicit state summaries, measuring the incremental contribution of sketches:
\begin{equation}
\label{eq:sketch_gain}
\Delta_{\mathrm{sk}}=\operatorname{Acc}(c_{\mathrm{both}})-\operatorname{Acc}(c_{\mathrm{txt}}),
\end{equation}
where $c_{\mathrm{txt}}$ and $c_{\mathrm{both}}$ denote text-only and combined histories, and $\operatorname{Acc}$ measures continuation accuracy.

\paragraph{Answer verification and edit testing.}
For executable tasks, an independent executor computes reference outcomes $y=f_{\mathrm{exec}}(q,\zeta)$ from the question and preserved source state. For baseline $C_0$, a relevant $d_t$-guided edit $C_1$, and an irrelevant edit $C_2$, outputs $(y_0,y_1,y_2)$ define expected changes and invariants over fields $k$:

\begin{equation}
\label{eq:intervention}
\mathcal{S}_{\mathrm{chg}}=\{k: y_1[k]\neq y_0[k]\},
\qquad
\mathcal{S}_{\mathrm{hold}}=\{k: y_0[k]=y_1[k]=y_2[k]\}.
\end{equation}
The model predicts changed and invariant fields without intervention labels or expected outcomes. We score these predictions using propagation F1 and invariant F1, check target invariance under $C_2$, and interpret intervention scores only for correct baselines. Verification evidence determines assignment to executable-process, audited-process, weak-process, answer-only, or diagnostic pools. Executable-process supervision requires passing execution and intervention checks. Natural-image tasks without executable ground truth instead rely on source-image evidence, independent model audits, and human review. Human review may resolve uncertainty or downgrade samples but cannot replace executable ground truth. We retain intermediate artifacts and distinguish source annotations from generated content for traceability.

\subsection{Multimodal serialization and packing}
\label{sec:serialization}

Visual and textual inputs are serialized into one autoregressive continuation sequence. Visual features from the Qwen3-VL vision transformer and the $2\times2$ patch merger replace placeholder embeddings at image positions, while ordinary text and textual image-boundary markers keep their token embeddings. Each position carries a source-modality label in $\{\mathrm{text},\mathrm{vision}\}$, and a supervised-token mask determines which positions contribute to the loss. Image spans, multi-image examples, end-of-document markers, and coordinate representations are declared in the serialization format rather than inferred at runtime. Vision-language task formats are converted to the same continuation objective, with no chat-style instruction template of any kind.

Packing concatenates independent examples up to the sequence length of the current stage. Sample boundaries are recorded as \texttt{\_sample\_lengths} together with position resets and attention metadata. The recurrent state resets at a logical sample boundary, whereas a visual span inside a sample does not reset context. This keeps independent examples isolated while leaving image and text within one context. Malformed images and extreme aspect ratios are handled before serialization so that visual-token counts stay within the budget of the current stage.

\subsection{Joint text-vision curriculum and deterministic blending}
\label{sec:curriculum}

Text and vision-language data are mixed from the first pre-training step, so cross-modal representations form inside the shared decoder rather than in a later alignment stage. The visual encoder is initialized from Qwen3-VL and held frozen while the decoder adapts to its features, after which joint training follows at a reduced learning rate.

Table~\ref{tab:curriculum} lists the five stages of Youtu-VL2.0-32B-A4B-Base and the volume trained in each stage. Stages 1 and 2 operate at 4K context, after which the curriculum moves to 32K, 128K, and 256K.

\begin{table}[!ht]
\centering
\small
\caption{Stage curriculum of Youtu-VL2.0-32B-A4B-Base. Volumes use the fixed reference tokenizer, with T and B denoting trillions and billions of tokens. MM share is the multimodal token fraction of each stage.}
\label{tab:curriculum}

\begin{tabular}{ccrrr}
\toprule
Stage & Context & MM share & Trained \\
\midrule
1 & 4K   & 10.0\%   & 9.7T \\
2 & 4K   & 15.0\% & 3.3T \\
3 & 32K  & 12.5\%   & 1.2T \\
4 & 128K & 12.5\%   & 650B  \\
5 & 256K & 12.5\%  & 350B  \\
\midrule
\multicolumn{2}{c}{All stages} & & 15.2T \\
\bottomrule
\end{tabular}
\end{table}

Each stage declares a target modality mixture.  Deterministic per-step blending converts the stage target into an exact vision-language sample quota per data-parallel rank, and a Bresenham placement schedule distributes that integer quota across the step. Integer quantization error is recorded separately from the realized modality ratio, which is measured on source-modality tokens. Controlled sample composition and realized visual-token exposure are therefore distinct quantities.

Deterministic resume, reshuffling, and dataset-exhaustion handling keep the realized mixture fixed across restarts. Corpus additions, removals, reweighting, and quality refreshes are versioned per stage and kept separate from context-length and optimizer changes, so a change in the data version is not confused with a change in the curriculum.

\subsection{Long-context data and loss normalization}
\label{sec:long_context}

Long-context training draws on naturally long documents and multimodal examples, packed independent examples, and synthetically composed long-range tasks where used. Only the first category retains coherence across the full sequence, and packed examples stay independent by construction. Long documents are cleaned, structurally validated, and sampled by length bucket. The ledger keeps separate token counts at 8K, 32K, 128K, and 256K. The context curriculum extends progressively to 256K tokens, and reporting is organized by length bucket rather than by the maximum sequence length alone.

Loss normalization is defined over logical packed examples. For sample $i$ with $n_i$ valid supervised tokens, we write the loss as
\begin{equation}
\label{eq:loss_weight}
\mathcal{L}=\sum_i \frac{n_i^{\alpha}}{\sum_j n_j^{\alpha}}\,\frac{1}{n_i}\sum_{t=1}^{n_i}\ell_{i,t},
\qquad \alpha\in\{1,\tfrac{1}{2},0\}.
\end{equation}
Shorter-context stages use $\alpha=1$, which averages the loss over valid tokens. The 256K stage uses $\alpha=1/2$, so a logical example contributes in proportion to the square root of its valid-token count. The weighting limits domination by exceptionally long samples without giving short examples equal weight. The $\alpha=0$ case corresponds to uniform sample weighting. Section~\ref{sec:discussion_loss_reweight} compares against the $\alpha=1$ token average. We report valid examples and tokens by length bucket, rather than only the maximum context length.

\section{Infrastructure}
\label{sec:infrastructure}
Youtu-VL 2.0 places three demanding requirements on the training stack simultaneously: a decoder that interleaves a recurrent token mixer with periodic sparse attention, a feed-forward path that is both sparsely routed and modality-dependent, and a large conditional-memory table. Left unoptimized, each component can introduce substantial host synchronization, communication, and activation-memory overhead. Therefore, we co-design the training stack around three principles: eliminating redundant computation, batching communication, and overlapping independent operations. Specifically, fused kernels paired with a continuous context-parallel partition keep the recurrent formulation efficient both within and across devices. \textbf{UniRoute} fuses routing, top-$k$ selection, and balancing statistics into a single kernel whose metadata feeds the dispatcher directly, eliminating dense routing maps and per-layer collectives for balancing statistics. \textbf{ModalPack} executes modality-specific shared experts entirely on-device without a host round trip. \textbf{DistEngram} uses node-aligned lookup groups to confine lookup traffic within each node, while accumulating sparse row gradients locally during the backward pass and deferring their communication until after backbone gradient synchronization. The same design principles govern system observability: telemetry relies on batched collectives and buffered writes, and sample-level attribution logs source indices rather than reconstructing samples online, keeping routine telemetry off the critical path.

\subsection{Algorithm-system co-design for the hybrid decoder}
\label{sec:infra_runtime}

\subsubsection{Recurrent-layer kernels}

The serial state update of the recurrent token mixer conflicts with the wide, uniform parallelism favored by GPUs, concentrating runtime overhead in two places.

The first bottleneck is intermediate data movement across the causal short convolution, query-key normalization, and chunked recurrence. A naive implementation writes and reloads intermediate activations between these stages. We fuse all three operations into a single custom autograd function. During the forward pass, the convolution kernel directly emits $\ell_2$-normalized queries and keys alongside their inverse $\ell_2$ normalization factors in one pass. During the backward pass, the normalization gradient is applied on the fly, keeping query and key gradients in full precision until the normalization backward consumes them. The standalone convolution forward achieves bitwise agreement with the reference implementation, whereas convolution gradients exhibit minor numerical differences due to floating-point operation reordering. Packed logical samples remain strictly isolated at both levels: recurrent states never cross micro-batch boundaries, and a sequence-identity mask on the convolution taps prevents any sample from reading tokens from its predecessor.

The second bottleneck lies in the backward kernel that jointly computes query, key, and decay gradients. Its upstream tiling is tuned for narrower heads: it caps key and value tiles at 64 and reduces occupancy to four warps when the head dimension exceeds 64. Consequently, a head of twice that width ($d_k=128$) must traverse both inner loops twice, even though shared memory is large enough to stage the entire tile. We retile this kernel to match our production head geometry and double the warp count. This optimization accelerates the dominant backward kernel by roughly $1.5\times$ across 2K-16K packed tokens while keeping all returned gradients within $10^{-7}$ relative error of the default tiling.

\subsubsection{Parallelizing the hybrid across devices}

Standard causal attention is typically parallelized along the sequence dimension using a mirrored two-chunk layout to balance the triangular workload induced by causal masking. However, a recurrent token mixer cannot adopt this layout since its hidden state must propagate sequentially along the token order. Therefore, we assign each token mixer a distinct context-parallel partition. The recurrent layers adopt a continuous partition in which each rank owns one contiguous sequence shard. Short-convolution halos are exchanged across shard boundaries, and recurrent boundary states are propagated whenever a logical sample spans multiple ranks. This preserves recurrence across devices while keeping distinct samples strictly isolated. By contrast, the attention layers redistribute the query rows of long logical samples using a per-sample zigzag layout, which balances the causal-attention workload across ranks without altering the set of keys each query attends to.

In the sparse attention layers, we further optimize memory and compute during activation checkpointing. The absorbed formulation in Section~\ref{sec:methods_dsa} reduces attention to a multi-query operation over the shared latent, and top-$k_{\mathrm{idx}}$ selection runs without gradients. Since the initial forward pass under activation checkpointing also executes without gradients, we cache the selected indices during the first forward pass and reuse them during backward recomputation. This avoids a redundant full-history scoring and selection pass while guaranteeing that recomputation attends to the exact same positions. Indexer scores on the selected support are still recomputed to evaluate the indexer's KL loss.

\subsection{Infrastructure for sparse multimodal pre-training}
\label{sec:infra_training}

While sparse activation scales stored expert capacity without proportionally increasing per-token compute, it still leaves full parameter storage intact and introduces dynamic execution bottlenecks. In our sparse multimodal stack, we address three primary sources of overhead using kernel fusion, communication scheduling, and activation recomputation: host synchronization induced by data-dependent expert batch shapes, all-to-all dispatch and combine communication, and the elevated activation memory of the four-stream residual path.

\subsubsection{UniRoute: routing, dispatch, and expert execution}

A conventional router materializes a dense token-to-expert routing map, forcing the dispatcher to recompute top-$k$ selection to recover indices the router already computed. Both sources of redundancy can be eliminated. We therefore propose \emph{UniRoute}, which consolidates routing into a single fused kernel: it computes sigmoid router scores, adds the bias corresponding to each token's modality, and executes the top-$(k{+}1)$ selection required by the quantile update in Section~\ref{sec:qb}. Within the same pass, the kernel normalizes the unbiased scores and accumulates both the required-bias histograms and expert load statistics. The dispatcher consumes UniRoute's output indices and probabilities directly, ensuring that the dense routing map is never materialized and expert selection is never repeated.

This design also eliminates fine-grained synchronization overhead. Because the required-bias histograms consist of integer counts, they are strictly additive: we accumulate them locally across micro-batches and recover the global per-modality distribution via a single all-reduce per optimizer step, stacking all MoE layers into one combined message. Consequently, no per-layer collective or per-micro-batch quantile computation is needed, and the routing diagnostics described in Section~\ref{sec:infra_observability} are obtained directly from UniRoute rather than via a separate pass over router scores.

Finally, routed experts are evaluated using a grouped matrix multiplication that fuses the gate and up projections, the bounded SiTU-GLU activation, and the router weights into a single kernel, reproducing reference numerical rounding boundaries exactly. Expert weight gradients are scheduled on a dedicated CUDA stream, allowing the weight-gradient computation to overlap with both the activation-gradient chain and dispatch transfers without altering gradient accumulation order.

\subsubsection{ModalPack: modality-partitioned shared experts}

Two naive approaches to executing modality-specific shared experts are both inefficient: evaluating both branches for every token and masking out the unused output wastes half of the compute, whereas gathering tokens per modality via host-side indexing incurs a host synchronization at every layer. To avoid both bottlenecks, we propose \emph{ModalPack}. First, two lightweight kernels count and partition tokens by modality directly on the device, generating a permutation index without host synchronization or data-dependent memory allocation. ModalPack then gathers the input activations into a packed buffer where each modality segment is padded to a tile boundary, guaranteeing aligned tensor-core inputs and valid weight-gradient computation. Next, each modality's matrix multiplication is launched separately and reads its weights in place via a tensor descriptor, eliminating stacked parameter copies across layers. Finally, the activation function is fused into the epilogue of the first projection, while the second projection scatters outputs directly back into the original token order within its epilogue, avoiding a standalone reordering pass.

\subsubsection{Activation memory, optimizer sharding, and precision}

Four residual streams multiply the activation memory of every sublayer input, so we place activation recomputation (replay) only where it is exact and free of approximation. Specifically, stream states are replayed at an interval equal to the square root of half the pipeline stage's layer count, reducing resident stream states from two per layer to the layer count divided by that interval plus twice the interval. This replay is exact because each stream write is a simple addition whose operands are already retained. Similarly, expert computation is replayed while dispatch transfers remain outside the checkpoint. On the pipeline stage that also hosts the visual encoder and the conditional-memory layer, we narrow expert replay further to token permutation and activation by retaining the first projection's pre-activation. The final pipeline stage is excluded from both replay schemes and instead retains recurrent chunk states, as its activations are consumed immediately by the backward pass and any recomputation would sit directly on the critical path.

Muon approximately orthogonalizes update matrices formed from momentum rather than the weights themselves, making a full update matrix the natural unit of work. We shard the optimizer accordingly: each parameter is assigned to a single owner rank via size-sorted round-robin across the data-parallel group, and each rank orthogonalizes only the updates it owns. This local orthogonalization requires no additional collectives beyond standard gradient and parameter synchronization. Same-shape updates are stacked into batched matrix multiplications, while packed expert parameters carry per-expert shape metadata so that each expert's update matrix is orthogonalized individually. For projections whose rows concatenate attention heads, such as the fused query-key-value projection in recurrent layers and the query and key-value up-projections in attention layers, we apply head-wise Muon to the corresponding update matrices.

The backbone uses BF16 arithmetic alongside FP32 master weights, gradient accumulation, gradient reduction, and router scores. Using BF16 enables tensor-core execution without FP8 per-tensor scaling, avoiding scale bookkeeping inside recurrent kernels and residual stream transport. Conversely, we use FP8 in the conditional-memory table strictly to reduce storage footprint rather than to accelerate backbone compute, as detailed in Section~\ref{sec:infra_memory}.

Long training runs inevitably require multiple restarts, necessitating exact restoration of both training state and data position. Each checkpoint stores the sharded optimizer and scheduler states together with the memory-table shard, its momentum, and its FP32 master copy. Node-local table replicas are rebuilt upon loading, and routing biases travel as persistent model buffers. The data iterator maintains no internal state of its own and is reconstructed directly from the consumed-sample count, removing sampler state from the checkpoint entirely. Across restarts of the training run, overlapping steps executed twice exhibited close numerical agreement, with a median relative loss difference of roughly $2\times10^{-4}$, rather than strict bitwise reproducibility of the full training state.

\subsection{DistEngram: distributed conditional memory}
\label{sec:infra_memory}

The Engram table adds substantial stored capacity while requiring only a small, fixed amount of lookup computation per token. However, replicating the full table on every device is memory-prohibitive, and a conventional dense-gradient embedding layer would materialize a table-sized gradient tensor during the backward pass. We therefore propose \emph{DistEngram}, which exploits three properties of the lookup: addressing is deterministic, each token touches a bounded number of rows, and each optimization step accesses only a small fraction of the table.

DistEngram shards the master table by row across the data-parallel group while maintaining a read-only lookup replica distributed over a dedicated lookup group. When the lookup group aligns with a physical node, all lookup traffic remains intra-node. After each optimizer update, the replica is marked stale. The first subsequent lookup refreshes each rank's replica slice (both quantized values and scales) via an all-gather across the corresponding replica group, which may span nodes. Subsequent lookups within the same step reuse this refreshed replica until the next update. During the backward pass, a custom autograd function records touched row identifiers and accumulates sparse row gradients locally, avoiding dense gradient materialization and deferring table-gradient communication. Once backbone gradient synchronization completes, DistEngram routes these sparse gradients to their owning ranks in two hops, first within the lookup group and then across replicas. To bound peak outgoing communication per round, the initial sender payload is partitioned into rounds with a fixed byte budget, though receiver buffer demand still scales with the aggregate payload from all senders. Finally, the table's squared gradient norm is folded into the global gradient norm so that a single clipping coefficient governs both backbone weights and memory-table rows.

Low-precision table storage introduces a subtle numerical hazard. Under block-scaled FP8 quantization, reading from the quantized representation at every step silently discards updates smaller than a single quantization step, which is the common case for rows accessed only a few times in a large batch. We therefore maintain an FP32 master copy of each shard on its owning rank, apply the Momentum $+$ Sinkhorn update to the master copy, and requantize only the rows touched in that step, allowing small repeated updates to accumulate rather than vanish. This master copy is retained strictly for training: during checkpoint export, the quantized values and scales are copied directly without intermediate dequantization, ensuring that the table read during serving is bit-identical to that read in the training forward pass.

\subsection{Observability and data attribution}
\label{sec:infra_observability}

Monitoring a sparse multimodal training run requires reading state distributed across modalities, experts, and pipeline stages, where a naive implementation turns every reader into a synchronization barrier. At every step, we log per-modality losses normalized by supervised tokens, the global gradient norm, per-expert loads, and maximum load violations computed separately for text and visual tokens. At a coarser interval, we additionally record per-layer routing statistics, shared-expert activations, encoder and decoder gradient norms, and memory-table update statistics. To minimize critical-path synchronization, we apply three optimizations. First, we buffer scalar writes and flush them after the step's collectives so that logging calls never block peer ranks. Second, we aggregate routing statistics across all layers in a single batched collective followed by one device-to-host copy. Third, we compute parameter-norm reductions periodically rather than at every step. Although these reductions, host copies, and buffered writes still incur minor overhead, they keep routine telemetry off the critical path.

While aggregate telemetry can localize an anomaly to a specific component, it cannot identify the underlying training samples. We therefore support sample-level inspection without interrupting training. Specifically, a monitor tracks per-modality losses against an exponential moving average and flags any step exhibiting a large deviation, a loss doubling, or a non-finite value. On a flagged step, we identify the micro-batch index with the highest supervised-token-normalized loss across data-parallel ranks using the metric that triggered the alert. Each writing rank then logs the dataset, shard, and row indices of its local micro-batch at that index, along with the configuration required to reconstruct it. No raw tokens, labels, or pixels are stored during training, and images are referenced solely via row indices. An offline tool subsequently materializes the referenced rows for inspection. While these records facilitate root-cause diagnosis, they do not by themselves establish the definitive cause of a loss spike.

\subsection{Inference latency}
\label{sec:infra_inference}

\begin{figure}[t]
    \centering
    \includegraphics[width=0.9\textwidth]{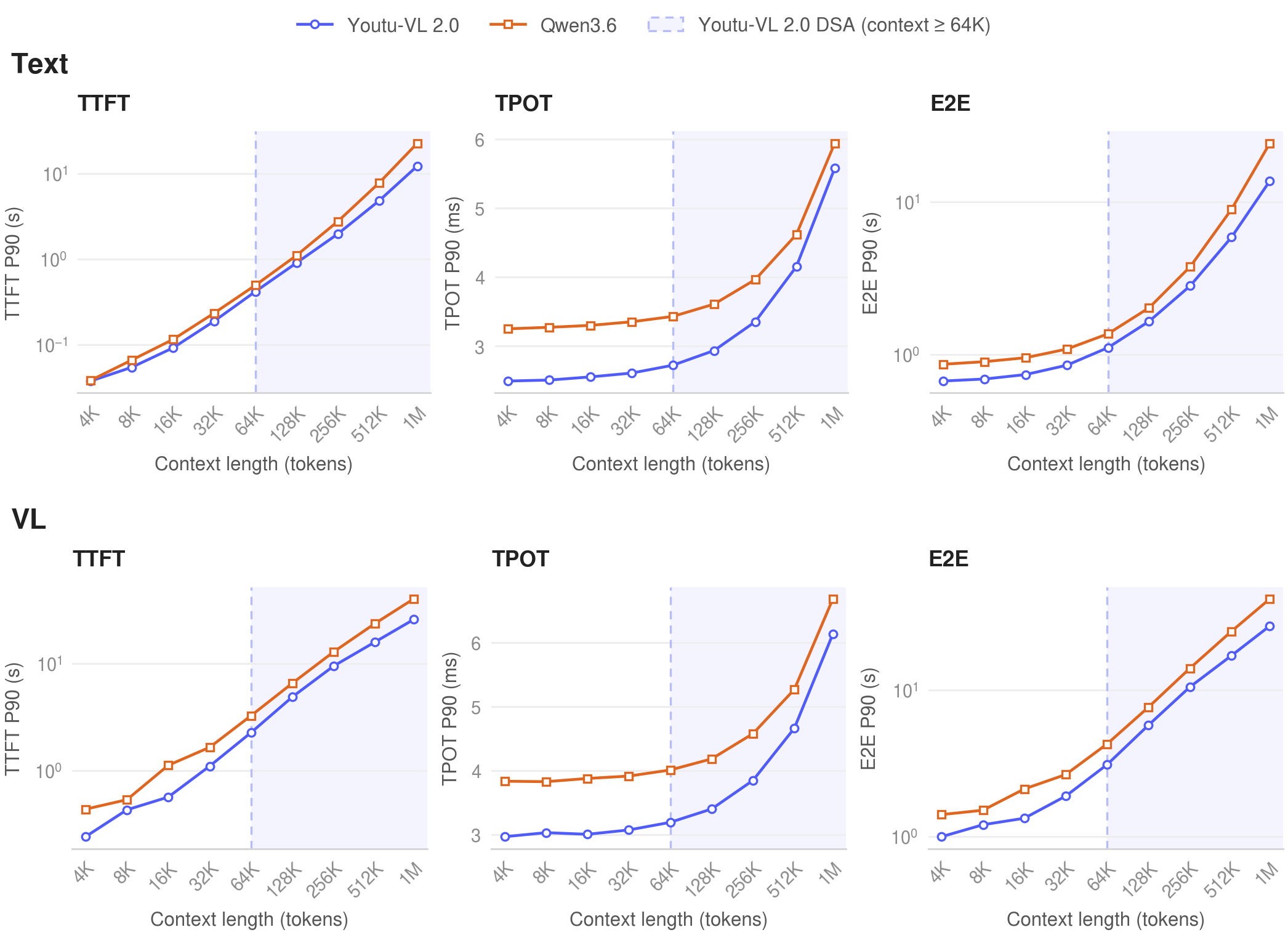}
    \caption{P90 inference latency of Youtu-VL2.0-32B-A4B-Base and Qwen3.6-35B-A3B across context lengths from 4K to 1M tokens. The top and bottom rows report results for text-only and vision-language inputs, respectively. Columns show TTFT (in seconds), TPOT (in milliseconds), and E2E latency (in seconds), where lower values are better. TTFT and E2E are plotted on logarithmic vertical axes. The shaded region denotes contexts of 64K tokens or longer, where sparse DSA is enabled for Youtu-VL2.0-32B-A4B-Base.}
    \label{fig:inference_latency}
\end{figure}

We evaluate inference latency to characterize the serving performance of the trained model. Figure~\ref{fig:inference_latency} compares Youtu-VL2.0-32B-A4B-Base against Qwen3.6-35B-A3B across text-only and vision-language (VL) inputs at context lengths ranging from 4K to 1M tokens. We report the 90th percentile (P90) of time to first token (TTFT), time per output token (TPOT), and end-to-end request latency (E2E). Specifically, TTFT captures the prefill delay before the first generated token, TPOT measures per-token decoding latency thereafter, and E2E records total request completion time. Note that these metrics measure single-request latency rather than aggregate serving throughput.

Both models are deployed using SGLang on a single eight-GPU node in BF16, with tensor and expert parallelism both set to 8. For contexts exceeding its native 256K window, Qwen3.6-35B-A3B employs YaRN scaling with a factor of 4. Requests are issued sequentially (concurrency of 1), with each request generating exactly 256 tokens via greedy decoding. Text-only inputs consist of PG-19~\cite{raecompressive2019} continuations truncated to the exact target context length. VL inputs combine image-text samples from MMLongBench~\cite{wang2025mmlongbench} and MM-NIAH~\cite{wang2024needle}, padded with PG-19 text to match the target length using identical images across both models. Prefix and preprocessing caches are disabled so that every request performs a full prefill. For each configuration, we execute 3 warm-up requests followed by 10 measured requests and report the P90 latency. TTFT and E2E are measured on the client side, and TPOT is computed as $(\mathrm{E2E}-\mathrm{TTFT})/255$.

On text-only inputs, Youtu-VL2.0-32B-A4B-Base achieves comparable TTFT at 4K and lower TTFT across all longer context lengths, while maintaining lower TPOT and E2E latency throughout the entire tested range. At 4K, P90 TPOT is approximately 2.5\,ms for Youtu-VL2.0-32B-A4B-Base compared with 3.3\,ms for Qwen3.6-35B-A3B. This TPOT advantage persists across longer contexts before narrowing toward 1M tokens, where both curves steepen. In the VL setting, Youtu-VL2.0-32B-A4B-Base outperforms Qwen3.6-35B-A3B across TTFT, TPOT, and E2E latency at every tested length, achieving a P90 TPOT of roughly 3.0\,ms versus 3.8\,ms at 4K. Across both models, VL TTFT curves sit above their text-only counterparts, reflecting the additional visual encoding and prefill overhead of multimodal inputs.

The shaded region indicates a serving-time configuration rather than a pre-training stage boundary: sparse DSA is activated starting at 64K tokens in this benchmark, whereas during pre-training it is introduced in the 256K stage (Section~\ref{sec:methods_dsa}). Youtu-VL2.0-32B-A4B-Base maintains lower latency within the shaded region for both modalities. We note that this end-to-end comparison does not isolate the standalone effect of DSA, as the two models also differ in overall architecture and serving kernels. While KDA maintains a fixed-size recurrent state and DSA bounds the number of positions read by the main attention, the indexer still scores the full causal history and historical cache entries remain stored. Consequently, neither total inference latency nor KV cache memory stays strictly constant with context length.

Finally, the 512K and 1M measurements characterize runtime scaling beyond the 256K window used during pre-training and capability evaluation, and therefore do not establish downstream long-context accuracy at those lengths. Likewise, the latency advantages reported here apply specifically to the evaluated single-request workloads and should not be generalized to aggregate throughput under arbitrary concurrency.

\section{Evaluations}
\label{sec:exp}
We evaluate the pre-trained Youtu-VL 2.0 base model along three axes: text
capabilities (Section~\ref{sec:eval-text}), multimodal capabilities
(Section~\ref{sec:eval-mm}), and long-context modeling (Section~\ref{sec:eval-longctx}).
Throughout this section we focus on \emph{base model} evaluation, i.e., all
models are evaluated without instruction tuning, chat templates, or reasoning
modes. Since most existing multimodal benchmarks are designed for
instruction-tuned models, we adapt them to the base-model setting. All
baselines are evaluated with our own evaluation pipeline under the same
settings as Youtu-VL 2.0.

\subsection{Text capabilities}
\label{sec:eval-text}

For text capabilities, we compare Youtu-VL 2.0 with recent open-source base models of comparable or
larger scale:
Qwen3.5-35B-A3B-Base~\citep{qwen2026qwen35}, DeepSeek-V4-Pro-Base~\citep{deepseekai2026deepseekv4}, DeepSeek-V4-Flash-Base~\citep{deepseekai2026deepseekv4},
Kimi-K2-Base~\citep{kimi2025k2}, MiMo-V2.5-Base~\citep{xiaomi2026mimov25}, and
MiMo-V2.5-Pro-Base~\citep{xiaomi2026mimov25pro}.
We report benchmarks implemented based on LM Evaluation
Harness~\citep{eval-harness} spanning the following three dimensions:
\begin{itemize}
  \item \textbf{World knowledge}: MMLU-Pro~\citep{wang2024mmlupro},
        MMLU-Redux~\citep{gema2025mmluredux},
        MMMLU~\citep{openai2024mmmlu},
        MMLU~\citep{hendrycks2021mmlu},
        SimpleQA-Verified~\citep{haas2025simpleqaverified},
        an in-house version of SimpleQA~\citep{wei2024simpleqa},
        INCLUDE~\citep{romanou2025include},
        AGIEval~\citep{zhong2023agieval},
        SuperGPQA~\citep{du2025supergpqa}, and
        GPQA~\citep{rein2024gpqa}.
  \item \textbf{Math and reasoning}: GSM8K~\citep{cobbe2021gsm8k},
        MGSM~\citep{shi2023mgsm},
        MATH~\citep{hendrycks2021math}, and BigBench-Hard
        (BBH)~\citep{suzgun2023bbh} with chain-of-thought demonstrations.
  \item \textbf{Code}: MBPP+ and HumanEval+~\citep{liu2023evalplus,austin2021mbpp,chen2021humaneval},
        LiveCodeBench v6~\citep{jain2024livecodebench},
        CRUXEval~\citep{gu2024cruxeval},
        MultiPL-E~\citep{cassano2023multiple}, and
        RepoBench~\citep{liu2024repobench}.
\end{itemize}
We use two scoring protocols. Standard multiple-choice knowledge
benchmarks, such as MMLU, MMMLU, INCLUDE, SuperGPQA, GPQA, and the
multiple-choice portion of AGIEval, are scored by ranking the given options according to token-level log-likelihood, without free-form decoding. All remaining text benchmarks are evaluated by greedy generation followed by rule-based answer extraction. The evaluation context window is 4K tokens,
including the few-shot demonstrations. The number of in-context examples for
each benchmark is listed in Table~\ref{tab:eval-text}. The average is computed over
all listed benchmarks with equal weights.

\begin{table*}[t]
\centering
\caption{Comparison of Youtu-VL 2.0 with open-source base models on text
benchmarks.
Best scores are in \textbf{bold} and second-best scores are \underline{underlined}.
SimpleQA-V / SimpleQA-IH denote SimpleQA-Verified and our in-house SimpleQA.
LCB v6 denotes LiveCodeBench v6.}
\label{tab:eval-text}
\setlength{\tabcolsep}{2.8pt}
\setlength{\aboverulesep}{0pt}
\setlength{\belowrulesep}{0pt}
\renewcommand{\arraystretch}{1.12}
\newcommand{\thead}[1]{\textbf{\begin{tabular}[t]{@{}c@{}}#1\end{tabular}}}
\resizebox{\textwidth}{!}{%
\begin{tabular}{@{}llc cccccc >{\columncolor{youtuhl}}c@{}}
\toprule
\textbf{Task} & \textbf{Benchmark} & \textbf{Shots}
 & \thead{Qwen3.5} & \thead{DeepSeek\\V4-Flash} & \thead{MiMo-V2.5} & \thead{Kimi-K2} & \thead{DeepSeek\\V4-Pro} & \thead{MiMo-V2.5\\Pro}
 & \thead{Youtu-VL 2.0} \\
\midrule
\multicolumn{2}{@{}l}{\textcolor{gray}{Size}} & - & \textcolor{gray}{35B-A3B} & \textcolor{gray}{284B-A13B} & \textcolor{gray}{310B-A15B} & \textcolor{gray}{1T-A32B} & \textcolor{gray}{1.6T-A49B} & \textcolor{gray}{1.02T-A42B} & \textcolor{gray}{32B-A4B} \\
\midrule
\multirow{10}{*}{\begin{tabular}[c]{@{}l@{}}World\\Knowl.\end{tabular}}
 & MMLU-Pro     & 5 & 65.01 & 65.43 & 66.26 & \underline{69.41} & 67.11 & 43.42 & \textbf{72.06} \\
 & MMLU-Redux   & 5 & 87.90 & \underline{92.20} & 91.31 & 90.07 & \textbf{93.21} & 72.29 & 89.36 \\
 & MMMLU        & 5 & 74.54 & \underline{84.26} & 79.09 & 81.26 & \textbf{86.48} & 83.77 & 77.16 \\
 & MMLU         & 5 & 82.83 & 88.67 & 86.22 & 87.99 & \textbf{90.06} & \underline{89.07} & 84.93 \\
 & SimpleQA-V   & 3 & 14.80 & 19.10 & 12.10 & \underline{23.40} & \textbf{34.10} & 18.30 & 10.80 \\
 & SimpleQA-IH  & 3 & 21.00 & \underline{40.00} & 33.75 & 36.50 & \textbf{45.25} & 37.50 & 25.25 \\
 & INCLUDE      & 5 & 76.29 & \underline{83.24} & 77.70 & 79.42 & \textbf{85.52} & 81.28 & 74.99 \\
 & AGIEval      & 0 & 65.06 & 71.53 & 70.45 & \textbf{76.55} & \underline{72.46} & 69.09 & 60.30 \\
 & SuperGPQA    & 5 & 42.66 & 46.65 & 42.67 & 47.40 & \textbf{54.20} & \underline{48.65} & 34.58 \\
 & GPQA         & 5 & 47.10 & \textbf{55.58} & 47.99 & 47.10 & \underline{55.36} & 53.57 & 47.32 \\
\cmidrule(l{0pt}r{0pt}){1-10}
\multirow{4}{*}{\begin{tabular}[c]{@{}l@{}}Math \&\\Reasoning\end{tabular}}
 & GSM8K        & 8 & 86.13 & \underline{92.42} & 91.74 & 92.04 & 91.96 & \textbf{97.35} & 86.50 \\
 & MGSM         & 8 & 81.60 & \textbf{82.70} & 73.70 & 77.40 & 79.90 & 81.00 & \underline{82.30} \\
 & MATH         & 4 & 52.18 & \underline{59.98} & 53.98 & 56.04 & \textbf{63.22} & 55.90 & 56.32 \\
 & BBH          & 3 & 86.85 & 88.08 & \textbf{88.82} & 88.16 & \underline{88.36} & \textbf{88.82} & 84.99 \\
\cmidrule(l{0pt}r{0pt}){1-10}
\multirow{6}{*}{Code}
 & MBPP+        & 3 & 80.95 & 79.10 & 83.07 & \textbf{89.95} & 84.13 & 87.57 & \underline{88.10} \\
 & HumanEval+   & 0 & 54.88 & 32.32 & 26.83 & \underline{64.63} & 61.00 & 26.83 & \textbf{71.34} \\
 & LCB v6       & 3 & 11.43 &  8.57 & 10.29 & \textbf{15.00} & 12.00 & 10.86 & \underline{14.86} \\
 & CRUXEval     & 1 & 71.94 & 70.25 & 72.63 & \textbf{77.81} & 69.88 & 69.13 & \underline{77.13} \\
 & MultiPL-E    & 0 & \underline{66.64} & 30.39 & 56.41 & 61.25 & 63.41 & 43.44 & \textbf{70.00} \\
 & RepoBench    & 3 & 28.42 & 32.27 & \textbf{35.57} & \underline{34.70} & 32.72 & 34.10 & 30.98 \\
\midrule
\multicolumn{2}{@{}l}{\textbf{Average}} & - & 59.91 & 61.14 & 60.03 & \underline{64.80} & \textbf{66.52} & 59.60 & 61.96 \\
\bottomrule
\end{tabular}%
}
\end{table*}

Across the 20 text benchmarks in Table~\ref{tab:eval-text}, Youtu-VL 2.0 achieves
an average score of 61.96 with 32B total and 4B active parameters, ranking
third overall. It outperforms Qwen3.5-35B-A3B at a comparable scale and also
surpasses the 300B-class models DeepSeek-V4-Flash and MiMo-V2.5, trailing only
the trillion-parameter Kimi-K2 and DeepSeek-V4-Pro.
Coding is the clearest strength of the model. Youtu-VL 2.0 attains the highest
code average across the six code benchmarks, 58.73 against 57.22 for Kimi-K2.
The advantage is most pronounced in zero-shot code
generation, where it reaches 71.34 on HumanEval+ and 70.00 on MultiPL-E, well
ahead of the remaining base models. On math and reasoning, Youtu-VL 2.0 outperforms
Qwen3.5 and MiMo-V2.5 overall, ranking second on MGSM and third on MATH.

World knowledge shows a clear split between reasoning-heavy benchmarks and pure
factual recall. Youtu-VL 2.0 ranks first overall on MMLU-Pro with 72.06 and
outperforms Qwen3.5 across the MMLU suite and GPQA. By contrast, on closed-book
long-tail factual and specialized benchmarks such as SimpleQA, SuperGPQA, and
AGIEval, performance is bounded by the 32B parameter capacity and falls behind
the 284B to 1.6T models. Finally, under our unified evaluation
protocol, MiMo-V2.5-Pro occasionally fails to follow the few-shot output format
on some generative tasks, including MMLU-Pro, MMLU-Redux, HumanEval+, MultiPL-E, and
several RULER subsets, which lowers its rule-extracted scores on those
benchmarks.

\subsection{Multimodal capabilities}
\label{sec:eval-mm}

Our multimodal evaluation is built on VLMEvalKit~\citep{duan2024vlmevalkit}.
Most existing multimodal benchmarks and their official prompts target instruction-tuned models. They rely on chat templates and natural-language instructions to elicit answers in the expected format, which base models cannot follow reliably. We therefore convert every benchmark into a few-shot format. For each benchmark, we reorganize and clean the prompts into a unified
question-answer layout, normalize the answer format, and prepend in-context demonstrations so that the expected output format is specified by examples rather than by instructions. The multimodal benchmarks we use can be classified into the following groups:
\begin{itemize}
  \item \textbf{STEM and math}: AI2D (test)~\citep{kembhavi2016ai2d},
        LogicVista~\citep{xiao2024logicvista},
        WeMath~\citep{qiao2025wemath},
        MMMU (val)~\citep{yue2024mmmu},
        MMMU-Pro (10-option setting)~\citep{yue2025mmmupro}, and
        MathVista (testmini)~\citep{lu2024mathvista}.
  \item \textbf{General VQA}: MMBench (v1.1 dev)~\citep{liu2024mmbench},
        SimpleVQA~\citep{cheng2025simplevqa},
        ChineseSimpleVQA~\citep{gu2025chinesesimplevqa},
        WorldVQA~\citep{zhou2026worldvqa},
        BabyVision~\citep{chen2026babyvision},
        PerceptionBench~\citep{lin2026perceptionbench},
        RealWorldQA~\citep{xai2024realworldqa}, and
        MMStar~\citep{chen2024mmstar}.
  \item \textbf{Document and chart understanding}:
        ChartQA (test)~\citep{masry2022chartqa},
        CharXiv reasoning (RQ)~\citep{wang2024charxiv}, and
        OCRBench v2 (EN/CN)~\citep{fu2024ocrbenchv2}.
  \item \textbf{Hallucination}: CRPE existence and relation~\citep{wang2024allseeingv2} and
        HallusionBench~\citep{guan2024hallusionbench}.
  \item \textbf{Grounding and spatial understanding}:
        EmbSpatial-Bench~\citep{du2024embspatial},
        CountBenchQA~\citep{beyer2024paligemma,paiss2023countbench},
        and RefCOCO~\citep{kazemzadeh2014referitgame,yu2016refcoco}
\end{itemize}

We compare Youtu-VL 2.0 with open-source multimodal base models:
Qwen3.5-35B-A3B-Base~\citep{qwen2026qwen35},
InternVL3.5-241B-A28B~\citep{wang2025internvl35} (pre-trained version),
ERNIE-4.5-VL-424B-A47B-Base~\citep{ernie2025technicalreport}, and
MiMo-V2.5-Base~\citep{xiaomi2026mimov25}.
All models are evaluated in the 3-shot and greedy generation setting.
Closed-form answers (e.g., multiple choice, yes/no, numbers, and coordinates) are scored by exact matching, while open-ended answers are scored by an LLM judge. The average is
computed over all listed benchmarks with equal weights.

Across the 25 multimodal benchmarks in Table~\ref{tab:eval-mm}, Youtu-VL 2.0
achieves an average score of 63.16, placing it in the top tier alongside
Qwen3.5 at 63.77 and MiMo-V2.5 at 63.33. It outperforms the much larger
InternVL3.5 and ERNIE-4.5-VL by a wide margin.
By category, the model is strongest in hallucination control, real-world visual
perception, and visual reasoning. It ranks second on all three hallucination
benchmarks, matching Qwen3.5 in visual faithfulness and clearly ahead of the
larger baselines. It ranks first on RealWorldQA, EmbSpatial-Bench, AI2D, and
WeMath.

The remaining gaps concentrate in two areas. On college-level discipline
benchmarks such as MMMU and MMMU-Pro, performance trails the leading baselines,
mirroring the SuperGPQA pattern in text evaluation and reflecting the capacity
limits of a 32B model for background knowledge. On RefCOCO and
OCRBench v2, the scores also leave room for improvement, indicating that
fine-grained bounding-box localization and dense text recognition lag behind
the model's stronger high-level spatial and counting abilities.

\begin{table*}[t]
\centering
\caption{Comparison of Youtu-VL 2.0 with open-source multimodal base models. All models are evaluated in the  3-shot and greedy generation setting.
Best scores are in \textbf{bold} and second-best scores are \underline{underlined}.}
\label{tab:eval-mm}
\setlength{\tabcolsep}{6pt}
\setlength{\aboverulesep}{0pt}
\setlength{\belowrulesep}{0pt}
\renewcommand{\arraystretch}{1.18}
\resizebox{0.95\textwidth}{!}{%
\begin{tabular}{@{}ll cccc >{\columncolor{youtuhl}}c@{}}
\toprule
\textbf{Task} & \textbf{Benchmark}
 & \textbf{Qwen3.5} & \textbf{InternVL3.5} & \textbf{ERNIE-4.5-VL} & \textbf{MiMo-V2.5}
 & \textbf{Youtu-VL 2.0} \\
\midrule
\multicolumn{2}{@{}l}{\textcolor{gray}{Size}} & \textcolor{gray}{35B-A3B} & \textcolor{gray}{241B-A28B} & \textcolor{gray}{424B-A47B} & \textcolor{gray}{310B-A15B} & \textcolor{gray}{32B-A4B} \\
\midrule
\multirow{6}{*}{STEM \& Math}
 & AI2D             & \underline{88.08} & 86.69 & 83.68 & 85.23 & \textbf{88.89} \\
 & LogicVista       & \underline{46.53} & 45.86 & 41.83 & \textbf{47.20} & 44.97 \\
 & WeMath           & 41.33 & 39.81 & 39.33 & \underline{45.14} & \textbf{50.00} \\
 & MMMU             & \underline{64.06} & 59.56 & 60.72 & \textbf{65.56} & 60.06 \\
 & MMMU-Pro         & \underline{50.06} & 48.44 & 49.42 & \textbf{50.12} & 43.64 \\
 & MathVista        & 69.50 & \textbf{71.60} & 67.70 & \underline{70.00} & 68.20 \\
\cmidrule(l{0pt}r{0pt}){1-7}
\multirow{9}{*}{General VQA}
 & MMBench-EN       & \textbf{87.85} & 87.00 & 85.14 & 85.14 & \underline{87.38} \\
 & MMBench-CN       & \textbf{87.31} & 84.75 & 83.51 & \underline{86.15} & 84.44 \\
 & SimpleVQA        & \textbf{51.01} & 45.73 & 49.04 & \underline{49.38} & 48.05 \\
 & ChineseSimpleVQA & \textbf{48.55} & 39.55 & \underline{47.95} & 47.27 & 47.45 \\
 & WorldVQA         & \underline{14.23} & 10.20 & \textbf{15.30} & 13.53 & 14.17 \\
 & BabyVision       & \underline{16.49} & 14.18 & 13.40 & \textbf{18.56} & \underline{16.49} \\
 & PerceptionBench  & \underline{31.63} & 27.13 & 26.17 & \textbf{32.57} & 31.00 \\
 & RealWorldQA      & 77.12 & 74.77 & 72.94 & \underline{78.56} & \textbf{79.61} \\
 & MMStar           & \textbf{70.07} & 67.67 & 64.00 & 68.07 & \underline{69.40} \\
\cmidrule(l{0pt}r{0pt}){1-7}
\multirow{4}{*}{Doc \& Chart}
 & ChartQA          & 86.96 & 73.88 & \textbf{87.96} & \underline{87.80} & 87.44 \\
 & CharXiv-RQ       & 58.10 & 43.90 & 42.20 & \textbf{65.40} & \underline{58.30} \\
 & OCRBench v2-EN   & 50.74 & \textbf{55.13} & \underline{52.43} & 52.00 & 50.48 \\
 & OCRBench v2-CN   & \underline{62.48} & 58.62 & \textbf{64.74} & 59.84 & 59.33 \\
\cmidrule(l{0pt}r{0pt}){1-7}
\multirow{3}{*}{Hallucination}
 & CRPE-Exist       & \textbf{97.06} & 95.78 & 96.20 & 94.14 & \underline{97.04} \\
 & CRPE-Relation    & 71.86 & \textbf{73.44} & 71.28 & 64.84 & \underline{72.86} \\
 & HallusionBench   & \textbf{61.16} & 54.49 & 54.27 & 57.65 & \underline{60.08} \\
\cmidrule(l{0pt}r{0pt}){1-7}
\multirow{3}{*}{Grounding}
 & EmbSpatial-Bench & \underline{82.06} & 75.41 & 69.86 & 80.00 & \textbf{82.69} \\
 & CountBenchQA     & 89.53 & 78.85 & 89.12 & \textbf{95.48} & \underline{94.87} \\
 & RefCOCO          & \textbf{90.42} & \underline{90.12} & 78.89 & 83.67 & 82.17 \\
\midrule
\multicolumn{2}{@{}l}{\textbf{Average}} & \textbf{63.77} & 60.10 & 60.28 & \underline{63.33} & 63.16 \\
\bottomrule
\end{tabular}%
}
\end{table*}

\subsection{Long-context modeling}
\label{sec:eval-longctx}
We evaluate long-context capabilities with three benchmarks covering synthetic retrieval, realistic long-document understanding, and long-range dialogue coreference:
\begin{itemize}
  \item \textbf{RULER}~\citep{hsieh2024ruler}: synthetic tasks including
        needle-in-a-haystack retrieval, multi-hop tracing, aggregation, and question answering, evaluated at context lengths of 4K, 8K, 16K, 32K, 64K, 128K, and 256K tokens.
  \item \textbf{LongBench v2}~\citep{bai2025longbenchv2}: realistic long-context multiple-choice questions. We report the
        \emph{short} and \emph{medium} subsets. The \emph{long} subset is excluded because most of its samples exceed 256K tokens.
  \item \textbf{MRCR v2}~\citep{vodrahalli2024michelangelo}: multi-round coreference resolution over long synthetic conversations, with 2, 4, and 8 needles. We group samples into three length bins ($<$32K, 32K-128K, and 128K-256K). Samples whose prompts exceed the evaluation window are discarded
\end{itemize}

We compare Youtu-VL 2.0 with Qwen3.5-35B-A3B-Base~\citep{qwen2026qwen35},
DeepSeek-V4-Flash-Base~\citep{deepseekai2026deepseekv4},
MiMo-V2.5-Base~\citep{xiaomi2026mimov25},
MiMo-V2.5-Pro-Base~\citep{xiaomi2026mimov25pro},
DeepSeek-V4-Pro-Base~\citep{deepseekai2026deepseekv4}, and
Kimi-K2-Base~\citep{kimi2025k2}. These baselines reach long contexts in different ways. Qwen3.5-35B-A3B-Base~\citep{qwen2026qwen35},  MiMo-V2.5-Base~\citep{xiaomi2026mimov25}, and MiMo-V2.5-Pro-Base~\citep{xiaomi2026mimov25pro}  is released with a 256K context. DeepSeek-V4-Flash-Base~\citep{deepseekai2026deepseekv4} and
DeepSeek-V4-Pro-Base~\citep{deepseekai2026deepseekv4} natively support a 1M-token context through hybrid compressed attention. Kimi-K2-Base~\citep{kimi2025k2}
officially supports a 128K context window, so we extend it to 256K with YaRN for evaluation. All long-context benchmarks are evaluated with the LM Evaluation Harness~\citep{eval-harness} with models served by
SGLang~\citep{zheng2024sglang}, using a maximum context length of 256K tokens. The overall average weights the nine groups of Table~\ref{tab:eval-longctx} equally.

\begin{table*}[!htbp]
\centering
\caption{Long-context evaluation of base models. ``Ctx.'' denotes the
officially supported context length of each base checkpoint. RULER $\le$32K is the equal-weight average of 4K, 8K,
16K, and 32K, and counts as one of the 9 groups in Avg. $^\dagger$Extended
to 256K with YaRN.
Best scores are in \textbf{bold} and second-best scores are \underline{underlined}.}
\label{tab:eval-longctx}
\setlength{\tabcolsep}{4.5pt}
\setlength{\aboverulesep}{0pt}
\setlength{\belowrulesep}{0pt}
\renewcommand{\arraystretch}{1.25}
\resizebox{\textwidth}{!}{%
\begin{tabular}{@{}lc cccc cc ccc c@{}}
\toprule
\multirow{2}{*}{\textbf{Model}} & \multirow{2}{*}{\textbf{Ctx.}}
 & \multicolumn{4}{c}{\textbf{RULER}}
 & \multicolumn{2}{c}{\textbf{LongBench v2}}
 & \multicolumn{3}{c}{\textbf{MRCR v2}}
 & \multirow{2}{*}{\textbf{Avg.}} \\
\cmidrule(lr){3-6}\cmidrule(lr){7-8}\cmidrule(lr){9-11}
 & & $\le$32K & 64K & 128K & 256K
 & Short & Medium
 & $\le$32K & 32K-128K & 128K-256K & \\
\midrule
Qwen3.5-35B-A3B & 256K
 & 88.33 & 87.94 & 87.53 & 84.70
 & 41.11 & 32.56
 & 32.04 & 25.69 & 15.94
 & 55.09 \\
DeepSeek-V4-Flash & 1M
 & 90.73 & 91.13 & 90.37 & \underline{90.31}
 & \textbf{52.22} & 43.26
 & \textbf{44.51} & 29.75 & \underline{27.18}
 & \underline{62.16} \\
MiMo-V2.5       & 256K
 & 94.34 & 92.48 & 90.62 & 87.32
 & 45.56 & \textbf{48.84}
 & 35.94 & 22.84 & 17.99
 & 59.55 \\
MiMo-V2.5-Pro   & 256K
 & 86.22 & 83.52 & 84.27 & 84.90
 & \underline{51.67} & \underline{47.44}
 & \underline{42.62} & 27.97 & 22.10
 & 58.97 \\
DeepSeek-V4-Pro & 1M
 & 93.04 & \underline{92.90} & \textbf{93.91} & \textbf{92.78}
 & \underline{51.67} & 40.93
 & 38.45 & 26.55 & 22.77
 & 61.44 \\
Kimi-K2         & 128K$^\dagger$
 & \textbf{95.53} & \textbf{93.86} & \underline{90.63} & 89.76
 & 50.00 & 38.14
 & 42.57 & \underline{35.08} & 25.02
 & \textbf{62.29} \\
\midrule
\rowcolor{youtuhl}
Youtu-VL 2.0    & 256K
 & \underline{94.55} & 92.30 & 88.58 & 83.94
 & 43.89 & 43.72
 & 41.69 & \textbf{35.90} & \textbf{28.58}
 & 61.46 \\
\bottomrule
\end{tabular}%
}
\end{table*}

Under the 256K-context evaluation in Table~\ref{tab:eval-longctx},
Youtu-VL 2.0 ranks third overall.
 Compared with Qwen3.5 at the same parameter scale, it leads by 6.37 points on average and wins eight of the nine groups.
Across model scales, it surpasses both MiMo-V2.5 models and matches the 1.6T
DeepSeek-V4-Pro despite the latter supporting a native 1M context window.
The model exhibits the strongest length robustness on realistic and multi-turn
long-context tasks. On MRCR v2, Youtu-VL 2.0 ranks first in both bins beyond 32K,
reaching 28.58 in the hardest bin. On
LongBench v2, its performance shows virtually no drop from the short to the
medium subset, enabling it to surpass Kimi-K2 and both DeepSeek-V4 models on
the medium split.

By contrast, on the synthetic RULER benchmark, Youtu-VL 2.0 remains strong up
to 64K tokens but degrades more noticeably toward the 256K boundary, trailing
the native 1M-context DeepSeek-V4 models. This gap reflects both the typical
boundary degradation of native 256K models at the edge of their context window
and the difference between synthetic stress tests and more realistic
long-document understanding or multi-turn coreference, where Youtu-VL 2.0 makes
more consistent use of long context.

\section{Discussion}
\label{sec:discussion}
\subsection{Visual encoder selection}
\label{sec:visual_encoder_selection}

We also trained an encoder-free visual embedder that replaces the pre-trained vision transformer by a direct projection from raw RGB patches into the decoder embedding space. Let $p_{i,j}\in\mathbb R^{3\cdot16^2}$ be the raw patch at fine-grid position $(i,j)$. The embedder normalizes and projects each patch, applies a depthwise convolution over the fine grid, and concatenates every $2\times2$ group in a fixed order,
\begin{equation}
z_{i,j}=\operatorname{GELU}\!\left(W_s\operatorname{LN}(p_{i,j})+b_s\right),
\qquad
\tilde z=z+\operatorname{DWConv}_{3}\!\left(\operatorname{LN}(z)\right),
\end{equation}
\begin{equation}
q_{u,v}=\big[\tilde z_{2u,2v},\tilde z_{2u,2v+1},\tilde z_{2u+1,2v},\tilde z_{2u+1,2v+1}\big],
\qquad
\bar z_{u,v}=\operatorname{LN}\!\left(\operatorname{LN}\!\left(W_{\mathrm{mrg}} q_{u,v}+b_{\mathrm{mrg}}\right)+e^x_u+e^y_v\right).
\end{equation}
Here $W_s\in\mathbb R^{d_s\times3\cdot16^2}$, $W_{\mathrm{mrg}}\in\mathbb R^{d_{\mathrm{mrg}}\times4d_s}$, and $e^x,e^y$ are coarse two-dimensional position tables. Two gated blocks then mix the merged tokens in sequence, first with kernel size $K=3$ and then with $K=5$. Each block applies a gated depthwise convolution followed by a gated SwiGLU channel mixer, and a normalization together with a final projection emits decoder-width embeddings,
\begin{equation}
r_{u,v}=\bar z_{u,v}+g_s\odot\operatorname{DWConv}_{K}\!\left(\operatorname{LN}(\bar z_{u,v})\right),
\qquad
c_{u,v}=r_{u,v}+g_c\odot\operatorname{SwiGLU}\!\left(\operatorname{LN}(r_{u,v})\right),
\end{equation}
\begin{equation}
e_{u,v}=W_o\operatorname{RMSNorm}(c_{u,v}).
\end{equation}
The gates $g_s,g_c$ are zero-initialized per-channel parameters, $c_{u,v}$ in the last equation is the output of the $K=5$ block, and $W_o\in\mathbb R^{d\times d_{\mathrm{mrg}}}$ with $d$ the decoder width. A fine grid of $h\times w$ patches yields $(h/2)(w/2)$ embeddings that replace the image-pad positions of the token sequence without changing its length. The whole embedder is initialized randomly and optimized jointly with the decoder, since no pre-trained weights exist for this design.

\begin{figure*}[!ht]
\centering
\includegraphics[width=0.98\textwidth]{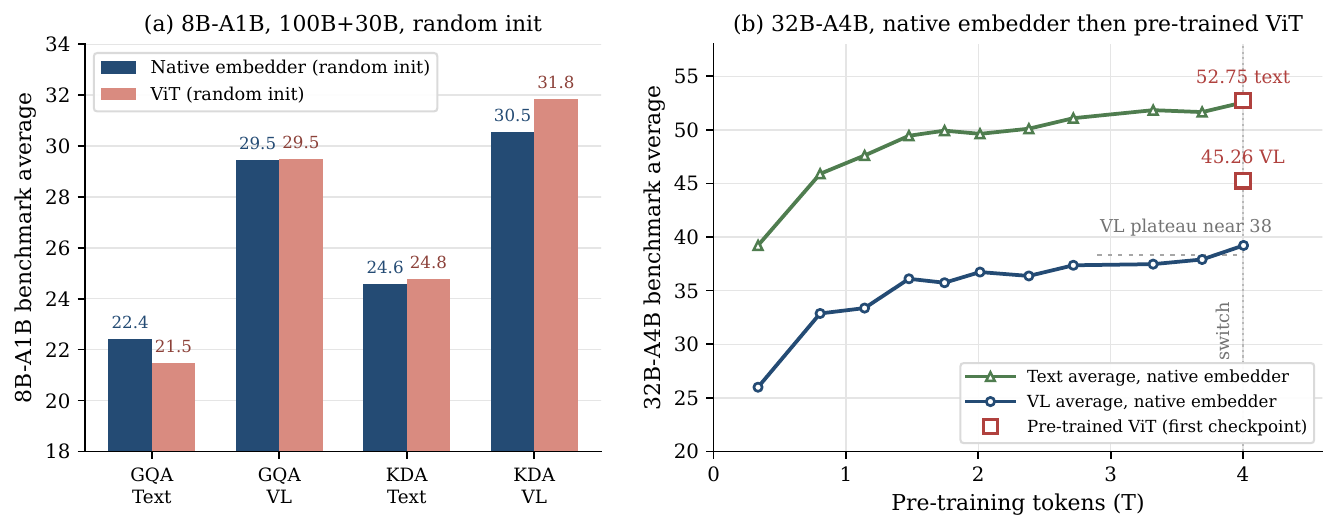}
\caption{Text and vision-language benchmark averages for two visual embedders. (a) At the Youtu-VL2.0-8B-A1B-Base scale and the 100B+30B budget, a randomly initialized native embedder and a randomly initialized ViT stay within 1.3 points of each other on both averages under GQA and KDA backbones. (b) At the 32B-A4B scale, the native embedder curves are shown up to 4T tokens, where the vision-language average plateaus near 38 and the text average reaches 52.57. The red squares mark the first checkpoint after switching to the pre-trained Qwen3-VL ViT, 45.26 on vision-language benchmarks and 52.75 on text, and are aligned with the 4T boundary for comparison. The two panels use different benchmark suites, so only within-panel differences are compared directly in this figure.}
\label{fig:native_cv}
\end{figure*}

At the Youtu-VL2.0-8B-A1B-Base scale we compared this embedder against a ViT of the Qwen3-VL architecture that was also initialized randomly under a shared training protocol. Figure~\ref{fig:native_cv}(a) reports text and vision-language averages for GQA and KDA backbones after training on 100B text tokens and 30B multimodal tokens, a budget we write as 100B+30B. Under GQA the two are within one point on both averages, at 22.42 against 21.47 on text and 29.46 against 29.48 on vision-language benchmarks. Under KDA the randomly initialized ViT leads by 0.22 on text and 1.29 on vision-language benchmarks. A controlled study that pretrains every variant from scratch attributes the encoder-free deficit on visual understanding to dedicated vision capacity~\cite{brown2026unbundling}. Our native embedder holds 31.95M vision-specific parameters against 443.03M for the ViT, so a smaller dedicated capacity is a plausible explanation for the KDA gap.

Figure~\ref{fig:native_cv}(b) tracks both averages through the first stage of the 32B-A4B run up to 4T tokens. The text average rises from 39.22 at 0.34T tokens to 52.57 at 4.0T tokens, while the vision-language average rises from 25.99 to a plateau near 38. For comparison, when a pre-trained ViT is used as the vision encoder, the average text score of its same 4.0T checkpoint is 52.75, a difference of less than 1 point compared to the score obtained using the native embedder. However, the average score for vision-language tasks reaches 45.26, which is approximately 7 points higher than the plateau level. The choice of visual encoder therefore leaves text-side scores largely unchanged in our run even when the encoder is pre-trained, while vision-side scores differ substantially. Brown et al.~\cite{brown2026unbundling} report the same separation, with encoder-free models trailing on Avg. VQA while Avg. Text stays comparable across their recipes.

We selected the pre-trained Qwen3-VL ViT for Youtu-VL 2.0 because its advantage in the measured comparison is clear, at 45.26 against 38.33 on vision-language benchmarks. Section~\ref{sec:method} describes the resulting architecture in detail.

\subsection{Latent routed experts}
\label{sec:discussion_latentmoe}

\begin{figure}[htbp]
\centering
\includegraphics[width=0.55\linewidth]{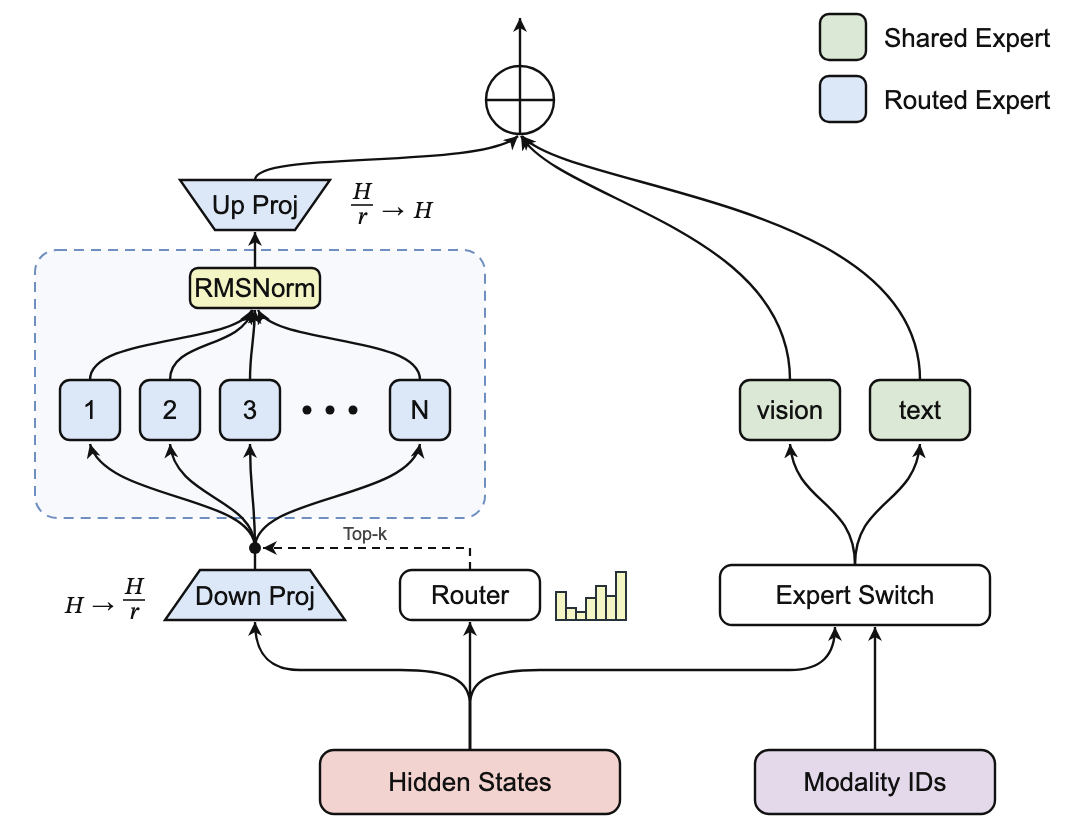}
\caption{Latent MMoE. Routed experts operate at reduced dimension $d/r$, with $r=2$ in our ablation, followed by RMSNorm and an up-projection. Both the router and modality-selected shared experts remain at the full dimension $d$. The figure labels the hidden dimension $H$ and the routed expert count $N$, which correspond to $d$ and $E$.}
\label{fig:discussion_latent_mmoe}
\end{figure}

LatentMoE~\cite{elango2026latentmoe} increases the number of routed experts by reducing the dimension at which each expert operates. We investigate this capacity allocation within our MMoE architecture, compressing the inputs and outputs of routed experts while keeping both the router and modality-specific shared experts at the full decoder hidden dimension (Figure~\ref{fig:discussion_latent_mmoe}). In our Latent MMoE variant, a shared down-projection halves the hidden dimension ($r=2$). The weighted routed outputs are then RMS-normalized and up-projected back to the full dimension before adding the selected text or vision shared branch. Consequently, the router retains access to the full input representation, while the modality-specific shared branch provides an uncompressed path alongside the latent routed computation.

To evaluate this design, we double both the routed expert count and top-$k$ activation, scaling from 128 experts (top-8) to 256 experts (top-16). At a fixed expert intermediate dimension, halving the input and output dimensions offsets the doubled expert count and activation, keeping leading expert parameters and matrix-multiplication FLOPs constant. This trades wider individual experts for a larger pool of narrower transformations. However, equal expert FLOPs do not translate to equal training throughput, as the latent variant introduces extra down- and up-projections, normalization, and routing overhead that must be evaluated in the full training stack.

In a 200B-token ablation at the Youtu-VL2.0-8B-A1B-Base scale using a shared backbone across both variants, Latent MMoE yields modest gains of 0.21 points on text benchmarks and 0.37 points on vision-language benchmarks over standard MMoE. However, the unoptimized implementation reduces training throughput by 27\%. Thus, despite comparable leading expert FLOPs, these small quality gains come at a substantial execution cost. Given this trade-off and the moderate sparsity of our final 32B-A4B model, we retain standard full-dimension MMoE in Youtu-VL2.0-32B-A4B-Base. Note that the measured throughput penalty reflects the specific implementation evaluated here rather than an intrinsic hardware limit, and revisiting this design choice would require a new comparison with an optimized latent implementation.

\subsection{Square-root loss reweighting}
\label{sec:discussion_loss_reweight}

At 256K context, packed sequences contain logical samples with widely varying numbers of supervised tokens. Under standard token averaging, a sample's loss coefficient scales linearly with its valid target count, allowing a few long samples to dominate the training objective. We apply square-root length reweighting to moderate this dependence while still assigning longer samples greater total weight than shorter ones. Specifically, for a sample with $n_i$ valid target tokens, each supervised token receives weight $1/\sqrt{n_i}$, giving the sample a total weight of $\sqrt{n_i}$ (corresponding to $\alpha=1/2$ in Equation~\ref{eq:loss_weight}). For instance, a sample with 16 times as many valid targets receives 4 times the total weight, rather than 16 times under token averaging.

Weights are computed independently over each logical sample within a packed sequence. Token counts exclude masked positions so that padding and unsupervised prompt tokens do not inflate a sample's coefficient, and fully masked samples contribute zero loss. Normalization uses the global weight sum aggregated across all micro-batches and data-parallel ranks. This mechanism complements the per-stage sample-type quotas in Section~\ref{sec:curriculum}: quotas govern which samples enter a training step, whereas length reweighting governs their relative contributions to the supervised objective. The weight depends solely on valid target length, so the reweighting rule introduces no explicit preference toward specific modalities or benchmark categories in the evaluation suite.

\begin{figure}[htbp]
\centering
\includegraphics[width=0.98\textwidth]{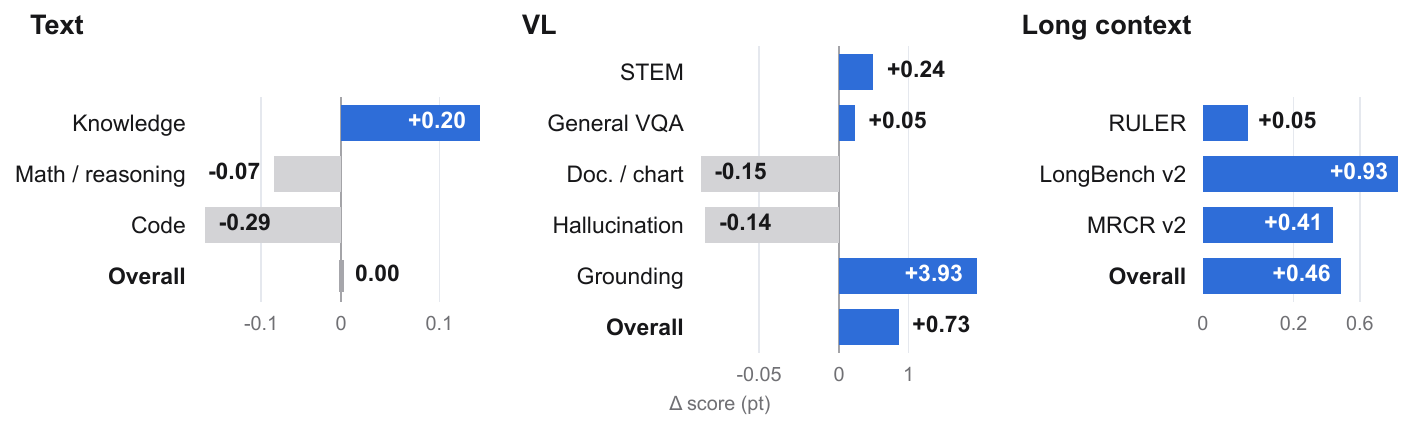}
\caption{Score differences (square-root reweighting minus token averaging) after 100B tokens per variant at 256K context. Axes use square-root scaling with separate positive and negative normalization, while annotations report raw point changes. Text and VL panels show category and overall averages. Long context equally averages RULER's 64K, 128K, and 256K scores and MRCR v2's 32K-128K and 128K-256K bins, alongside LongBench v2's 32K-128K subset. Its overall score is the equal-weight mean of these three benchmark averages, excluding results below 32K.}
\label{fig:loss_reweight_summary}
\end{figure}

A 100B-token ablation per variant at 256K compares square-root reweighting against token averaging (Figure~\ref{fig:loss_reweight_summary}). While the text benchmark average holds steady at 61.90, this aggregate reflects a small gain in knowledge ($+0.20$) offset by minor declines in math/reasoning ($-0.07$) and code ($-0.29$). By contrast, the vision-language average rises from 61.81 to 62.54 ($+0.73$). Grounding accounts for the largest category gain ($+3.93$), driven by improvements on RefSpatial ($+9.20$), RefCOCO ($+4.78$), and OSWorld-G ($+4.61$), while changes across all other VL categories remain within 0.25 points. The visual improvement is thus concentrated in grounding rather than uniformly distributed across the evaluation suite.

For long-context evaluation, we summarize only results at 32K tokens and above. RULER, LongBench v2, and MRCR v2 improve by 0.05, 0.93, and 0.41 points, respectively, yielding an equal-weight overall gain of 0.46 points. Note that these benchmark means summarize performance across the selected length bins rather than implying improvements at every individual length. Taken together, the substantial grounding gains, stable aggregate text performance, and modest long-context improvements support retaining square-root reweighting in the 256K stage. At the same time, this ablation does not isolate which specific sample lengths drive these gains, nor does it establish training-stability benefits.

\section{Conclusion}
\label{sec:conclusion}
We have presented \textbf{Youtu-VL2.0-32B-A4B-Base}, a native multimodal Mixture-of-Experts base model with 32B total and 4B active parameters. The architecture pairs a 3:1 KDA and gated DSA hybrid decoder with token-level modality specialization across shared experts, quantile-balanced routing, and lexical $N$-gram memory. Supported by a co-designed training stack and a progressive 256K curriculum, the model achieves strong performance across text, vision-language, coding, and long-context benchmarks while delivering efficient serving latency. Our ablation studies further demonstrate how training objectives and systems execution shape model quality, showing that square-root length reweighting boosts visual grounding and that architectural efficiency depends on full-stack kernel and communication design.

\setcitestyle{square,numbers,comma}
\bibliography{youtu_bib}

@inproceedings{vaswani2017attention,
  author    = {Ashish Vaswani and Noam Shazeer and Niki Parmar and others},
  title     = {Attention Is All You Need},
  booktitle = {Advances in Neural Information Processing Systems},
  year      = {2017},
  eprint    = {1706.03762},
  archiveprefix = {arXiv},
  url       = {https://arxiv.org/abs/1706.03762}
}

@inproceedings{shazeer2017outrageously,
  author    = {Noam Shazeer and Azalia Mirhoseini and Krzysztof Maziarz and others},
  title     = {Outrageously Large Neural Networks: The Sparsely-Gated Mixture-of-Experts Layer},
  booktitle = {International Conference on Learning Representations},
  year      = {2017},
  eprint    = {1701.06538},
  archiveprefix = {arXiv},
  url       = {https://arxiv.org/abs/1701.06538}
}

@inproceedings{lepikhin2021gshard,
  author    = {Dmitry Lepikhin and HyoukJoong Lee and Yuanzhong Xu and others},
  title     = {{GShard}: Scaling Giant Models with Conditional Computation and Automatic Sharding},
  booktitle = {International Conference on Learning Representations},
  year      = {2021},
  eprint    = {2006.16668},
  archiveprefix = {arXiv},
  url       = {https://arxiv.org/abs/2006.16668}
}

@article{fedus2022switch,
  author    = {William Fedus and Barret Zoph and Noam Shazeer},
  title     = {Switch Transformers: Scaling to Trillion Parameter Models with Simple and Efficient Sparsity},
  journal   = {Journal of Machine Learning Research},
  volume    = {23},
  number    = {120},
  year      = {2022},
  eprint    = {2101.03961},
  archiveprefix = {arXiv},
  url       = {https://arxiv.org/abs/2101.03961}
}

@article{deepseekai2024deepseekv2,
  author    = {{DeepSeek-AI}},
  title     = {{DeepSeek-V2}: A Strong, Economical, and Efficient Mixture-of-Experts Language Model},
  journal   = {arXiv preprint arXiv:2405.04434},
  year      = {2024},
  eprint    = {2405.04434},
  archiveprefix = {arXiv},
  url       = {https://arxiv.org/abs/2405.04434}
}

@article{deepseekai2024deepseekv3,
  author    = {{DeepSeek-AI}},
  title     = {{DeepSeek-V3} Technical Report},
  journal   = {arXiv preprint arXiv:2412.19437},
  year      = {2024},
  eprint    = {2412.19437},
  archiveprefix = {arXiv},
  url       = {https://arxiv.org/abs/2412.19437}
}

@misc{deepseekai2025deepseekv32,
  title        = {DeepSeek-V3.2: Pushing the Frontier of Open Large Language Models},
  author       = {{DeepSeek-AI}},
  year         = {2025},
  eprint       = {2512.02556},
  archivePrefix = {arXiv},
  primaryClass = {cs.CL},
  url          = {https://arxiv.org/abs/2512.02556}
}

@inproceedings{yang2024parallelizing,
  author    = {Songlin Yang and Bailin Wang and Yu Zhang and Yikang Shen and Yoon Kim},
  title     = {Parallelizing Linear Transformers with the Delta Rule over Sequence Length},
  booktitle = {Advances in Neural Information Processing Systems},
  year      = {2024},
  eprint    = {2406.06484},
  archiveprefix = {arXiv},
  url       = {https://arxiv.org/abs/2406.06484}
}

@inproceedings{yang2025gated,
  author    = {Songlin Yang and Jan Kautz and Ali Hatamizadeh},
  title     = {Gated Delta Networks: Improving {Mamba2} with Delta Rule},
  booktitle = {International Conference on Learning Representations},
  year      = {2025},
  eprint    = {2412.06464},
  archiveprefix = {arXiv},
  url       = {https://arxiv.org/abs/2412.06464}
}

@article{kimiteam2025kimilinear,
  author    = {{Kimi Team} and Yu Zhang and Zongyu Lin and others},
  title     = {{Kimi Linear}: An Expressive, Efficient Attention Architecture},
  journal   = {arXiv preprint arXiv:2510.26692},
  year      = {2025},
  eprint    = {2510.26692},
  archiveprefix = {arXiv},
  url       = {https://arxiv.org/abs/2510.26692}
}

@misc{inclusionai2026ling30flash,
  author       = {{inclusionAI}},
  year         = {2026},
  title        = {{Ling-3.0-Flash} Model Card},
  howpublished = {Hugging Face model card},
  url          = {https://huggingface.co/inclusionAI/Ling-3.0-flash},
  note         = {Accessed: 2026-10-20}
}

@misc{inclusionai2026ling31flash,
  author       = {{inclusionAI}},
  year         = {2026},
  title        = {{Ling-3.1-Flash} Model Card},
  howpublished = {Hugging Face model card},
  url          = {https://huggingface.co/inclusionAI/Ling-3.1-flash},
  note         = {Accessed: 2026-10-20}
}

@article{kimiteam2026kimik3,
  author    = {{Kimi Team} and Tongtong Bai and Yifan Bai and others},
  title     = {{Kimi K3}: Open Frontier Intelligence},
  journal   = {arXiv preprint arXiv:2607.24653},
  year      = {2026},
  eprint    = {2607.24653},
  archiveprefix = {arXiv},
  url       = {https://arxiv.org/abs/2607.24653},
  note      = {Model card: https://huggingface.co/moonshotai/Kimi-K3}
}

@article{qiu2026qwen38next,
  author    = {Zihan Qiu and Zekun Wang and Xiao Li and others},
  title     = {On the Design of {Qwen3.8-Next} Architecture: Evaluation, Efficiency, and Training Stability},
  journal   = {arXiv preprint arXiv:2608.30320},
  year      = {2026},
  eprint    = {2608.30320},
  archiveprefix = {arXiv},
  url       = {https://arxiv.org/abs/2608.30320}
}

@misc{zai2026glm53flash,
  author       = {{Z.ai}},
  year         = {2026},
  title        = {{GLM-5.3-Flash} Model Card},
  howpublished = {Hugging Face model card},
  url          = {https://huggingface.co/zai-org/GLM-5.3-Flash},
  note         = {Model configuration: https://huggingface.co/zai-org/GLM-5.3-Flash/blob/main/config.json. Accessed: 2026-10-20}
}

@article{deepseekai2026deepseekv41flash,
  author    = {{DeepSeek-AI}},
  title     = {{DeepSeek-V4.1-Flash}: Pushing the Limits of {KV} Cache Compression},
  journal   = {arXiv preprint arXiv:2609.19969},
  year      = {2026},
  eprint    = {2609.19969},
  archiveprefix = {arXiv},
  url       = {https://arxiv.org/abs/2609.19969}
}

@article{zhu2024hyperconnections,
  author    = {Defa Zhu and Hongzhi Huang and Zihao Huang and Yutao Zeng and Yunyao Mao and Banggu Wu and Qiyang Min and Xun Zhou},
  title     = {Hyper-Connections},
  journal   = {arXiv preprint arXiv:2409.19606},
  year      = {2024},
  eprint    = {2409.19606},
  archiveprefix = {arXiv},
  url       = {https://arxiv.org/abs/2409.19606}
}

@article{xie2025mhc,
  author    = {Zhenda Xie and Yixuan Wei and Huanqi Cao and others},
  title     = {{mHC}: Manifold-Constrained Hyper-Connections},
  journal   = {arXiv preprint arXiv:2512.24880},
  year      = {2025},
  eprint    = {2512.24880},
  archiveprefix = {arXiv},
  url       = {https://arxiv.org/abs/2512.24880}
}

@inproceedings{alayrac2022flamingo,
  author    = {Jean-Baptiste Alayrac and Jeff Donahue and Pauline Luc and others},
  title     = {Flamingo: A Visual Language Model for Few-Shot Learning},
  booktitle = {Advances in Neural Information Processing Systems},
  year      = {2022},
  eprint    = {2204.14198},
  archiveprefix = {arXiv},
  url       = {https://arxiv.org/abs/2204.14198}
}

@inproceedings{liu2023visual,
  author    = {Haotian Liu and Chunyuan Li and Qingyang Wu and Yong Jae Lee},
  title     = {Visual Instruction Tuning},
  booktitle = {Advances in Neural Information Processing Systems},
  year      = {2023},
  eprint    = {2304.08485},
  archiveprefix = {arXiv},
  url       = {https://arxiv.org/abs/2304.08485}
}

@inproceedings{dehghani2023patch,
  author    = {Mostafa Dehghani and Basil Mustafa and Josip Djolonga and others},
  title     = {Patch n' Pack: {NaViT}, a Vision Transformer for Any Aspect Ratio and Resolution},
  booktitle = {Advances in Neural Information Processing Systems},
  year      = {2023},
  eprint    = {2307.06304},
  archiveprefix = {arXiv},
  url       = {https://arxiv.org/abs/2307.06304}
}

@article{wang2024qwen2vl,
  author    = {Peng Wang and Shuai Bai and Sinan Tan and others},
  title     = {{Qwen2-VL}: Enhancing Vision-Language Model's Perception of the World at Any Resolution},
  journal   = {arXiv preprint arXiv:2409.12191},
  year      = {2024},
  eprint    = {2409.12191},
  archiveprefix = {arXiv},
  url       = {https://arxiv.org/abs/2409.12191}
}

@article{bai2025qwen3vl,
  author    = {Shuai Bai and Yuxuan Cai and Ruizhe Chen and others},
  title     = {{Qwen3-VL} Technical Report},
  journal   = {arXiv preprint arXiv:2511.21631},
  year      = {2025},
  eprint    = {2511.21631},
  archiveprefix = {arXiv},
  url       = {https://arxiv.org/abs/2511.21631}
}

@article{wei2026youtuvl,
  author    = {Zhixiang Wei and Yi Li and Zhehan Kan and others},
  title     = {{Youtu-VL}: Unleashing Visual Potential via Unified Vision-Language Supervision},
  journal   = {arXiv preprint arXiv:2601.19798},
  year      = {2026},
  eprint    = {2601.19798},
  archiveprefix = {arXiv},
  url       = {https://arxiv.org/abs/2601.19798}
}

@article{abtahizadeh2026museglimmer,
  author    = {Arash Ebrahim Abtahizadeh and Muna Aghamelu and Adil Ahmed and others},
  title     = {{Muse Glimmer 30B}: An On-Device Agentic Model},
  journal   = {arXiv preprint arXiv:2610.24628},
  year      = {2026},
  eprint    = {2610.24628},
  archiveprefix = {arXiv},
  url       = {https://arxiv.org/abs/2610.24628},
  note      = {Model card: https://huggingface.co/meta-models/Muse-Glimmer-30B}
}

@article{wortsman2023smallscale,
  author    = {Mitchell Wortsman and Peter J. Liu and Lechao Xiao and others},
  title     = {Small-Scale Proxies for Large-Scale Transformer Training Instabilities},
  journal   = {arXiv preprint arXiv:2309.14322},
  year      = {2023},
  eprint    = {2309.14322},
  archiveprefix = {arXiv},
  url       = {https://arxiv.org/abs/2309.14322}
}

@inproceedings{yue2024mmmu,
  author    = {Xiang Yue and Yuansheng Ni and Kai Zhang and others},
  title     = {{MMMU}: A Massive Multi-Discipline Multimodal Understanding and Reasoning Benchmark for Expert {AGI}},
  booktitle = {IEEE/CVF Conference on Computer Vision and Pattern Recognition},
  year      = {2024},
  eprint    = {2311.16502},
  archiveprefix = {arXiv},
  url       = {https://arxiv.org/abs/2311.16502}
}

@inproceedings{liu2024mmbench,
  author    = {Yuan Liu and Haodong Duan and Yuanhan Zhang and others},
  title     = {{MMBench}: Is Your Multi-Modal Model an All-Around Player?},
  booktitle = {European Conference on Computer Vision},
  pages     = {216-233},
  year      = {2024},
  doi       = {10.1007/978-3-031-72658-3_13},
  eprint    = {2307.06281},
  archiveprefix = {arXiv},
  url       = {https://doi.org/10.1007/978-3-031-72658-3_13}
}

@article{liu2024ocrbench,
  author    = {Yuliang Liu and Zhang Li and Mingxin Huang and others},
  title     = {{OCRBench}: On the Hidden Mystery of {OCR} in Large Multimodal Models},
  journal   = {Science China Information Sciences},
  volume    = {67},
  pages     = {220102},
  year      = {2024},
  doi       = {10.1007/s11432-024-4235-6},
  eprint    = {2305.07895},
  archiveprefix = {arXiv},
  url       = {https://doi.org/10.1007/s11432-024-4235-6}
}

@inproceedings{masry2022chartqa,
  author    = {Ahmed Masry and Do Xuan Long and Jia Qing Tan and Shafiq Joty and Enamul Hoque},
  title     = {{ChartQA}: A Benchmark for Question Answering about Charts with Visual and Logical Reasoning},
  booktitle = {Findings of the Association for Computational Linguistics: ACL 2022},
  pages     = {2263-2279},
  year      = {2022},
  eprint    = {2203.10244},
  archiveprefix = {arXiv},
  url       = {https://aclanthology.org/2022.findings-acl.177/}
}

@inproceedings{rein2024gpqa,
  author    = {David Rein and Betty Li Hou and Asa Cooper Stickland and others},
  title     = {{GPQA}: A Graduate-Level Google-Proof {Q\&A} Benchmark},
  booktitle = {Conference on Language Modeling},
  year      = {2024},
  eprint    = {2311.12022},
  archiveprefix = {arXiv},
  url       = {https://openreview.net/forum?id=Ti67584b98}
}

@article{cassano2023multiple,
  author    = {Federico Cassano and John Gouwar and Daniel Nguyen and others},
  title     = {{MultiPL-E}: A Scalable and Polyglot Approach to Benchmarking Neural Code Generation},
  journal   = {IEEE Transactions on Software Engineering},
  volume    = {49},
  number    = {7},
  pages     = {3675-3691},
  year      = {2023},
  doi       = {10.1109/TSE.2023.3267446},
  eprint    = {2208.08227},
  archiveprefix = {arXiv},
  url       = {https://doi.org/10.1109/TSE.2023.3267446}
}

@inproceedings{liu2024repobench,
  author    = {Tianyang Liu and Canwen Xu and Julian McAuley},
  title     = {{RepoBench}: Benchmarking Repository-Level Code Auto-Completion Systems},
  booktitle = {International Conference on Learning Representations},
  year      = {2024},
  eprint    = {2306.03091},
  archiveprefix = {arXiv},
  url       = {https://openreview.net/forum?id=pPjZIOuQuF}
}

@article{tong2026beyond,
  author    = {Shengbang Tong and David Fan and John Nguyen and others},
  title     = {Beyond Language Modeling: An Exploration of Multimodal Pretraining},
  journal   = {arXiv preprint arXiv:2603.03276},
  year      = {2026},
  eprint    = {2603.03276},
  archiveprefix = {arXiv},
  url       = {https://arxiv.org/abs/2603.03276}
}

@article{cheng2026conditional,
  author    = {Xin Cheng and Rui Tian and Wangding Zeng and others},
  title     = {Conditional Memory via Scalable Lookup: A New Axis of Sparsity for Large Language Models},
  journal   = {arXiv preprint arXiv:2601.07372},
  year      = {2026},
  eprint    = {2601.07372},
  archiveprefix = {arXiv},
  url       = {https://arxiv.org/abs/2601.07372}
}

@misc{eval-harness,
  title        = {The Language Model Evaluation Harness},
  author       = {Gao, Leo and Tow, Jonathan and Abbasi, Baber and Biderman, Stella and Black, Sid and DiPofi, Anthony and Foster, Charles and Golding, Laurence and Hsu, Jeffrey and Le Noac'h, Alain and Li, Haonan and McDonell, Kyle and Muennighoff, Niklas and Ociepa, Chris and Phang, Jason and Reynolds, Laria and Schoelkopf, Hailey and Skowron, Aviya and Sutawika, Lintang and Tang, Eric and Thite, Anish and Wang, Ben and Wang, Kevin and Zou, Andy},
  year         = {2024},
  publisher    = {Zenodo},
  version      = {v0.4.3},
  doi          = {10.5281/zenodo.12608602},
  url          = {https://zenodo.org/records/12608602}
}

@misc{duan2024vlmevalkit,
  title         = {{VLMEvalKit: An Open-Source Toolkit for Evaluating Large Multi-Modality Models}},
  author        = {Haodong Duan and Xinyu Fang and Junming Yang and Xiangyu Zhao and Zerun Ma and Yuxuan Qiao and Mo Li and Tianhao Liang and Lin Zhu and Amit Agarwal and Xiaozhe Li and Shengyuan Ding and Jiazi Bu and Ziyu Liu and Zhangyang Qi and Yifei Li and Yuhang Zang and Zhe Chen and Lin Chen and Yuan Liu and Yubo Ma and Hailong Sun and Yifan Zhang and Shiyin Lu and Tack Hwa Wong and Weiyun Wang and Peiheng Zhou and Chaoyou Fu and Junbo Cui and Jixuan Chen and Enxin Song and Song Mao and Junming Lin and Xilin Wei and Jinsong Li and Zeyi Sun and Zhaowei Wang and Zicheng Zhang and Xiaoyi Dong and Junjun He and Pan Zhang and Jiaqi Wang and Dahua Lin and Kai Chen},
  year          = {2024},
  eprint        = {2407.11691},
  archivePrefix = {arXiv},
  primaryClass  = {cs.CV},
  url           = {https://arxiv.org/abs/2407.11691}
}

@misc{zheng2024sglang,
  title         = {{SGLang: Efficient Execution of Structured Language Model Programs}},
  author        = {Lianmin Zheng and Liangsheng Yin and Zhiqiang Xie and Chuyue Sun and Jeff Huang and Cody Hao Yu and Shiyi Cao and Christos Kozyrakis and Ion Stoica and Joseph E. Gonzalez and Clark Barrett and Ying Sheng},
  year          = {2023},
  eprint        = {2312.07104},
  archivePrefix = {arXiv},
  primaryClass  = {cs.AI},
  url           = {https://arxiv.org/abs/2312.07104}
}

@misc{deepseekai2026deepseekv4,
  title         = {{DeepSeek-V4: Towards Highly Efficient Million-Token Context Intelligence}},
  author        = {{DeepSeek-AI}},
  year          = {2026},
  eprint        = {2606.19348},
  archivePrefix = {arXiv},
  primaryClass  = {cs.CL},
  url           = {https://arxiv.org/abs/2606.19348}
}

@misc{kimi2025k2,
  title         = {{Kimi K2: Open Agentic Intelligence}},
  author        = {{Kimi Team}},
  year          = {2025},
  eprint        = {2507.20534},
  archivePrefix = {arXiv},
  primaryClass  = {cs.LG},
  url           = {https://arxiv.org/abs/2507.20534}
}

@misc{xiaomi2026mimov25pro,
  title        = {{MiMo-V2.5-Pro}},
  author       = {{Xiaomi MiMo Team}},
  year         = {2026},
  howpublished = {\url{https://mimo.xiaomi.com/mimo-v2-5-pro/}},
  note         = {Model card: \url{https://huggingface.co/XiaomiMiMo/MiMo-V2.5-Pro-Base}}
}

@misc{xiaomi2026mimov25,
  title        = {{MiMo-V2.5}},
  author       = {{Xiaomi MiMo Team}},
  year         = {2026},
  howpublished = {\url{https://mimo.xiaomi.com/mimo-v2-5}}
}

@misc{qwen2026qwen35,
  title        = {{Qwen3.5}: Towards Native Multimodal Agents},
  author       = {{Qwen Team}},
  year         = {2026},
  month        = {February},
  howpublished = {\url{https://qwen.ai/blog?id=qwen3.5}}
}

@misc{wang2025internvl35,
  title         = {{InternVL3.5: Advancing Open-Source Multimodal Models in Versatility, Reasoning, and Efficiency}},
  author        = {Weiyun Wang and Zhangwei Gao and Lixin Gu and Hengjun Pu and Long Cui and Xingguang Wei and Zhaoyang Liu and Linglin Jing and Shenglong Ye and Jie Shao and Zhaokai Wang and Zhe Chen and Hongjie Zhang and Ganlin Yang and Haomin Wang and Qi Wei and Jinhui Yin and Wenhao Li and Erfei Cui and Guanzhou Chen and Zichen Ding and Changyao Tian and Zhenyu Wu and Jingjing Xie and Zehao Li and Bowen Yang and Yuchen Duan and Xuehui Wang and Zhi Hou and Haoran Hao and Tianyi Zhang and Songze Li and Xiangyu Zhao and Haodong Duan and Nianchen Deng and Bin Fu and Yinan He and Yi Wang and Conghui He and Botian Shi and Junjun He and Yingtong Xiong and Han Lv and Lijun Wu and Wenqi Shao and Kaipeng Zhang and Huipeng Deng and Biqing Qi and Jiaye Ge and Qipeng Guo and Wenwei Zhang and Songyang Zhang and Maosong Cao and Junyao Lin and Kexian Tang and Jianfei Gao and Haian Huang and Yuzhe Gu and Chengqi Lyu and Huanze Tang and Rui Wang and Haijun Lv and Wanli Ouyang and Limin Wang and Min Dou and Xizhou Zhu and Tong Lu and Dahua Lin and Jifeng Dai and Weijie Su and Bowen Zhou and Kai Chen and Yu Qiao and Wenhai Wang and Gen Luo},
  year          = {2025},
  eprint        = {2508.18265},
  archivePrefix = {arXiv},
  primaryClass  = {cs.CV},
  url           = {https://arxiv.org/abs/2508.18265}
}

@misc{ernie2025technicalreport,
  title        = {{ERNIE 4.5} Technical Report},
  author       = {{Baidu ERNIE Team}},
  year         = {2025},
  eprint       = {},
  archivePrefix = {arXiv},
  primaryClass = {cs.CL},
  url          = {https://github.com/PaddlePaddle/ERNIE}
}

@misc{hendrycks2021mmlu,
  title         = {{Measuring Massive Multitask Language Understanding}},
  author        = {Dan Hendrycks and Collin Burns and Steven Basart and Andy Zou and Mantas Mazeika and Dawn Song and Jacob Steinhardt},
  year          = {2020},
  eprint        = {2009.03300},
  archivePrefix = {arXiv},
  primaryClass  = {cs.CY},
  url           = {https://arxiv.org/abs/2009.03300}
}

@misc{gema2025mmluredux,
  title         = {{Are We Done with MMLU?}},
  author        = {Aryo Pradipta Gema and Joshua Ong Jun Leang and Giwon Hong and Alessio Devoto and Alberto Carlo Maria Mancino and Rohit Saxena and Xuanli He and Yu Zhao and Xiaotang Du and Mohammad Reza Ghasemi Madani and Claire Barale and Robert McHardy and Joshua Harris and Jean Kaddour and Emile van Krieken and Pasquale Minervini},
  year          = {2024},
  eprint        = {2406.04127},
  archivePrefix = {arXiv},
  primaryClass  = {cs.CL},
  url           = {https://arxiv.org/abs/2406.04127}
}

@misc{wang2024mmlupro,
  title         = {{MMLU-Pro: A More Robust and Challenging Multi-Task Language Understanding Benchmark}},
  author        = {Yubo Wang and Xueguang Ma and Ge Zhang and Yuansheng Ni and Abhranil Chandra and Shiguang Guo and Weiming Ren and Aaran Arulraj and Xuan He and Ziyan Jiang and Tianle Li and Max Ku and Kai Wang and Alex Zhuang and Rongqi Fan and Xiang Yue and Wenhu Chen},
  year          = {2024},
  eprint        = {2406.01574},
  archivePrefix = {arXiv},
  primaryClass  = {cs.CL},
  url           = {https://arxiv.org/abs/2406.01574}
}

@misc{openai2024mmmlu,
  title        = {{MMMLU}: Multilingual Massive Multitask Language Understanding},
  author       = {{OpenAI}},
  year         = {2024},
  howpublished = {\url{https://huggingface.co/datasets/openai/MMMLU}}
}

@misc{wei2024simpleqa,
  title         = {{Measuring short-form factuality in large language models}},
  author        = {Jason Wei and Nguyen Karina and Hyung Won Chung and Yunxin Joy Jiao and Spencer Papay and Amelia Glaese and John Schulman and William Fedus},
  year          = {2024},
  eprint        = {2411.04368},
  archivePrefix = {arXiv},
  primaryClass  = {cs.CL},
  url           = {https://arxiv.org/abs/2411.04368}
}

@misc{haas2025simpleqaverified,
  title         = {{SimpleQA Verified: A Reliable Factuality Benchmark to Measure Parametric Knowledge}},
  author        = {Lukas Haas and Gal Yona and Giovanni D'Antonio and Sasha Goldshtein and Dipanjan Das},
  year          = {2025},
  eprint        = {2509.07968},
  archivePrefix = {arXiv},
  primaryClass  = {cs.CL},
  url           = {https://arxiv.org/abs/2509.07968}
}

@misc{romanou2025include,
  title         = {{INCLUDE: Evaluating Multilingual Language Understanding with Regional Knowledge}},
  author        = {Angelika Romanou and Negar Foroutan and Anna Sotnikova and Zeming Chen and Sree Harsha Nelaturu and Shivalika Singh and Rishabh Maheshwary and Micol Altomare and Mohamed A. Haggag and Snegha A and Alfonso Amayuelas and Azril Hafizi Amirudin and Viraat Aryabumi and Danylo Boiko and Michael Chang and Jenny Chim and Gal Cohen and Aditya Kumar Dalmia and Abraham Diress and Sharad Duwal and Daniil Dzenhaliou and Daniel Fernando Erazo Florez and Fabian Farestam and Joseph Marvin Imperial and Shayekh Bin Islam and Perttu Isotalo and Maral Jabbarishiviari and Börje F. Karlsson and Eldar Khalilov and Christopher Klamm and Fajri Koto and Dominik Krzemiński and Gabriel Adriano de Melo and Syrielle Montariol and Yiyang Nan and Joel Niklaus and Jekaterina Novikova and Johan Samir Obando Ceron and Debjit Paul and Esther Ploeger and Jebish Purbey and Swati Rajwal and Selvan Sunitha Ravi and Sara Rydell and Roshan Santhosh and Drishti Sharma and Marjana Prifti Skenduli and Arshia Soltani Moakhar and Bardia Soltani Moakhar and Ran Tamir and Ayush Kumar Tarun and Azmine Toushik Wasi and Thenuka Ovin Weerasinghe and Serhan Yilmaz and Mike Zhang and Imanol Schlag and Marzieh Fadaee and Sara Hooker and Antoine Bosselut},
  year          = {2024},
  eprint        = {2411.19799},
  archivePrefix = {arXiv},
  primaryClass  = {cs.CL},
  url           = {https://arxiv.org/abs/2411.19799}
}

@misc{zhong2023agieval,
  title         = {{AGIEval: A Human-Centric Benchmark for Evaluating Foundation Models}},
  author        = {Wanjun Zhong and Ruixiang Cui and Yiduo Guo and Yaobo Liang and Shuai Lu and Yanlin Wang and Amin Saied and Weizhu Chen and Nan Duan},
  year          = {2023},
  eprint        = {2304.06364},
  archivePrefix = {arXiv},
  primaryClass  = {cs.CL},
  url           = {https://arxiv.org/abs/2304.06364}
}

@misc{du2025supergpqa,
  title         = {{SuperGPQA: Scaling LLM Evaluation across 285 Graduate Disciplines}},
  author        = {{M-A-P Team}},
  year          = {2025},
  eprint        = {2502.14739},
  archivePrefix = {arXiv},
  primaryClass  = {cs.CL},
  url           = {https://arxiv.org/abs/2502.14739}
}

@misc{cobbe2021gsm8k,
  title         = {{Training Verifiers to Solve Math Word Problems}},
  author        = {Karl Cobbe and Vineet Kosaraju and Mohammad Bavarian and Mark Chen and Heewoo Jun and Lukasz Kaiser and Matthias Plappert and Jerry Tworek and Jacob Hilton and Reiichiro Nakano and Christopher Hesse and John Schulman},
  year          = {2021},
  eprint        = {2110.14168},
  archivePrefix = {arXiv},
  primaryClass  = {cs.LG},
  url           = {https://arxiv.org/abs/2110.14168}
}

@inproceedings{shi2023mgsm,
  title         = {{Language Models are Multilingual Chain-of-Thought Reasoners}},
  author        = {Freda Shi and Mirac Suzgun and Markus Freitag and Xuezhi Wang and Suraj Srivats and Soroush Vosoughi and Hyung Won Chung and Yi Tay and Sebastian Ruder and Denny Zhou and Dipanjan Das and Jason Wei},
  booktitle     = {The Eleventh International Conference on Learning Representations},
  year          = {2023},
  eprint        = {2210.03057},
  archivePrefix = {arXiv},
  primaryClass  = {cs.CL},
  url           = {https://arxiv.org/abs/2210.03057}
}

@misc{hendrycks2021math,
  title         = {{Measuring Mathematical Problem Solving With the MATH Dataset}},
  author        = {Dan Hendrycks and Collin Burns and Saurav Kadavath and Akul Arora and Steven Basart and Eric Tang and Dawn Song and Jacob Steinhardt},
  year          = {2021},
  eprint        = {2103.03874},
  archivePrefix = {arXiv},
  primaryClass  = {cs.LG},
  url           = {https://arxiv.org/abs/2103.03874}
}

@misc{suzgun2023bbh,
  title         = {{Challenging BIG-Bench Tasks and Whether Chain-of-Thought Can Solve Them}},
  author        = {Mirac Suzgun and Nathan Scales and Nathanael Schärli and Sebastian Gehrmann and Yi Tay and Hyung Won Chung and Aakanksha Chowdhery and Quoc V. Le and Ed H. Chi and Denny Zhou and Jason Wei},
  year          = {2022},
  eprint        = {2210.09261},
  archivePrefix = {arXiv},
  primaryClass  = {cs.CL},
  url           = {https://arxiv.org/abs/2210.09261}
}

@misc{chen2021humaneval,
  title         = {{Evaluating Large Language Models Trained on Code}},
  author        = {Mark Chen and Jerry Tworek and Heewoo Jun and Qiming Yuan and Henrique Ponde de Oliveira Pinto and Jared Kaplan and Harri Edwards and Yuri Burda and Nicholas Joseph and Greg Brockman and Alex Ray and Raul Puri and Gretchen Krueger and Michael Petrov and Heidy Khlaaf and Girish Sastry and Pamela Mishkin and Brooke Chan and Scott Gray and Nick Ryder and Mikhail Pavlov and Alethea Power and Lukasz Kaiser and Mohammad Bavarian and Clemens Winter and Philippe Tillet and Felipe Petroski Such and Dave Cummings and Matthias Plappert and Fotios Chantzis and Elizabeth Barnes and Ariel Herbert-Voss and William Hebgen Guss and Alex Nichol and Alex Paino and Nikolas Tezak and Jie Tang and Igor Babuschkin and Suchir Balaji and Shantanu Jain and William Saunders and Christopher Hesse and Andrew N. Carr and Jan Leike and Josh Achiam and Vedant Misra and Evan Morikawa and Alec Radford and Matthew Knight and Miles Brundage and Mira Murati and Katie Mayer and Peter Welinder and Bob McGrew and Dario Amodei and Sam McCandlish and Ilya Sutskever and Wojciech Zaremba},
  year          = {2021},
  eprint        = {2107.03374},
  archivePrefix = {arXiv},
  primaryClass  = {cs.LG},
  url           = {https://arxiv.org/abs/2107.03374}
}

@misc{austin2021mbpp,
  title         = {{Program Synthesis with Large Language Models}},
  author        = {Jacob Austin and Augustus Odena and Maxwell Nye and Maarten Bosma and Henryk Michalewski and David Dohan and Ellen Jiang and Carrie Cai and Michael Terry and Quoc Le and Charles Sutton},
  year          = {2021},
  eprint        = {2108.07732},
  archivePrefix = {arXiv},
  primaryClass  = {cs.PL},
  url           = {https://arxiv.org/abs/2108.07732}
}

@misc{liu2023evalplus,
  title         = {{Is Your Code Generated by ChatGPT Really Correct? Rigorous Evaluation of Large Language Models for Code Generation}},
  author        = {Jiawei Liu and Chunqiu Steven Xia and Yuyao Wang and Lingming Zhang},
  year          = {2023},
  eprint        = {2305.01210},
  archivePrefix = {arXiv},
  primaryClass  = {cs.SE},
  url           = {https://arxiv.org/abs/2305.01210}
}

@misc{jain2024livecodebench,
  title         = {{LiveCodeBench: Holistic and Contamination Free Evaluation of Large Language Models for Code}},
  author        = {Naman Jain and King Han and Alex Gu and Wen-Ding Li and Fanjia Yan and Tianjun Zhang and Sida Wang and Armando Solar-Lezama and Koushik Sen and Ion Stoica},
  year          = {2024},
  eprint        = {2403.07974},
  archivePrefix = {arXiv},
  primaryClass  = {cs.SE},
  url           = {https://arxiv.org/abs/2403.07974}
}

@misc{gu2024cruxeval,
  title         = {{CRUXEval: A Benchmark for Code Reasoning, Understanding and Execution}},
  author        = {Alex Gu and Baptiste Rozière and Hugh Leather and Armando Solar-Lezama and Gabriel Synnaeve and Sida I. Wang},
  year          = {2024},
  eprint        = {2401.03065},
  archivePrefix = {arXiv},
  primaryClass  = {cs.SE},
  url           = {https://arxiv.org/abs/2401.03065}
}

@inproceedings{kembhavi2016ai2d,
  title     = {A Diagram is Worth a Dozen Images},
  author    = {Kembhavi, Aniruddha and Salvato, Mike and Kolve, Eric and Seo, Minjoon and Hajishirzi, Hannaneh and Farhadi, Ali},
  booktitle = {European Conference on Computer Vision},
  year      = {2016},
  eprint    = {1603.07396},
  archivePrefix = {arXiv},
  url       = {https://arxiv.org/abs/1603.07396}
}

@misc{xiao2024logicvista,
  title         = {{LogicVista: Multimodal LLM Logical Reasoning Benchmark in Visual Contexts}},
  author        = {Yijia Xiao and Edward Sun and Tianyu Liu and Wei Wang},
  year          = {2024},
  eprint        = {2407.04973},
  archivePrefix = {arXiv},
  primaryClass  = {cs.AI},
  url           = {https://arxiv.org/abs/2407.04973}
}

@misc{qiao2025wemath,
  title         = {{We-Math: Does Your Large Multimodal Model Achieve Human-like Mathematical Reasoning?}},
  author        = {Runqi Qiao and Qiuna Tan and Guanting Dong and Minhui Wu and Chong Sun and Xiaoshuai Song and Zhuoma GongQue and Shanglin Lei and Zhe Wei and Miaoxuan Zhang and Runfeng Qiao and Yifan Zhang and Xiao Zong and Yida Xu and Muxi Diao and Zhimin Bao and Chen Li and Honggang Zhang},
  year          = {2024},
  eprint        = {2407.01284},
  archivePrefix = {arXiv},
  primaryClass  = {cs.AI},
  url           = {https://arxiv.org/abs/2407.01284}
}

@misc{yue2025mmmupro,
  title         = {{MMMU-Pro: A More Robust Multi-discipline Multimodal Understanding Benchmark}},
  author        = {Xiang Yue and Tianyu Zheng and Yuansheng Ni and Yubo Wang and Kai Zhang and Shengbang Tong and Yuxuan Sun and Botao Yu and Ge Zhang and Huan Sun and Yu Su and Wenhu Chen and Graham Neubig},
  year          = {2024},
  eprint        = {2409.02813},
  archivePrefix = {arXiv},
  primaryClass  = {cs.CL},
  url           = {https://arxiv.org/abs/2409.02813}
}

@misc{lu2024mathvista,
  title         = {{MathVista: Evaluating Mathematical Reasoning of Foundation Models in Visual Contexts}},
  author        = {Pan Lu and Hritik Bansal and Tony Xia and Jiacheng Liu and Chunyuan Li and Hannaneh Hajishirzi and Hao Cheng and Kai-Wei Chang and Michel Galley and Jianfeng Gao},
  year          = {2023},
  eprint        = {2310.02255},
  archivePrefix = {arXiv},
  primaryClass  = {cs.CV},
  url           = {https://arxiv.org/abs/2310.02255}
}

@misc{cheng2025simplevqa,
  title         = {{SimpleVQA: Multimodal Factuality Evaluation for Multimodal Large Language Models}},
  author        = {Xianfu Cheng and Wei Zhang and Shiwei Zhang and Jian Yang and Xiangyuan Guan and Xianjie Wu and Xiang Li and Ge Zhang and Jiaheng Liu and Yuying Mai and Yutao Zeng and Zhoufutu Wen and Ke Jin and Baorui Wang and Weixiao Zhou and Yunhong Lu and Tongliang Li and Wenhao Huang and Zhoujun Li},
  year          = {2025},
  eprint        = {2502.13059},
  archivePrefix = {arXiv},
  primaryClass  = {cs.CL},
  url           = {https://arxiv.org/abs/2502.13059}
}

@misc{gu2025chinesesimplevqa,
  title         = {{"See the World, Discover Knowledge": A Chinese Factuality Evaluation for Large Vision Language Models}},
  author        = {Jihao Gu and Yingyao Wang and Pi Bu and Chen Wang and Ziming Wang and Tengtao Song and Donglai Wei and Jiale Yuan and Yingxiu Zhao and Yancheng He and Shilong Li and Jiaheng Liu and Meng Cao and Jun Song and Yingshui Tan and Xiang Li and Wenbo Su and Zhicheng Zheng and Xiaoyong Zhu and Bo Zheng},
  year          = {2025},
  eprint        = {2502.11718},
  archivePrefix = {arXiv},
  primaryClass  = {cs.CL},
  url           = {https://arxiv.org/abs/2502.11718}
}

@misc{zhou2026worldvqa,
  title         = {{WorldVQA: Measuring Atomic World Knowledge in Multimodal Large Language Models}},
  author        = {Runjie Zhou and Youbo Shao and Haoyu Lu and Bowei Xing and Tongtong Bai and Yujie Chen and Jie Zhao and Lin Sui and Haotian Yao and Zijia Zhao and Hao Yang and Haoning Wu and Zaida Zhou and Jinguo Zhu and Zhiqi Huang and Yiping Bao and Yangyang Liu and Y. Charles and Xinyu Zhou},
  year          = {2026},
  eprint        = {2602.02537},
  archivePrefix = {arXiv},
  primaryClass  = {cs.CV},
  url           = {https://arxiv.org/abs/2602.02537}
}

@misc{chen2026babyvision,
  title         = {{BabyVision: Visual Reasoning Beyond Language}},
  author        = {Liang Chen and Weichu Xie and Yiyan Liang and Hongfeng He and Hans Zhao and Zhibo Yang and Zhiqi Huang and Haoning Wu and Haoyu Lu and Y. charles and Yiping Bao and Yuantao Fan and Guopeng Li and Haiyang Shen and Xuanzhong Chen and Wendong Xu and Shuzheng Si and Zefan Cai and Wenhao Chai and Ziqi Huang and Fangfu Liu and Tianyu Liu and Baobao Chang and Ming Wu and Xiaobo Hu and Kaiyuan Chen and Yixin Ren and Yang Liu and Yuan Gong and Kuan Li},
  year          = {2026},
  eprint        = {2601.06521},
  archivePrefix = {arXiv},
  primaryClass  = {cs.CV},
  url           = {https://arxiv.org/abs/2601.06521}
}

@misc{lin2026perceptionbench,
  title         = {{PerceptionBench: Evaluating Atomic Visual Perception in Multimodal Large Language Models}},
  author        = {Zichao Lin and Yifeng Xie and Bowen Qu and Haiming Wang and Jia Li and Haoning Wu and Yuhao Dong and Zuhao Yang and Jinguo Zhu and Haoyu Lu and Zijia Zhao and Tongtian Yue and Zhangyang Qi and Junwei Yang and Mengfan Dong and Peizhou Cao and Chenzhuang Du and Zaida Zhou and Haotian Yao and Hao Yang and Hongcheng Gao and Lin Sui and Weihong Li and Xinxing Zu and Jia Chen and Yao Wang and Xiaoxue Wu and Yalin Wang and Y. Charles and Yiping Bao and Yangyang Liu and Zhiqi Huang and Xinyu Zhou},
  year          = {2026},
  eprint        = {2607.24957},
  archivePrefix = {arXiv},
  primaryClass  = {cs.CV},
  url           = {https://arxiv.org/abs/2607.24957}
}

@misc{xai2024realworldqa,
  title        = {{Grok-1.5 Vision} Preview},
  author       = {{xAI}},
  year         = {2024},
  howpublished = {\url{https://x.ai/news/grok-1.5v}},
  note         = {Introduces the RealWorldQA benchmark}
}

@misc{chen2024mmstar,
  title         = {{Are We on the Right Way for Evaluating Large Vision-Language Models?}},
  author        = {Lin Chen and Jinsong Li and Xiaoyi Dong and Pan Zhang and Yuhang Zang and Zehui Chen and Haodong Duan and Jiaqi Wang and Yu Qiao and Dahua Lin and Feng Zhao},
  year          = {2024},
  eprint        = {2403.20330},
  archivePrefix = {arXiv},
  primaryClass  = {cs.CV},
  url           = {https://arxiv.org/abs/2403.20330}
}

@misc{wang2024charxiv,
  title         = {{CharXiv: Charting Gaps in Realistic Chart Understanding in Multimodal LLMs}},
  author        = {Zirui Wang and Mengzhou Xia and Luxi He and Howard Chen and Yitao Liu and Richard Zhu and Kaiqu Liang and Xindi Wu and Haotian Liu and Sadhika Malladi and Alexis Chevalier and Sanjeev Arora and Danqi Chen},
  year          = {2024},
  eprint        = {2406.18521},
  archivePrefix = {arXiv},
  primaryClass  = {cs.CL},
  url           = {https://arxiv.org/abs/2406.18521}
}

@misc{fu2024ocrbenchv2,
  title         = {{OCRBench v2: An Improved Benchmark for Evaluating Large Multimodal Models on Visual Text Localization and Reasoning}},
  author        = {Ling Fu and Zhebin Kuang and Jiajun Song and Mingxin Huang and Biao Yang and Yuzhe Li and Linghao Zhu and Qidi Luo and Xinyu Wang and Hao Lu and Zhang Li and Guozhi Tang and Bin Shan and Chunhui Lin and Qi Liu and Binghong Wu and Hao Feng and Hao Liu and Can Huang and Jingqun Tang and Wei Chen and Lianwen Jin and Yuliang Liu and Xiang Bai},
  year          = {2024},
  eprint        = {2501.00321},
  archivePrefix = {arXiv},
  primaryClass  = {cs.CV},
  url           = {https://arxiv.org/abs/2501.00321}
}

@misc{wang2024allseeingv2,
  title         = {{The All-Seeing Project V2: Towards General Relation Comprehension of the Open World}},
  author        = {Weiyun Wang and Yiming Ren and Haowen Luo and Tiantong Li and Chenxiang Yan and Zhe Chen and Wenhai Wang and Qingyun Li and Lewei Lu and Xizhou Zhu and Yu Qiao and Jifeng Dai},
  year          = {2024},
  eprint        = {2402.19474},
  archivePrefix = {arXiv},
  primaryClass  = {cs.CV},
  url           = {https://arxiv.org/abs/2402.19474},
  note          = {Introduces the CRPE benchmark}
}

@misc{guan2024hallusionbench,
  title         = {{HallusionBench: An Advanced Diagnostic Suite for Entangled Language Hallucination and Visual Illusion in Large Vision-Language Models}},
  author        = {Tianrui Guan and Fuxiao Liu and Xiyang Wu and Ruiqi Xian and Zongxia Li and Xiaoyu Liu and Xijun Wang and Lichang Chen and Furong Huang and Yaser Yacoob and Dinesh Manocha and Tianyi Zhou},
  year          = {2023},
  eprint        = {2310.14566},
  archivePrefix = {arXiv},
  primaryClass  = {cs.CV},
  url           = {https://arxiv.org/abs/2310.14566}
}

@misc{du2024embspatial,
  title         = {{EmbSpatial-Bench: Benchmarking Spatial Understanding for Embodied Tasks with Large Vision-Language Models}},
  author        = {Mengfei Du and Binhao Wu and Zejun Li and Xuanjing Huang and Zhongyu Wei},
  year          = {2024},
  eprint        = {2406.05756},
  archivePrefix = {arXiv},
  primaryClass  = {cs.AI},
  url           = {https://arxiv.org/abs/2406.05756}
}

@misc{paiss2023countbench,
  title         = {{Teaching CLIP to Count to Ten}},
  author        = {Roni Paiss and Ariel Ephrat and Omer Tov and Shiran Zada and Inbar Mosseri and Michal Irani and Tali Dekel},
  year          = {2023},
  eprint        = {2302.12066},
  archivePrefix = {arXiv},
  primaryClass  = {cs.CV},
  url           = {https://arxiv.org/abs/2302.12066}
}

@misc{beyer2024paligemma,
  title         = {{PaliGemma: A versatile 3B VLM for transfer}},
  author        = {Lucas Beyer and Andreas Steiner and André Susano Pinto and Alexander Kolesnikov and Xiao Wang and Daniel Salz and Maxim Neumann and Ibrahim Alabdulmohsin and Michael Tschannen and Emanuele Bugliarello and Thomas Unterthiner and Daniel Keysers and Skanda Koppula and Fangyu Liu and Adam Grycner and Alexey Gritsenko and Neil Houlsby and Manoj Kumar and Keran Rong and Julian Eisenschlos and Rishabh Kabra and Matthias Bauer and Matko Bošnjak and Xi Chen and Matthias Minderer and Paul Voigtlaender and Ioana Bica and Ivana Balazevic and Joan Puigcerver and Pinelopi Papalampidi and Olivier Henaff and Xi Xiong and Radu Soricut and Jeremiah Harmsen and Xiaohua Zhai},
  year          = {2024},
  eprint        = {2407.07726},
  archivePrefix = {arXiv},
  primaryClass  = {cs.CV},
  url           = {https://arxiv.org/abs/2407.07726},
  note          = {Introduces the CountBenchQA benchmark}
}

@inproceedings{kazemzadeh2014referitgame,
  title     = {{ReferItGame}: Referring to Objects in Photographs of Natural Scenes},
  author    = {Kazemzadeh, Sahar and Ordonez, Vicente and Matten, Mark and Berg, Tamara},
  booktitle = {Proceedings of the 2014 Conference on Empirical Methods in Natural Language Processing ({EMNLP})},
  year      = {2014},
  pages     = {787-798},
  publisher = {Association for Computational Linguistics},
  doi       = {10.3115/v1/D14-1086},
  url       = {https://aclanthology.org/D14-1086/}
}

@misc{yu2016refcoco,
  title         = {{Modeling Context in Referring Expressions}},
  author        = {Licheng Yu and Patrick Poirson and Shan Yang and Alexander C. Berg and Tamara L. Berg},
  year          = {2016},
  eprint        = {1608.00272},
  archivePrefix = {arXiv},
  primaryClass  = {cs.CV},
  url           = {https://arxiv.org/abs/1608.00272}
}

@misc{hsieh2024ruler,
  title         = {{RULER: What's the Real Context Size of Your Long-Context Language Models?}},
  author        = {Cheng-Ping Hsieh and Simeng Sun and Samuel Kriman and Shantanu Acharya and Dima Rekesh and Fei Jia and Yang Zhang and Boris Ginsburg},
  year          = {2024},
  eprint        = {2404.06654},
  archivePrefix = {arXiv},
  primaryClass  = {cs.CL},
  url           = {https://arxiv.org/abs/2404.06654}
}

@inproceedings{bai2025longbenchv2,
  title         = {{LongBench v2: Towards Deeper Understanding and Reasoning on Realistic Long-context Multitasks}},
  author        = {Yushi Bai and Shangqing Tu and Jiajie Zhang and Hao Peng and Xiaozhi Wang and Xin Lv and Shulin Cao and Jiazheng Xu and Lei Hou and Yuxiao Dong and Jie Tang and Juanzi Li},
  booktitle     = {Proceedings of the 63rd Annual Meeting of the Association for Computational Linguistics (Volume 1: Long Papers)},
  year          = {2025},
  pages         = {3639-3664},
  eprint        = {2412.15204},
  archivePrefix = {arXiv},
  primaryClass  = {cs.CL},
  url           = {https://arxiv.org/abs/2412.15204}
}

@misc{vodrahalli2024michelangelo,
  title         = {{Michelangelo: Long Context Evaluations Beyond Haystacks via Latent Structure Queries}},
  author        = {Kiran Vodrahalli and Santiago Ontanon and Nilesh Tripuraneni and Kelvin Xu and Sanil Jain and Rakesh Shivanna and Jeffrey Hui and Nishanth Dikkala and Mehran Kazemi and Bahare Fatemi and Rohan Anil and Ethan Dyer and Siamak Shakeri and Roopali Vij and Harsh Mehta and Vinay Ramasesh and Quoc Le and Ed Chi and Yifeng Lu and Orhan Firat and Angeliki Lazaridou and Jean-Baptiste Lespiau and Nithya Attaluri and Kate Olszewska},
  year          = {2024},
  eprint        = {2409.12640},
  archivePrefix = {arXiv},
  primaryClass  = {cs.CL},
  url           = {https://arxiv.org/abs/2409.12640},
  note          = {MRCR v2 data: \url{https://github.com/google-deepmind/eval_hub/tree/master/eval_hub/mrcr_v2}}
}

@article{lu2025youtullm,
  author        = {Junru Lu and Jiarui Qin and Lingfeng Qiao and Yinghui Li and Xinyi Dai and others},
  title         = {{Youtu-LLM}: Unlocking the Native Agentic Potential for Lightweight Large Language Models},
  journal       = {arXiv preprint arXiv:2512.24618},
  year          = {2025},
  eprint        = {2512.24618},
  archivePrefix = {arXiv},
  primaryClass  = {cs.CL},
  url           = {https://arxiv.org/abs/2512.24618}
}

@article{pan2026elephantbench,
  author        = {Zhuoshi Pan and Junru Lu and Yan Qian and H. Vicky Zhao and Di Yin and Xing Sun},
  title         = {Blind Men and the Elephant: Probing the Epistemic Myopia of {LLMs} under Long-Tail Divergent Knowledge},
  journal       = {arXiv preprint arXiv:2608.28478},
  year          = {2026},
  eprint        = {2608.28478},
  archivePrefix = {arXiv},
  primaryClass  = {cs.CL},
  url           = {https://arxiv.org/abs/2608.28478}
}

@article{tao2026textbooks,
  author        = {Jiawen Tao and Miao Peng and Yaoming Li and Xiaokun Yuan and Mengzhou Wu and Wenhan Yu and Guoan Wang and Nuo Chen and Tong Yang and Maxm Pan},
  title         = {Beyond Rephrasing: Book-Level Organization Improves Synthetic Textbook Data for Mid-Training},
  journal       = {arXiv preprint arXiv:2607.28109},
  year          = {2026},
  eprint        = {2607.28109},
  archivePrefix = {arXiv},
  primaryClass  = {cs.AI},
  url           = {https://arxiv.org/abs/2607.28109}
}

@article{raecompressive2019,
  author = {Rae, Jack W and Potapenko, Anna and Jayakumar, Siddhant M and
            Hillier, Chloe and Lillicrap, Timothy P},
  title = {Compressive Transformers for Long-Range Sequence Modelling},
  journal = {arXiv preprint},
  url = {https://arxiv.org/abs/1911.05507},
  year = {2019},
}

@inproceedings{wang2025mmlongbench,
      title={MMLongBench: Benchmarking Long-Context Vision-Language Models Effectively and Thoroughly}, 
      author={Zhaowei Wang and Wenhao Yu and Xiyu Ren and Jipeng Zhang and Yu Zhao and Rohit Saxena and Liang Cheng and Ginny Wong and Simon See and Pasquale Minervini and Yangqiu Song and Mark Steedman},
      year={2025},
      eprint={2505.10610},
      booktitle={The 39th (2025) Annual Conference on Neural Information Processing Systems},
      url={https://arxiv.org/abs/2505.10610}, 
}

@article{wang2024needle,
  title={Needle In A Multimodal Haystack},
  author={Wang, Weiyun and Zhang, Shuibo and Ren, Yiming and Duan, Yuchen and Li, Tiantong and Liu, Shuo and Hu, Mengkang and Chen, Zhe and Zhang, Kaipeng and Lu, Lewei and others},
  journal={arXiv preprint arXiv:2406.07230},
  year={2024}
}

@article{brown2026unbundling,
  author    = {Ellis Brown and Chris Hoang and Shengyi Qian and others},
  title     = {Unbundling Multimodal Vision Encoding},
  journal   = {arXiv preprint arXiv:2611.17334},
  year      = {2026},
  eprint    = {2611.17334},
  archiveprefix = {arXiv},
  url       = {https://arxiv.org/abs/2611.17334}
}

@misc{elango2026latentmoe,
      title={LatentMoE: Toward Optimal Accuracy per FLOP and Parameter in Mixture of Experts}, 
      author={Venmugil Elango and Nidhi Bhatia and Roger Waleffe and Rasoul Shafipour and Tomer Asida and Abhinav Khattar and Nave Assaf and Maximilian Golub and Joey Guman and Tiyasa Mitra and Ritchie Zhao and Ritika Borkar and Ran Zilberstein and Mostofa Patwary and Mohammad Shoeybi and Bita Rouhani},
      year={2026},
      eprint={2601.18089},
      archivePrefix={arXiv},
      primaryClass={cs.LG},
      url={https://arxiv.org/abs/2601.18089}, 
}

\appendix
\section{Contributions and Acknowledgments}
\label{sec:contributions}
We would like to express our sincere gratitude to all contributors, including those not listed in the paper, for their invaluable support and efforts. The contributors within each group are listed in no particular order.

\vspace{0.8em}
\noindent\textbf{Core Contributors}
\vspace{0.3em}

\noindent\begin{tabularx}{\textwidth}{@{}XXXX@{}}
Qinglong Zhang & Yun Liu & Yunjia Xi & Junru Lu \\
Lingfeng Qiao & Xinyi Dai & Qianwen Zhang & Haoyu Cao \\
Di Yin$^\dagger$ & Qiang Lin \\
\end{tabularx}
\begingroup
\renewcommand\thefootnote{}\footnote{$^\dagger$Corresponding author: \texttt{endymecyyin@tencent.com}}
\addtocounter{footnote}{-1}
\endgroup

\vspace{0.8em}
\noindent\textbf{Contributors}
\vspace{0.3em}

\noindent\begin{tabularx}{\textwidth}{@{}XXXX@{}}
Xing Sun & Yinghui Li & Jiarui Qin & Yinsong Liu \\
Deqiang Jiang & Yunhang Shen & Ke Li  & Xin Li \\
Yongheng Zhang & Biao Liu & Daixian Liu & Ao Shen \\
Xianjie Liu & Qian Cao & Xueyan Wu & Guotao Liang \\
Yucheng Zhou & Hang Zheng & Renting Rui & Qingyao Li \\
Yanshu Li & Zhuoshi Pan & Yingqi Shi & Jiayi Kuang \\
Tianxiang Jiang & Yilin Long

\end{tabularx}

\end{document}